\documentclass[runningheads]{llncs}
 
\usepackage{accv}

\usepackage{accvabbrv}

\usepackage{graphicx}
\usepackage{booktabs}

\usepackage[accsupp]{axessibility}  

\usepackage{multirow}
\usepackage{subcaption}
\usepackage{capt-of}
\usepackage{algorithm}
\usepackage{algorithmic}

\usepackage{amsmath}
\usepackage{amssymb}

\usepackage[pagebackref,breaklinks,colorlinks,citecolor=accvblue]{hyperref}

\usepackage{orcidlink}
\usepackage{todonotes}

\begin{document}

\title{On Adversarial Vulnerability of Vision-Language Models through the Lens of Intermediate Spectral Subspaces} 
\titlerunning{Adversarial Vulnerability in VLMs via Spectral Subspaces}

\author{Chethan Krishnamurthy Ramanaik \inst{1} \and 
Tobias Callies\inst{1} \and 
Michael Hecht\inst{2} \and 
Eirini Ntoutsi\inst{1}}

\authorrunning{C.K. Ramanaik et al.}

\institute{University of the Bundeswehr Munich, Germany \\
\email{\{chethan.krishnamurthy,tobias.callies,eirini.ntoutsi\}@unibw.de}\\
\and
University of Wrocław, Poland\\
\email{michael.hecht@math.uni.wroc.pl}}

\maketitle

\begin{abstract}
Adversarial vulnerability in deep neural networks (DNNs) has been studied from the perspectives of decision-boundary geometry, feature robustness, input–output Jacobians, and the instability of inverse problems. Here, we focus on the spectral structure of intermediate linear transformations that propagate information through modern DNNs, an unexplored mechanism of adversarial vulnerability. Specifically, we investigate transformer-based vision–language models, whose linear layers admit interpretable spectral decompositions and whose widespread adoption makes understanding their robustness increasingly important. We propose a white-box spectral-subspace-guided attack (SSGRA) that aligns intermediate representations with the subspace spanned by the bottom right singular vectors. Our experiments show improved attack effectiveness over existing baselines. In addition, SSGRA offers a spectral interpretation of adversarial vulnerability in VLMs, providing insights for improving their robustness.

\end{abstract}

\section{Introduction}
\label{sec:intro}

Deep neural networks, including modern vision--language models (VLMs), are known to be vulnerable to adversarial perturbations that substantially alter model predictions while remaining visually imperceptible~\cite{szegedy2013intriguing,goodfellow2014explaining, yin2023vlattack}. Despite extensive research, the mechanisms governing how such perturbations propagate through deep networks remain incompletely understood. Existing explanations primarily analyze adversarial vulnerability from input-space \cite{fawzi2018analysis,khoury2018geometry,Goodfellow-et-al-2016,song2017pixeldefend,samangouei2018defense,gilmer1801adversarial,zhang2019theoretically} or end-to-end perspectives, including decision-boundary geometry \cite{moosavi2017universal}, robust and non-robust features \cite{ilyas2019adversarial}, Jacobian analysis \cite{hoffman2019robust,jakubovitz2018improving,paniagua2025adversarial,khrulkov2018art}, inverse problem instability and Lipschitz properties \cite{antun2021deep,antun2020instabilities,gottschling2025troublesome}. Despite these advances, existing theories predominantly explain adversarial vulnerability from the input space or through end-to-end network properties, leaving the spectral behavior of intermediate linear transformations largely unexplored. 

Transformer-based VLMs provide a natural setting for such an analysis. Their architectures comprise numerous learnable intermediate linear transformations including the projection matrices in self-attention, feed-forward networks, and multimodal fusion modules \cite{vaswani2017attention,dosovitskiy2020image,gemmateam2025gemma3technicalreport,bai2025qwen25vltechnicalreport}, making spectral decomposition a principled tool for studying representations. Because singular values govern how different representation directions (singular vectors) are \emph{amplified} or \emph{suppressed} by each linear transformation, they naturally provide a lens for studying how adversarial signals propagate through transformer layers. Moreover, their widespread adoption makes them an important testbed for adversarial robustness. 

This motivates us to investigate adversarial vulnerability through the singular-vector basis of intermediate linear transformations. Inspired by the instability of ill-posed inverse problems, where near-null singular directions govern information loss, we study how adversarial intermediate representations align with top and bottom singular-vector subspaces during attack optimization. Guided by this perspective, we formulate a spectral-guidance principle and instantiate it through a white-box attack that serves to empirically validate the proposed mechanism. Our results suggest that, beyond constraining large singular values, explicitly controlling near-null singular directions may offer a complementary strategy for improving adversarial robustness. 

\begin{figure*}[t]
    \centering
    \includegraphics[width=0.7\textwidth]{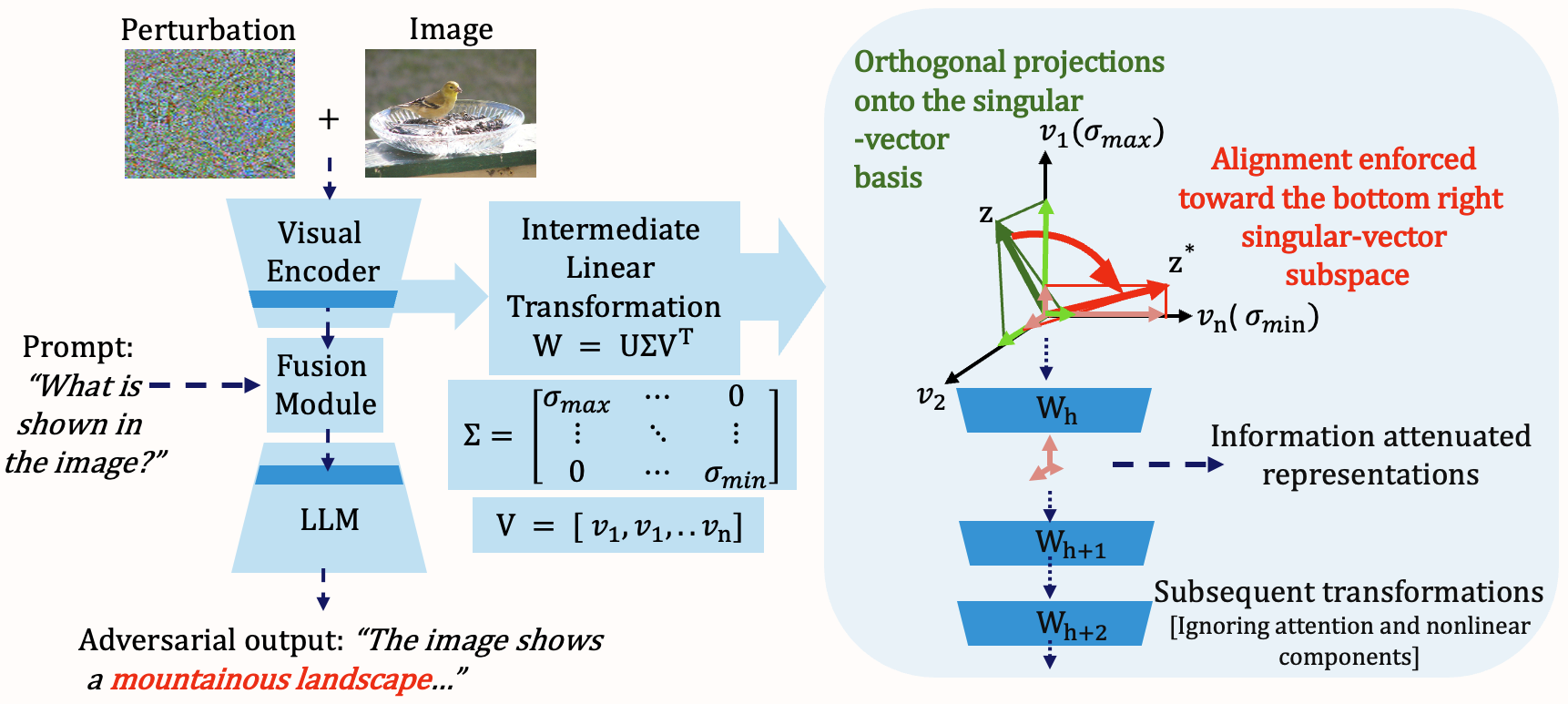}
    \caption{\textbf{Overview of the proposed spectral framework.}}
    \label{fig:overview}
\end{figure*}

\noindent\textbf{Contributions.} We identify bottom singular-vector subspaces of intermediate linear transformations as a previously overlooked spectral attack surface in transformer-based VLMs. We show that untargeted adversarial optimization naturally tends to increase the alignment of intermediate representations with these information-attenuating subspaces, even without explicit enforcement. Building on this insight, we propose the Spectral Subspace Guided Representation Attack (SSGRA), a spectrally guided white-box attack that demonstrates improved attack effectiveness compared with representative feature-space and output-space attacks on three state-of-the-art VLMs.

In the subsequent sections, the paper reviews the related work, followed by the preliminaries, methodology, and experimental evaluation. 

\section{Related Work}
We briefly review the theoretical foundations of adversarial vulnerability, followed by relevant representative methods for crafting adversarial examples.

\subsection{Theoretical Perspectives.}

\noindent\textbf{Manifolds \& Decision Boundaries.}
Adversarial vulnerability has been attributed to the geometry of high-dimensional decision boundaries \cite{fawzi2018analysis,khoury2018geometry}. Adversarial examples have also been explained as off-manifold inputs \cite{Goodfellow-et-al-2016,song2017pixeldefend,samangouei2018defense,gilmer1801adversarial,zhang2019theoretically}, while universal perturbations suggest that decision boundaries around different inputs share a low-dimensional subspace of normal vectors \cite{moosavi2017universal}.

\noindent\textbf{Non-robust features. } Standard training encourages models to rely on highly predictive but non-robust features, whose sensitivity to small input perturbations leads to adversarial vulnerability \cite{ilyas2019adversarial}.

\noindent\textbf{Linearity approximation.}
Adversarial vulnerability has been attributed to local linearity, where high-dimensional gradient--perturbation interactions amplify small perturbations \cite{goodfellow2014explaining}. In linear settings, vulnerability arises when the input lies close to the decision boundary \cite{etmann2019connection,hein2017formal}. However, experimental evidence suggests that DNNs are only locally linear and remain globally nonlinear \cite{luo2015foveation}.


\noindent\textbf{Internal weights.}
Large singular values of weight matrices have been linked to adversarial vulnerability through their connection to local Lipschitz constants, motivating spectral regularization \cite{szegedy2013intriguing,cisse2017parseval,yoshida2017spectral}. Near-zero singular values suppress gradient flow, while restoring these gradients strengthens adversarial attacks \cite{ramanaik2025grill,gupta2022improved}.

\noindent\textbf{End-to-end Jacobians.}
Adversarial vulnerability has been linked to large input--output Jacobians \cite{hoffman2019robust,jakubovitz2018improving}. Targeted perturbations can be expressed as linear combinations of the right singular vectors of the logit-to-image Jacobian \cite{paniagua2025adversarial}, while intermediate layer Jacobians identify sensitive input directions \cite{khrulkov2018art}. Noisy and poorly aligned input Jacobians have also been associated with adversarial vulnerability \cite{chan2019jacobian}.

\noindent\textbf{Instability of inverse problems.}
Studies on the instability of inverse problems show that information lost along null-space and near-null singular directions leads to unstable reconstruction during inversion \cite{antun2021deep,antun2020instabilities,gottschling2025troublesome,ramanaik2024ensuring}. However, most adversarial robustness research has focused on constraining large singular values through spectral normalization and Lipschitz regularization \cite{barrett2022certifiably,fazlyab2019efficient,yoshida2017spectral,miyato2018spectral,cisse2017parseval,gulrajani2017improved,virmaux2018lipschitz}, as well as on the implicit self-regularization of dominant singular values \cite{martin2021implicit}. Comparatively little attention has been paid to whether bottom singular-vector subspaces of intermediate linear transformations constitute a source of adversarial vulnerability, which is the focus of this work.

\subsection{Methods for Generating Adversarial Examples}
Adversarial attacks have evolved from early gradient-based methods such as FGSM, PGD, and optimization-based attacks \cite{szegedy2013intriguing,goodfellow2014explaining,mkadry2017towards,carlini2017towards} to feature-alignment attacks \cite{sabour2015adversarial,inkawhich2019feature} and attacks on generative models \cite{gondim2018adversarial,cemgil2020adversarially,willetts2019improving,kuzina2021diagnosing,ramanaik2025grill}. Recent VLM attacks have explored computational availability \cite{gao2024inducing}, cross-prompt transferability \cite{luo2024image}, CoT reasoning \cite{wang2024stop}, visual grounding \cite{gao2024adversarial}, gray-box SVD-based attacks \cite{liu2026attacking}, black-box attacks \cite{dong2023robust}, and behavior hijacking \cite{bailey2023image} targeting different threat models, tasks, or objectives including targeted attacks, visual reasoning, gray-box or black-box settings, and behavior control. 


We focus on untargeted white-box attacks to study adversarial vulnerability through information attenuation rather than predefined attack objectives. Accordingly, we compare against representative untargeted white-box feature- and output-level attacks. Feature Discrepancy Attack (FDA) perturbs inputs by maximizing discrepancies between intermediate representations~\cite{ganeshan2019fda}. Similarly, Self-Supervised Perturbation Attack (SSPA) maximizes the discrepancy between clean and adversarial feature representations in pretrained models~\cite{naseer2020self,jia2025adversarial}. Dispersion Reduction Attack (DRA) minimizes the variance of intermediate features, forcing representations to collapse and become less discriminative \cite{lu2020enhancing}. Blockwise Similarity Attack (BSA) targets transformer-based VLMs by maximizing cosine discrepancies between block-wise intermediate representations of clean and adversarial inputs, thereby disrupting semantic alignment throughout the network \cite{yin2023vlattack}.

Beyond feature-space objectives, output-level attacks maximize the task loss using cross-entropy (CE) or negative log-likelihood \cite{cui2024robustness}. Entropy-Guided Attacks (EGA) maximize output entropy to induce uncertain model responses \cite{he2025few}. We compare against representative feature-space attacks (FDA, SSPA, DRA, and BSA) and output-level attacks (CE, EGA). 



\section{Preliminaries}
\label{sec:preliminaries}

\noindent{\textbf{Notation}.}
Let \(x\in\mathbb{R}^{d_x}\) denote a flattened input image. 
We consider perturbations constrained to the \(L_p\)-ball
$
B_c^p(x)
=
\left\{
x_a\in\mathbb{R}^{d_x}
\;\middle|\;
\|x_a-x\|_p\le c
\right\},
$
where \(c\) is the perturbation budget. A VLM can be abstractly described by a function
$
\mathcal{F}:\mathbb{R}^{d_x}\times\mathcal{P}\rightarrow\mathcal{\zeta},
$
where \(\mathcal{P}\) is the space of input prompts and \(\mathcal{\zeta}\) is the associated tokenizer dictionary.

\subsection*{VLM pipeline}
Modern VLMs typically consist of a visual encoder, a multimodal projection or fusion module, and a large language model (LLM).

\noindent{\textbf{Visual Encoder.}} The visual encoder $\phi:\mathbb{R}^{d_x}\rightarrow\mathbb{R}^{N_v\times d_v}$ consists of $K$ sequentially applied blocks $\phi_k$, each producing a visual token representation $h^v_k\in\mathbb{R}^{N_v\times d_v}$ consisting of  \(N_v\)  visual tokens of the visual embedding dimensionality $d_v$ for a given input image \(x\).

\noindent{\textbf{Fusion Module. }} The final visual embedding $h_K^v(x)=\phi(x)$ is projected into the embedding space of the language model through a projection module \(P\), outputting $H^v_0=P(\phi(x)) \in \mathbb{R}^{N_v\times d}$, where \(d\) is the hidden dimension of the language model. For a given textual prompt, let $H_0^t\in\mathbb{R}^{N_t\times d}$ denote the corresponding textual embedding, using \(N_t\) tokens. The concatenated visual and textual embeddings then form the input sequence to the language model, $H_0 = \bigl[\,H_0^v;\;H_0^t\,\bigr] \in \mathbb{R}^{N\times d}$, where $N=N_v+N_t$
is the total sequence length after multimodal fusion. 

\noindent{\textbf{LLM.}} The LLM consists of $L$ transformer blocks and a final decoder. The \(\ell\)-th transformer block outputs the intermediate tokenized representation $H_\ell\in\mathbb{R}^{N\times d}$.


\subsection*{Singular Subspaces and Orthogonal Projection}
For a linear transformation $W\in\mathbb{R}^{m\times n}$, the singular value decomposition (SVD) is given by
$W = U\Sigma V^\top$,
where \(U\in\mathbb{R}^{m\times m}\) and \(V\in\mathbb{R}^{n\times n}\) are orthonormal matrices, and
$
\Sigma\in\mathbb{R}^{m\times n}$
contains the singular values of \(W\), ordered as
$
\sigma_1 \ge \sigma_2 \ge \cdots \ge \sigma_r \ge 0,
$ on its diagonal entries: $\Sigma_{ii}=\sigma_i$ for 
\(1\leq i \leq r=\mathrm{rank}(W)\leq \min\{m,n\}\), and zeros otherwise. The columns of
$
V=[v_1,v_2,\ldots,v_n]
$
are referred to as the \emph{right singular vectors} of \(W\).
For a given \(1\le k\le n\), let
\[
V_k^{\mathrm{top}}
=
\{v_1,\ldots,v_k\},
\qquad
V_k^{\mathrm{bottom}}
=
\{v_{n-k+1},\ldots,v_n\},
\]
denote the top-\(k\) and bottom-\(k\) right singular vectors of \(W\), respectively, and the corresponding subspaces are referred to as top-\(k\) and bottom-\(k\) singular subspaces, and denoted by \(\mathrm{span}(V_k^{\mathrm{top}})\) and \(\mathrm{span}(V_k^{\mathrm{bottom}})\).


Due to the orthonormality of $V$, we can measure the alignment of a vector $z\in\mathbb{R}^n$ with such subspaces using the norm of their projections and the identity $\|z\|^2_2=\|P_k^{\mathrm{top}}(z)\|^2_2+\|P_{n-k}^{\mathrm{bottom}}(z)\|^2_2$. Here, $P_k^{\mathrm{top}}(z)=(v_1^\top z, \ldots, v_k^\top z )$ denotes the projection onto $\mathrm{span}(V_k^{\mathrm{top}})$, and $P_k^{\mathrm{bottom}}$ is defined analogously. The projection norms motivate the interpretation as corresponding \emph{energies}.

\subsection*{Effect of Near-Null Singular Directions: An Analogy to Ill-Posed Inverse Problems}
\label{sec:IP}

Consider the linear transformation $W:\mathbb{R}^n\rightarrow\mathbb{R}^m$ with reconstruction map $\Gamma:\mathbb{R}^m\rightarrow\mathbb{R}^n$. The instability of ill-posed inverse problems states that, in general, stable reconstruction cannot be guaranteed \cite{szegedy2013intriguing,antun2020instabilities,huang2018some}, with deeper theoretical treatments in \cite{gottschling2025troublesome,antun2020instabilities}. A common characterization of this instability is the local Lipschitz constant of the reconstruction map at a measurement $y\in\mathbb{R}^m$:
\[
L_\varepsilon(\Gamma,y)
=
\sup_{0<\|y'-y\|<\varepsilon}
\frac{\|\Gamma(y')-\Gamma(y)\|_2}
{\|y'-y\|_2},
\qquad
\varepsilon>0,
\]
where $y'\in\mathbb{R}^m$ denotes a perturbed measurement. The local Lipschitz constant may become unbounded, causing large reconstruction errors. We draw an analogy between this instability phenomenon and intermediate linear transformations in VLMs, extending the inverse-problem viewpoint of \cite{jain2025automated}. In inverse problems, the reconstruction map attempts to recover information attenuated by the forward operator, whereas VLMs propagate representations through successive transformations. Consequently, representations aligned with the bottom singular-vector subspace of $W$ are strongly attenuated by the forward transformation, motivating our study of bottom singular-vector subspace alignment.

\subsection*{Spectral Alignment Measure}

For a set of vectors
$
H=\{h_1,\ldots,h_N\}\subset\mathbb{R}^n
$ (as arising in the tokenized representations with $n=d$), we quantify the average alignment with a subspace by computing the average projection energy of the normalized token representations. For $\mathrm{span}(V_k^{\mathrm{top}})$ this takes the form:

\begin{equation}
\label{alignmentMeasure}
\Psi_k(H,V^{\mathrm{top}}_k)
=
\Psi_k(H,\mathrm{span}(V^{\mathrm{top}}_k))
=
\frac{1}{N}
\sum_{i=1}^{N}
\sum_{j=1}^{k}
\left(
\frac{v_j^\top h_i}
{\|h_i\|_2}
\right)^2.
\end{equation}


Consequently,
$0 \le \Psi_k(H,V_k^{\mathrm{top}}) \le 1$, and larger values indicate greater concentration of representational energy within the selected singular subspace. The alignment measure is defined analogously for the bottom-$k$ singular subspace $V_k^{\mathrm{bottom}}$. In the proposed attack, we instantiate this general definition using a fixed subspace dimension $s$, i.e., $\Psi_s(\cdot, V_s^{\mathrm{bottom}})$.




\subsection*{Threat Model}
We consider an untargeted white-box attack where the adversary has full access to the model architecture, parameters, and intermediate representations. Given an image $x$ and a text prompt, the adversary generates an adversarial image $x_a$ that degrades the model's response while remaining visually similar to $x$. The perturbation is constrained by an $L_\infty$ budget $c$, i.e.,
$
x_a \in B_c^\infty(x).
$
The prompt, model parameters, and inference procedure remain unchanged, and each image is attacked independently.

\section{A Spectral Framework for Adversarial Vulnerability}
We first introduce the Spectral Subspace Guided Representation Attack (SSGRA), which instantiates the proposed spectral-guidance principle, and then present the layer-wise probing framework used to analyze the spectral dynamics of adversarial optimization.

\subsection{Spectral Subspace Guided Representation Attack (SSGRA)}
SSGRA extends the Blockwise Similarity Attack (BSA)~\cite{yin2023vlattack} by introducing a spectral guidance term based on the alignment measure in Eq.~\eqref{alignmentMeasure}. Motivated by the instability phenomenon (Section~\ref{sec:IP}), this guidance aligns intermediate adversarial representations with the bottom singular-vector subspaces of selected linear transformations, where information is most attenuated. We hypothesize that steering representations toward these subspaces progressively weakens semantic information propagation, improving attack effectiveness.

SSGRA combines two complementary objectives (Eq~\ref{eq:ssgra}). The first maximizes the discrepancy between clean and adversarial intermediate representations following BSA~\cite{yin2023vlattack}, thereby disrupting the learned feature hierarchy. The second maximizes the spectral alignment measure defined in Eq.~\eqref{alignmentMeasure}, encouraging adversarial representations to concentrate their energy within bottom singular-vector subspaces. 


For each selected layer $m\in\mathcal S$, let $z_m(\cdot)$ denote the collection of token representations immediately preceding the corresponding linear transformation, and let $V^{\mathrm{bottom}}_{m,s}$ denote the subspace spanned by the bottom-$s$ right singular vectors of that transformation. Since the textual prompt remains fixed during optimization, we suppress its dependence in the notation and write $h_k^v(x)$, $H_\ell(x)$, and $z_m(x)$ instead of $h_k^v(x,p)$, $H_\ell(x,p)$, and $z_m(x,p)$.
\begin{definition}[Spectral Subspace Guided Representation Attack (SSGRA)]
The SSGRA adversarial example is defined as the solution to the following optimization problem:
\begin{equation}
\label{eq:ssgra}
\begin{aligned}
x_a^*
=
\arg\max_{x_a\in B_c^p(x)}
\Bigg\{
&
-\lambda
\Bigg[
\sum_{k=1}^{K}
\sum_{j=1}^{N_v}
\cos
\!\left(
h_k^{v,(j)}(x),
h_k^{v,(j)}(x_a)
\right)
\\
&
+
\sum_{\ell=1}^{L}
\sum_{i=1}^{N}
\cos
\!\left(
H_\ell^{(i)}(x),
H_\ell^{(i)}(x_a)
\right)
\Bigg]
\\
&
+
(1-\lambda)
\sum_{m\in\mathcal S}
\Psi_s
\!\left(
z_m(x_a),
V^{\mathrm{bottom}}_{m,s}
\right)
\Bigg\}.
\end{aligned}
\end{equation}
where $h_k^{v,(j)}$ and $H_\ell^{(i)}$ denote the representations of the $j$-th visual token and the $i$-th multimodal token at the outputs of the $k$-th visual encoder block and the $\ell$-th LLM block, respectively. Furthermore, $\Psi_s(\cdot,\cdot)$ is the spectral alignment measure defined in Eq.~\eqref{alignmentMeasure}, $s$ denotes the chosen dimension of the bottom singular subspace, and $\lambda\in[0,1]$ controls the trade-off between representation discrepancy and spectral subspace alignment.
\end{definition}
The optimization procedure is summarized in Algorithm~\ref{alg:ssgra} in the Appendix. Rather than applying spectral guidance to all layers, we use it only on a selected subset of intermediate linear transformations, denoted by $\mathcal{S}$. The layers are selected by evaluating each visual encoder, fusion, and LLM layer independently on a small validation set and retaining those that yield the strongest attack performance. Developing adaptive layer-selection methods that avoid validation-based tuning is left for future work.

\subsection{Layer-wise Probing of Spectral Alignment}
To analyze the spectral alignment of adversarial representations, we perform layer-wise adversarial probing. For each transformer block \(i\), we generate an adversarial example by minimizing the cosine similarity between the clean and adversarial feature maps:
\begin{equation}
\label{eq:layerwise_probe}
x_{a,i}^{*}
=
\arg\min_{x_a \in B_{c}(x)}
\frac{
\left\langle H_i(x), H_i(x_a) \right\rangle_F
}{
\|H_i(x)\|_F \, \|H_i(x_a)\|_F
},
\end{equation}
where \(H_i(\cdot)\) denotes the feature map at layer \(i\), \(\langle \cdot,\cdot\rangle_F\) is the Frobenius inner product, and \(B_{c}(x)\) is the admissible perturbation set.

For each adversarial example \(x_{a,i}^{*}\), we compute the spectral subspace alignment across all vision and language layers using Eq.~\eqref{alignmentMeasure}. Repeating this procedure over all target layers and input samples enables us to analyze alignment with top and bottom singular-vector subspaces during and after attack optimization.

\subsection{Evaluation Metrics}

We evaluate attack effectiveness by comparing the adversarial output $\hat{y}_a$ with the corresponding clean output $\hat{y}_c$ using BERTScore~\cite{zhang2019bertscore} and ROUGE-L~\cite{lin2004rouge}. 
BERTScore measures semantic similarity using contextual token embeddings from RoBERTa-large.
ROUGE-L measures lexical similarity based on the longest common subsequence (LCS), capturing structural degradation of the generated response.
For both metrics, we report Precision, Recall, and F1, where lower scores indicate stronger attacks. Additional details are provided in Appendix~\ref{sec:evalDetails}.

\section{Experiments}
We evaluate the proposed attacks against representative baselines, analyze their spectral mechanisms, and present ablation studies.
\subsection{Experimental Setup}
\label{expSetup}
We evaluate attacks on different VLMs, namely Gemma-3 (4B) \cite{gemmateam2025gemma3technicalreport}, Qwen2.5-VL (7B) \cite{bai2025qwen25vltechnicalreport}, and LLaVA-1.5 
(7B) \cite{an2025llava}. Experiments are conducted on ImageNet \cite{deng2009imagenet}, whose
diverse visual categories enable assessment of generalization across image 
content. Given an input image and the prompt \textit{``What is shown in 
the image?''}, we optimize sample-specific adversarial perturbations to 
degrade the model's image description while remaining visually 
imperceptible. Each experimental instance is defined by a perturbation 
budget and an attack method, evaluated over 100 images. 
The perturbation 
budget ranges from 0.002 to 0.005 in the \(\ell_{\infty}\) norm, selected 
via grid search such that the lower bound captures the regime where outputs remain 
semantically similar across methods, and the upper bound where performance 
differences become pronounced. 

We compare SSGRA against six representative baselines BSA, DRA, FDA, EGA, SSPA, and NLL. 
All attacks are optimized using Adam following \cite{carlini2017towards} with a fixed computational budget of $1000$ gradient steps to ensure a fair comparison across methods and enable evaluation on a sufficiently large number of samples for statistically reliable quantitative results. Grid search over learning rates $\{10^{-2},\,5\times10^{-3},\,10^{-3},\,5\times10^{-4},\,10^{-4}\}$ identified $10^{-3}$ as consistently yielding the best attack performance across the three models.
The adversarial perturbation is initialized with small random noise of near-zero magnitude, following standard practice in iterative adversarial optimization.


\subsection{Quantitative comparison with State-of-the-Art Attacks}
\label{sec:quantitativeComparision}
\begin{figure*}[t]
\centering
\begin{subfigure}[t]{\textwidth}
\centering
\begin{subfigure}[b]{0.7\textwidth}
    \centering
    \includegraphics[width=\linewidth]{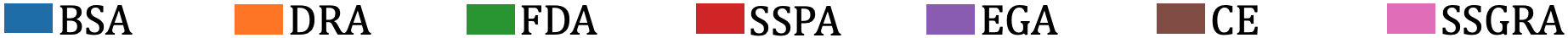}
\end{subfigure}
\begin{subfigure}[b]{0.3\textwidth}
    \centering
    \includegraphics[width=\linewidth, trim=0cm 0cm 0cm 2.4cm, clip]{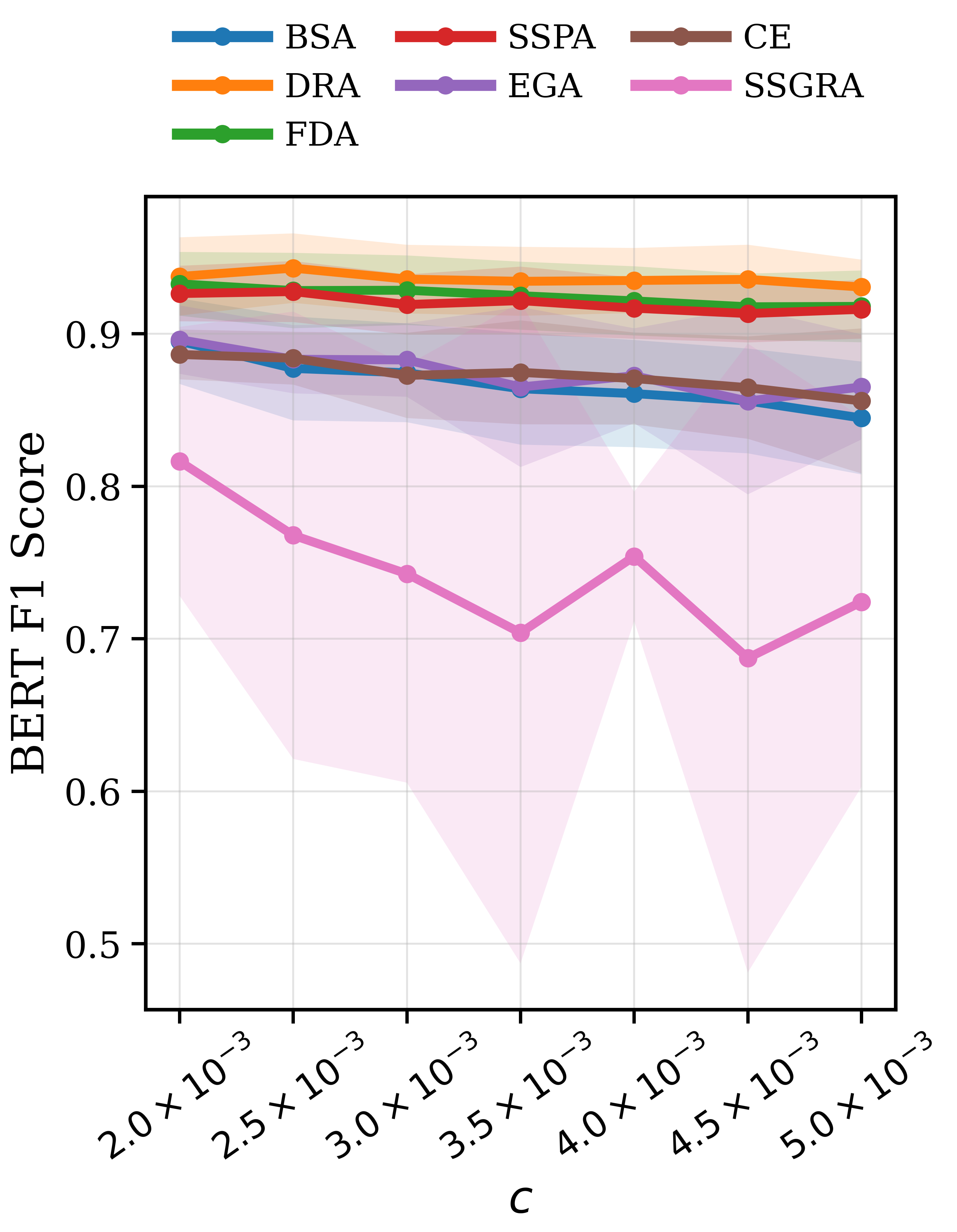}
    \caption{Qwen~2.5-VL BERT F1}
\end{subfigure}
\hfill
\begin{subfigure}[b]{0.3\textwidth}
    \centering
    \includegraphics[width=\linewidth, trim=0cm 0cm 0cm 2.4cm, clip]{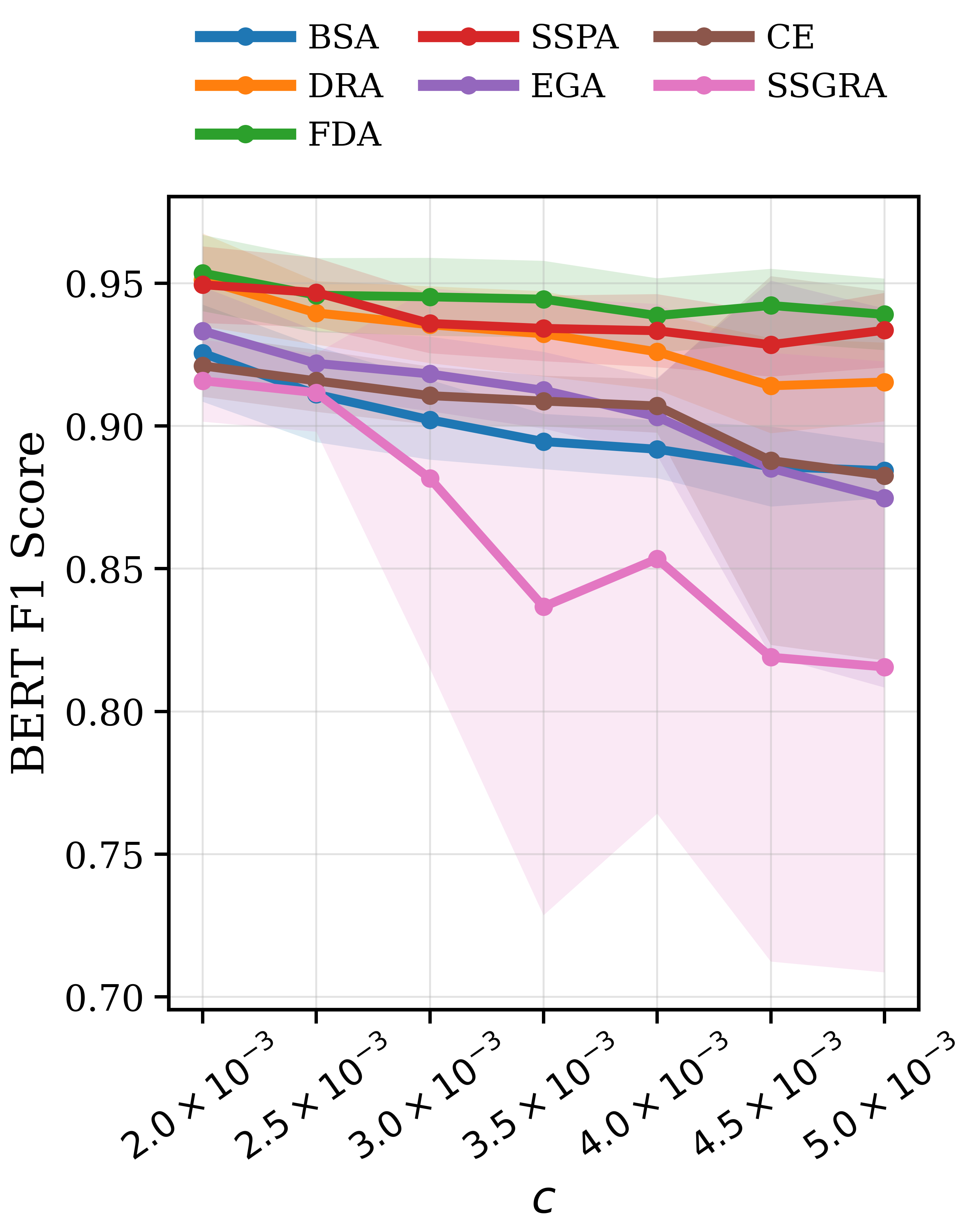}
    \caption{LLaVA~1.5 BERT F1}
\end{subfigure}
\hfill
\begin{subfigure}[b]{0.3\textwidth}
    \centering
    \includegraphics[width=\linewidth, trim=0cm 0cm 0cm 2.4cm, clip]{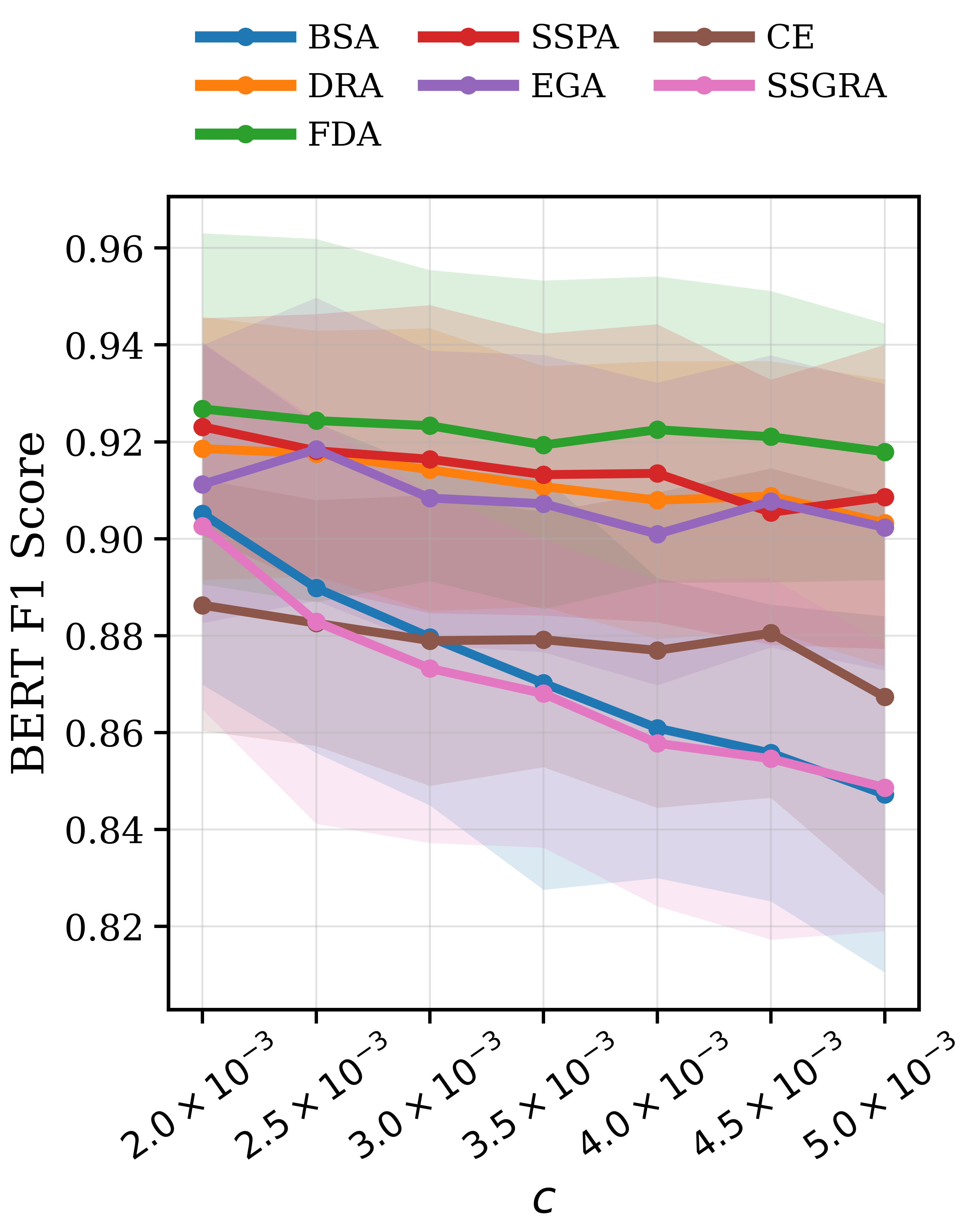}
    \caption{Gemma~3 BERT F1}
\end{subfigure}
\label{fig:bert_comparison}
\end{subfigure}
\vspace{0.8em}
\begin{subfigure}[t]{\textwidth}
\centering
\begin{subfigure}[b]{0.3\textwidth}
    \centering
    \includegraphics[width=\linewidth, trim=0cm 0cm 0cm 2.4cm, clip]{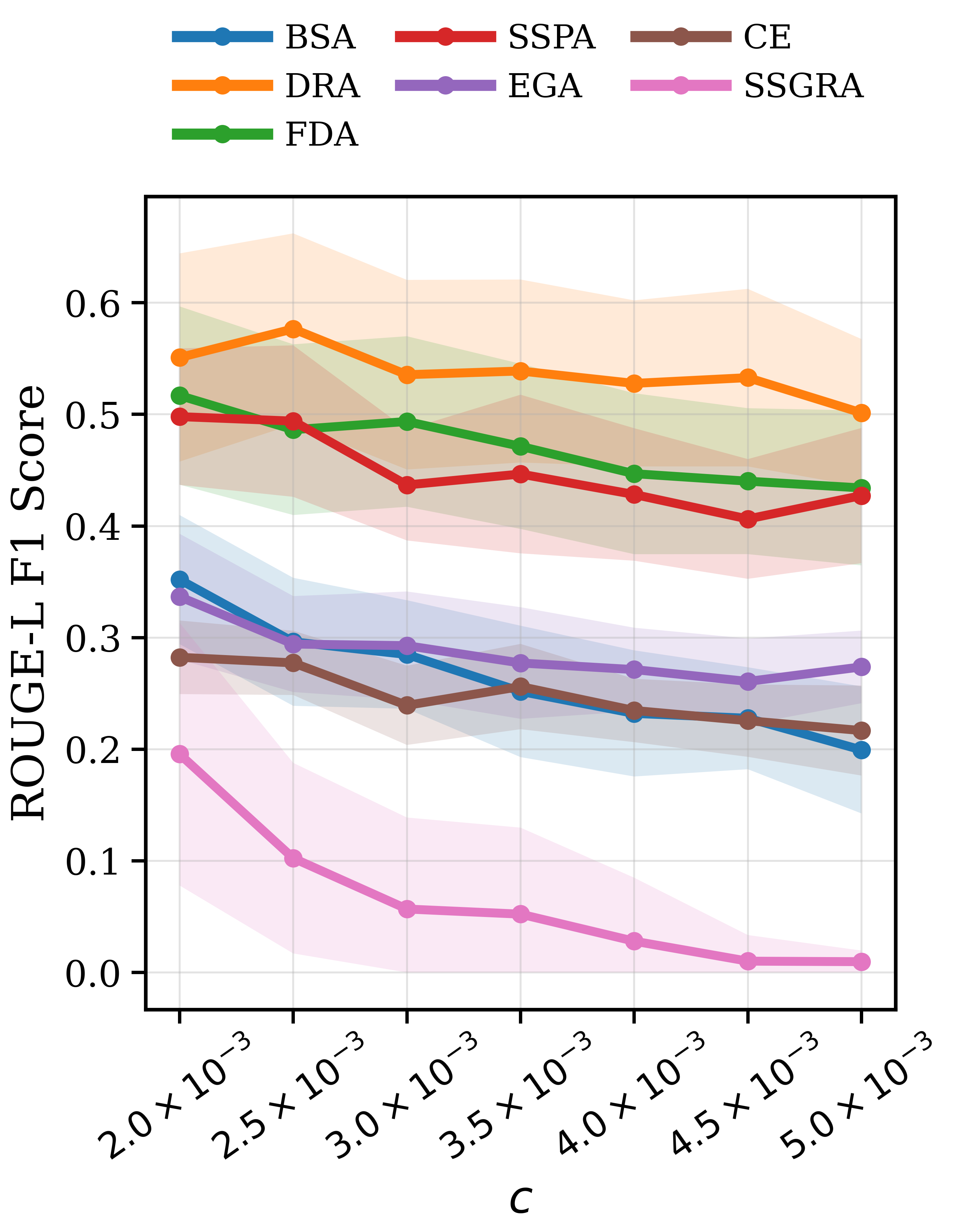}
    \caption{Qwen~2.5-VL ROUGE-L F1}
\end{subfigure}
\hfill
\begin{subfigure}[b]{0.3\textwidth}
    \centering
    \includegraphics[width=\linewidth, trim=0cm 0cm 0cm 2.4cm, clip]{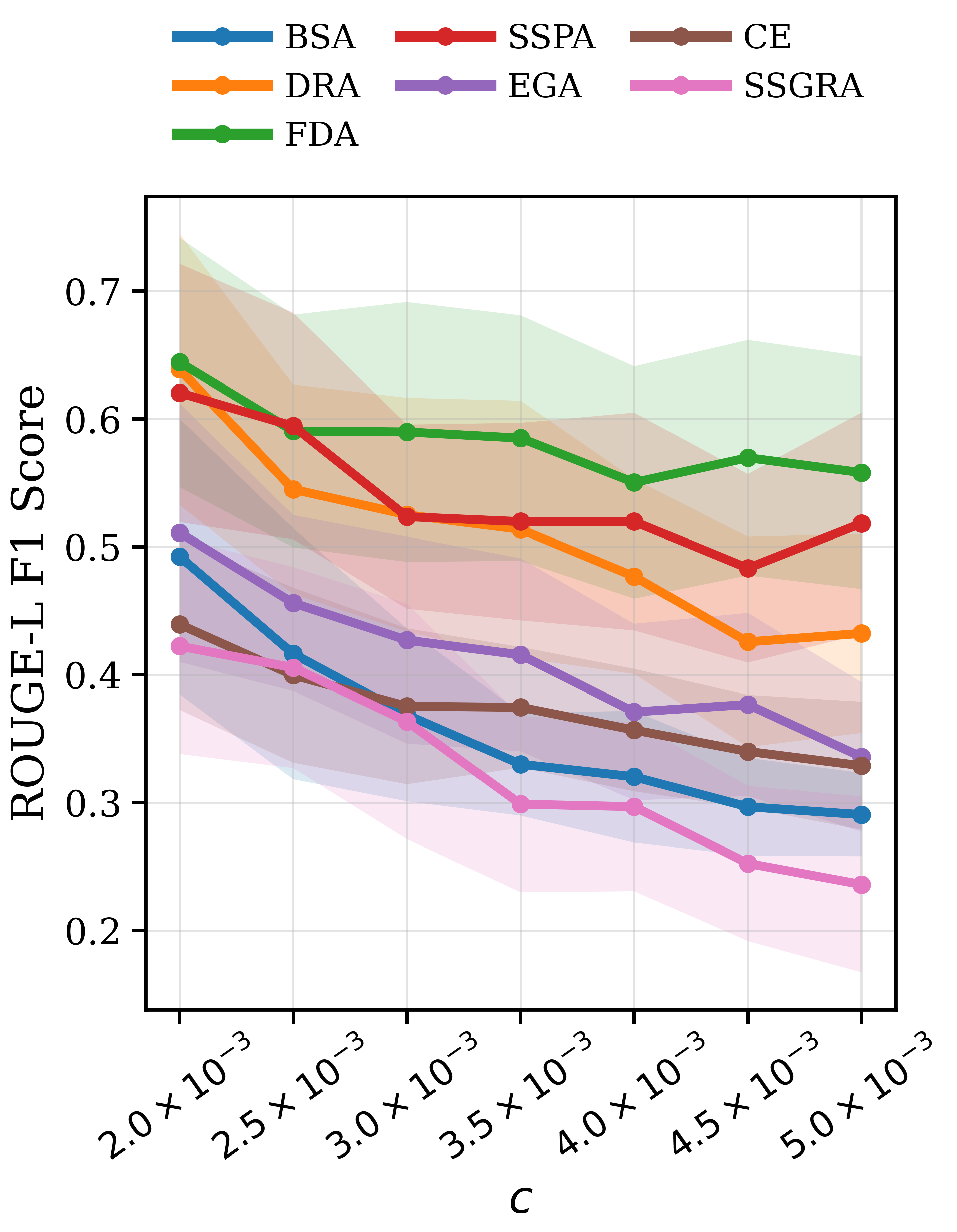}
    \caption{LLaVA~1.5 ROUGE-L F1}
\end{subfigure}
\hfill
\begin{subfigure}[b]{0.3\textwidth}
    \centering
    \includegraphics[width=\linewidth, trim=0cm 0cm 0cm 2.4cm, clip]{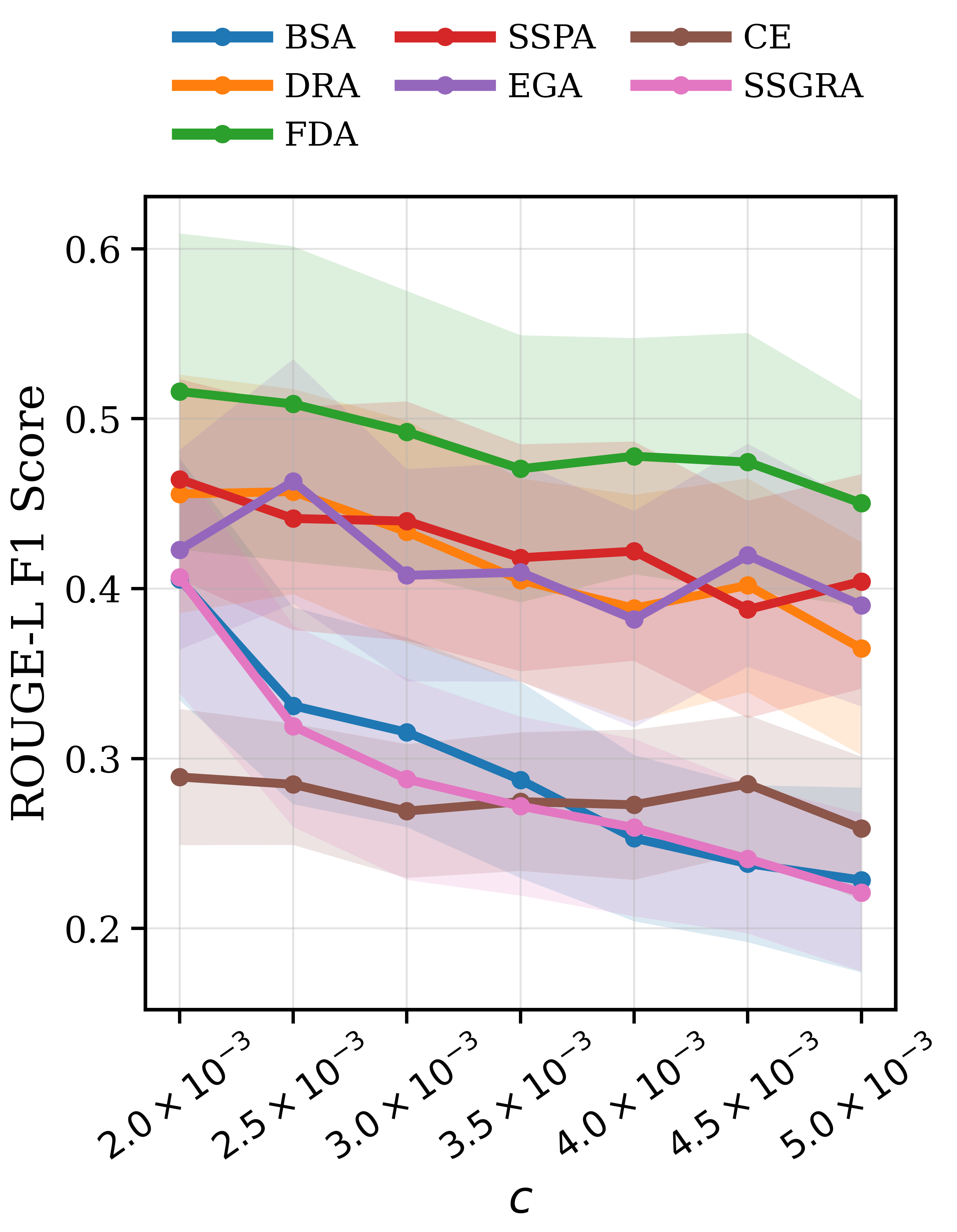}
    \caption{Gemma~3 ROUGE-L F1}
\end{subfigure}
\label{fig:rouge_comparison}
\end{subfigure}
\caption{SSGRA vs representative baselines across perturbation budgets. Lower scores indicate stronger attacks.
}
\label{fig:comparativeEvaluation}

\end{figure*}

Figure~\ref{fig:comparativeEvaluation} shows the performance of SSGRA and the selected baselines across perturbation budgets using BERTScore F1 and ROUGE-L F1.\\
\noindent\textbf{Qwen~2.5-VL:} SSGRA consistently outperforms all baselines, achieving an additional \textbf{7.90--19.74\%} relative degradation under BERTScore F1 and \textbf{30.50--97.93\%} under ROUGE-L F1 over the strongest baseline. \\
\noindent\textbf{LLaVA~1.5:} SSGRA achieves up to \textbf{7.46\%} and \textbf{18.90\%} additional relative degradation over the strongest baseline under BERTScore F1 and ROUGE-L F1, respectively. The improvement increases with the perturbation budget, indicating that spectral subspace guidance becomes more effective at larger perturbation budgets.\\
\noindent\textbf{Gemma~3: } Although the margins are smaller, SSGRA consistently achieves the strongest attacks, providing up to \textbf{0.80\%} and \textbf{3.07\%} additional relative degradation over the strongest baseline under BERTScore F1 and ROUGE-L F1, respectively. \\
Among the three VLMs, Qwen~2.5-VL exhibits the largest degradation, consistent with its having the highest proportion of near-null singular directions (Table~\ref{tab:singular_value_statistics_all_models}), and thus the largest spectral attack surface.
In contrast, Gemma~3 shows the smallest relative degradation, consistent with its lower proportion of near-null singular directions compared with Qwen~2.5-VL and LLaVA~1.5, limiting the opportunity to exploit bottom singular-vector subspaces.  A detailed discussion on spectral characterization follows in Section~\ref{sec:spectralCharacterization}.

Additional results are provided in the Appendix. 
In particular, Table~\ref{tab:all_results} summarizes the F1 scores and the additional relative degradation (\%) over the strongest baseline attack. The corresponding Precision, Recall, and F1 scores are provided in Tables~\ref{tab:Qwen_all_results}--\ref{tab:gemma_all_results}, while the corresponding trends are shown in Figures~\ref{fig:bertscore_comparison_all_models} and~\ref{fig:rougel_comparison_all_models} in the Appendix. 

Overall, the results suggest a relationship between spectral conditioning and adversarial vulnerability and provide empirical support for our hypothesis that bottom singular-vector subspaces play a central role in transformer-based VLMs.

\subsection{Qualitative Analysis}
\begin{figure}[t]
    \centering
    \begin{subfigure}{\linewidth}
        \centering
        \includegraphics[width=\linewidth]{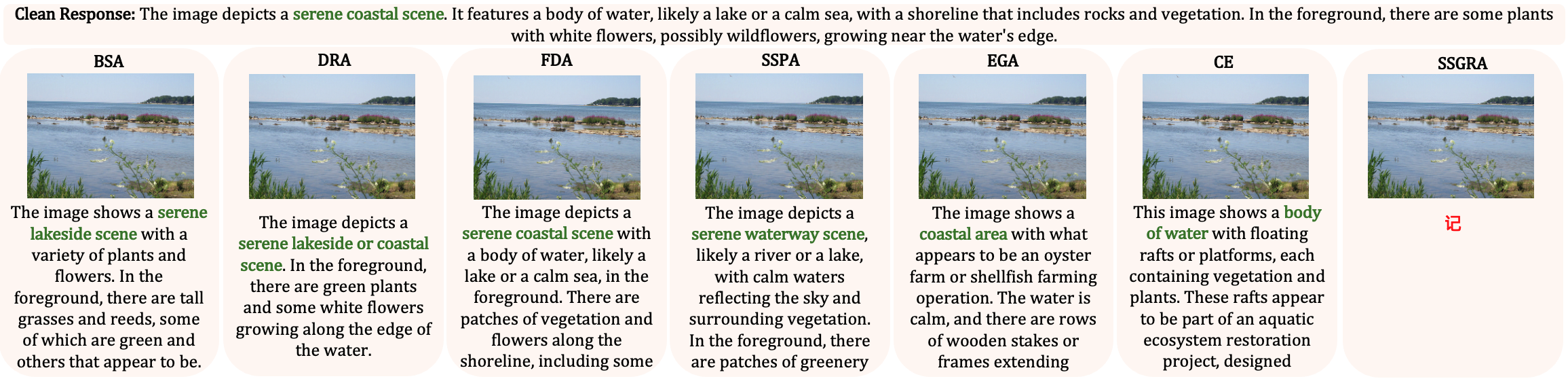}
        \caption{Qwen2.5-VL}
        \label{fig:qual_qwen}
    \end{subfigure}
    \begin{subfigure}{\linewidth}
        \centering
        \includegraphics[width=\linewidth]{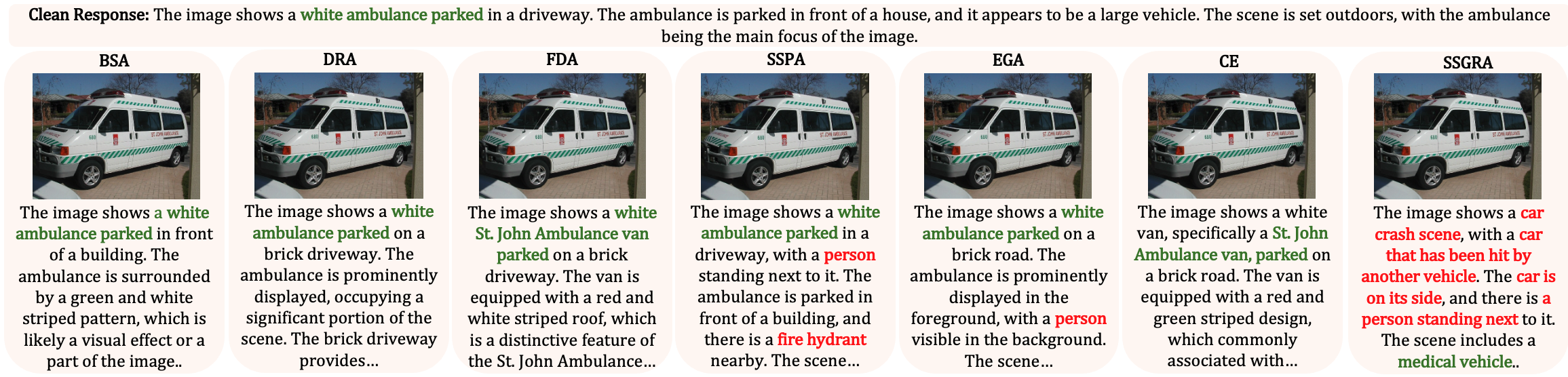}
        \caption{LLaVA-1.5}
        \label{fig:qual_llava}
    \end{subfigure}
    \begin{subfigure}{\linewidth}
        \centering
        \includegraphics[width=\linewidth]{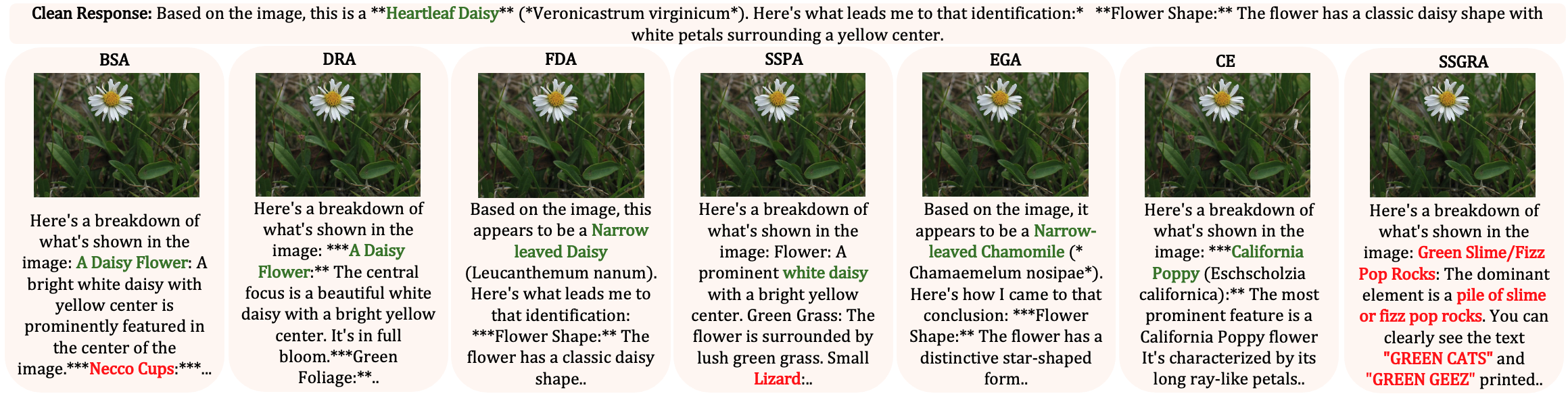}
        \caption{Gemma 3}
        \label{fig:qual_gemma}
    \end{subfigure}
    \caption{Qualitative adversarial examples generated under a common perturbation budget of $c=0.003$ across three vision--language models.}
    \label{fig:qualitative_eps003}
\end{figure}

Figure~\ref{fig:qualitative_eps003} shows representative adversarial examples generated under a common extremely small perturbation budget of $c=0.003$ for the chosen models. Compared with existing baselines, \textbf{SSGRA} consistently induces larger semantic deviations while preserving the perceptual appearance of the input images.

For the coastal scene (Qwen2.5-VL), baseline attacks largely preserve the correct scene description, whereas SSGRA causes the model response to become meaningless. Likewise, for the ambulance image (LLaVA-1.5), baseline attacks still identify the ambulance despite minor hallucinations, whereas SSGRA instead describes an unrelated car crash scene. For the flower image (Gemma~3), baseline attacks largely preserve the correct flower category despite minor hallucinations involving a cup and a lizard, whereas SSGRA generates an unrelated prediction ("Green Slime/Fizz Pop Rocks"). More qualitative examples are presented in Appendix~\ref{sec:additionalQualitative}.

Overall, SSGRA induces larger semantic shifts than existing baselines, consistent with the quantitative findings in Section~\ref{sec:quantitativeComparision}.

\subsection{Spectral Characterization of Adversarial Representations}
\label{sec:spectralCharacterization}
\begin{figure*}[t]
\centering

\begin{subfigure}[t]{0.32\textwidth}
    \centering
    \includegraphics[width=0.48\linewidth]{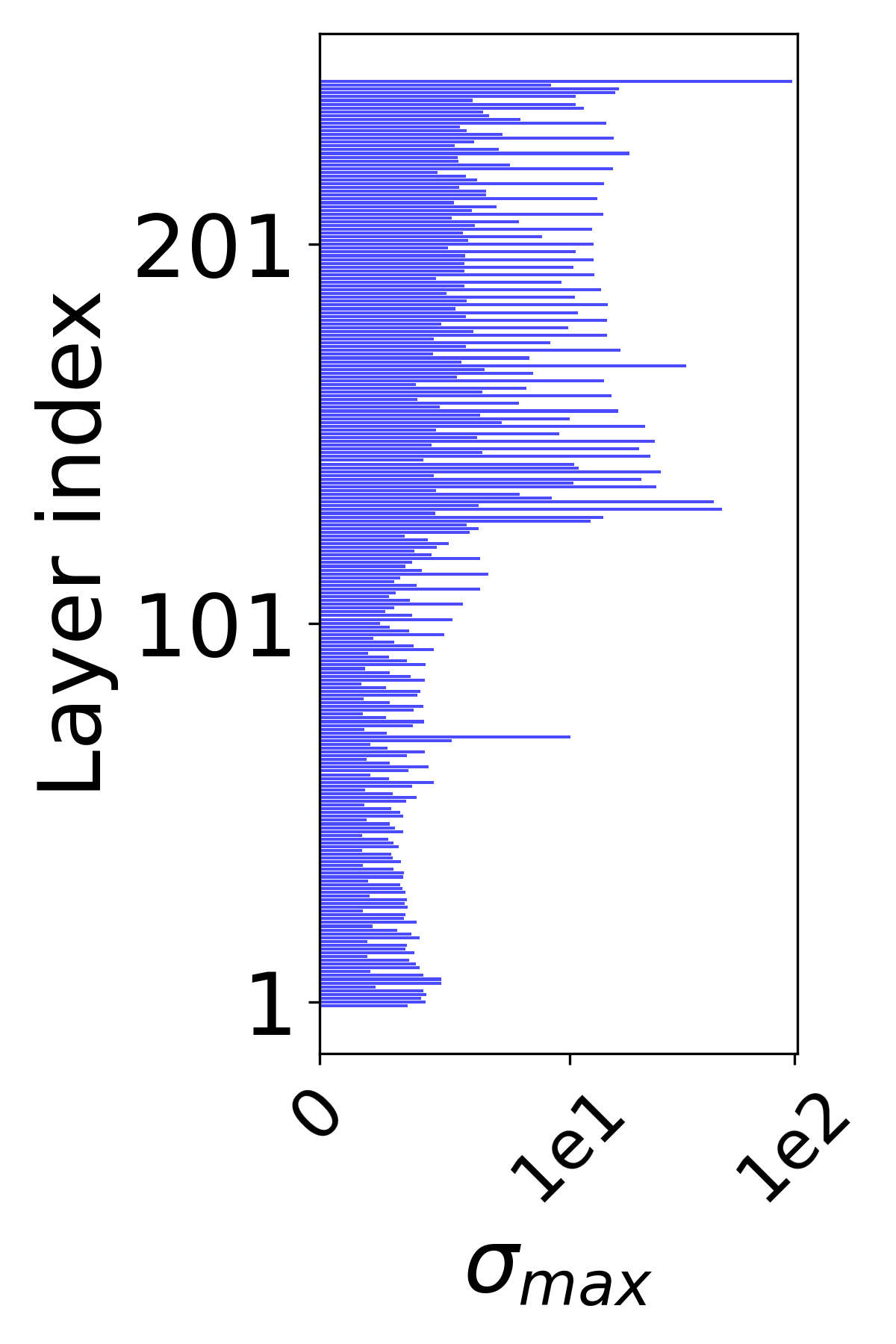}
    \hfill
    \includegraphics[width=0.48\linewidth]{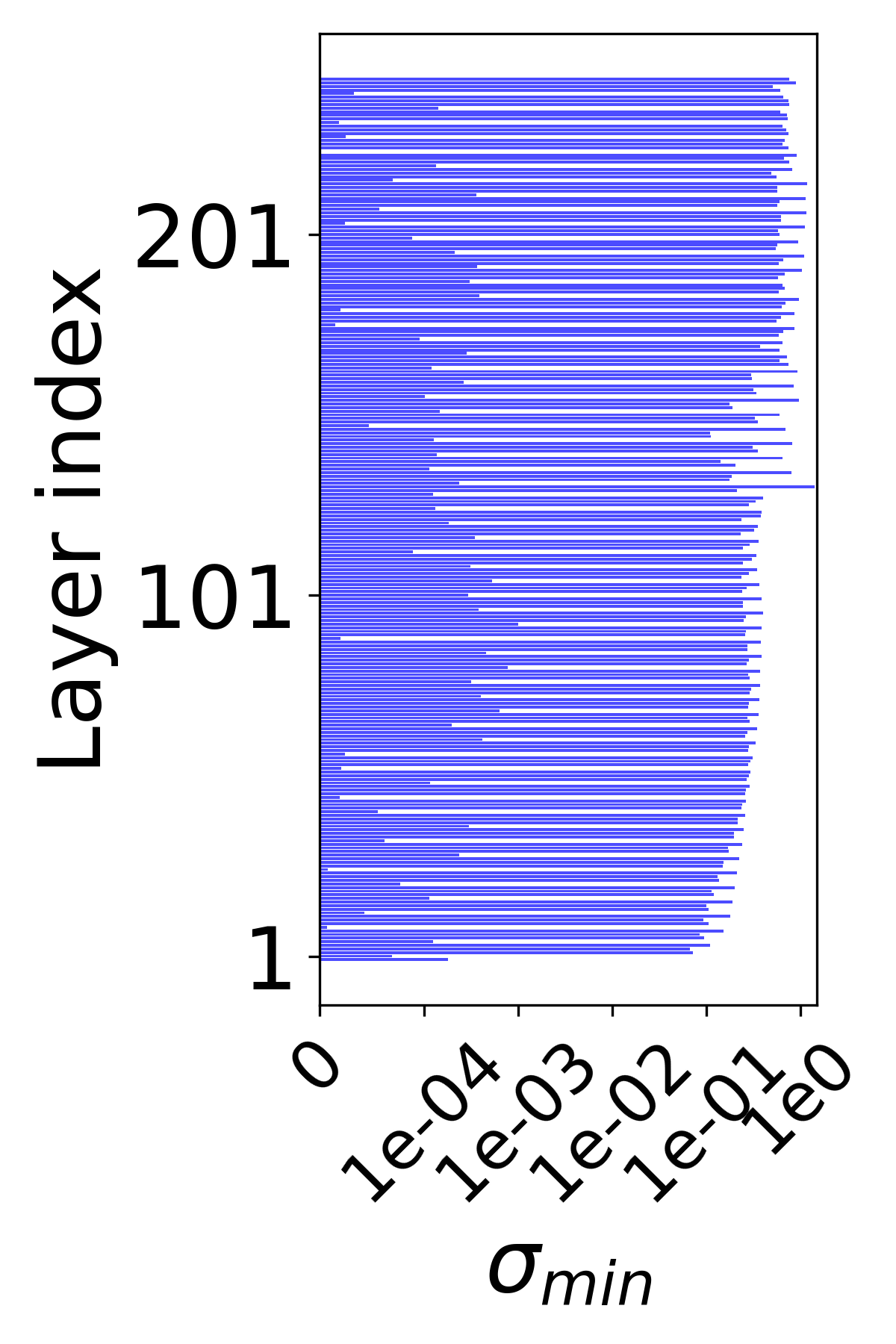}
    \caption{Qwen2.5-VL}
\end{subfigure}
\hfill
\begin{subfigure}[t]{0.32\textwidth}
    \centering
    \includegraphics[width=0.48\linewidth]{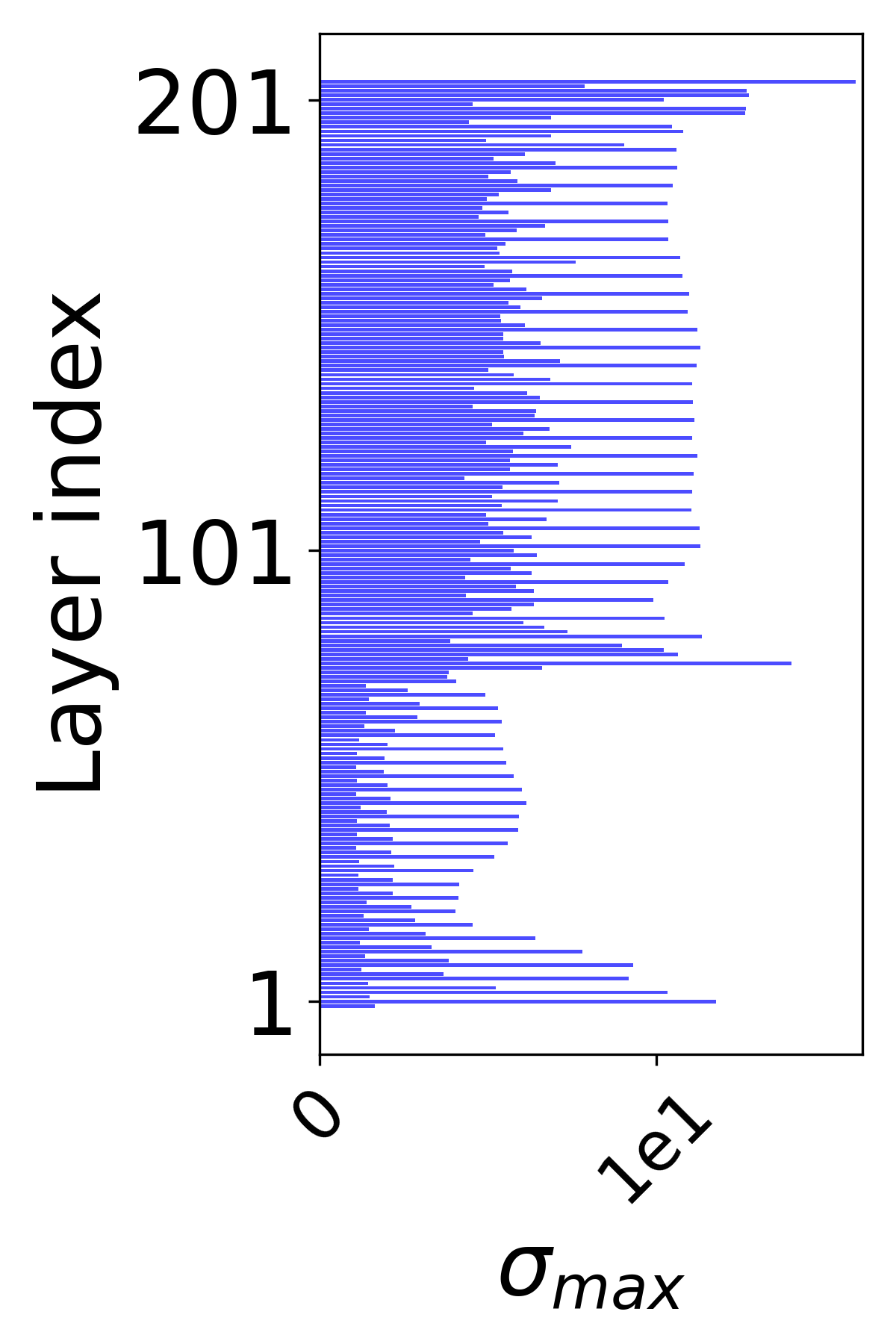}
    \hfill
    \includegraphics[width=0.48\linewidth]{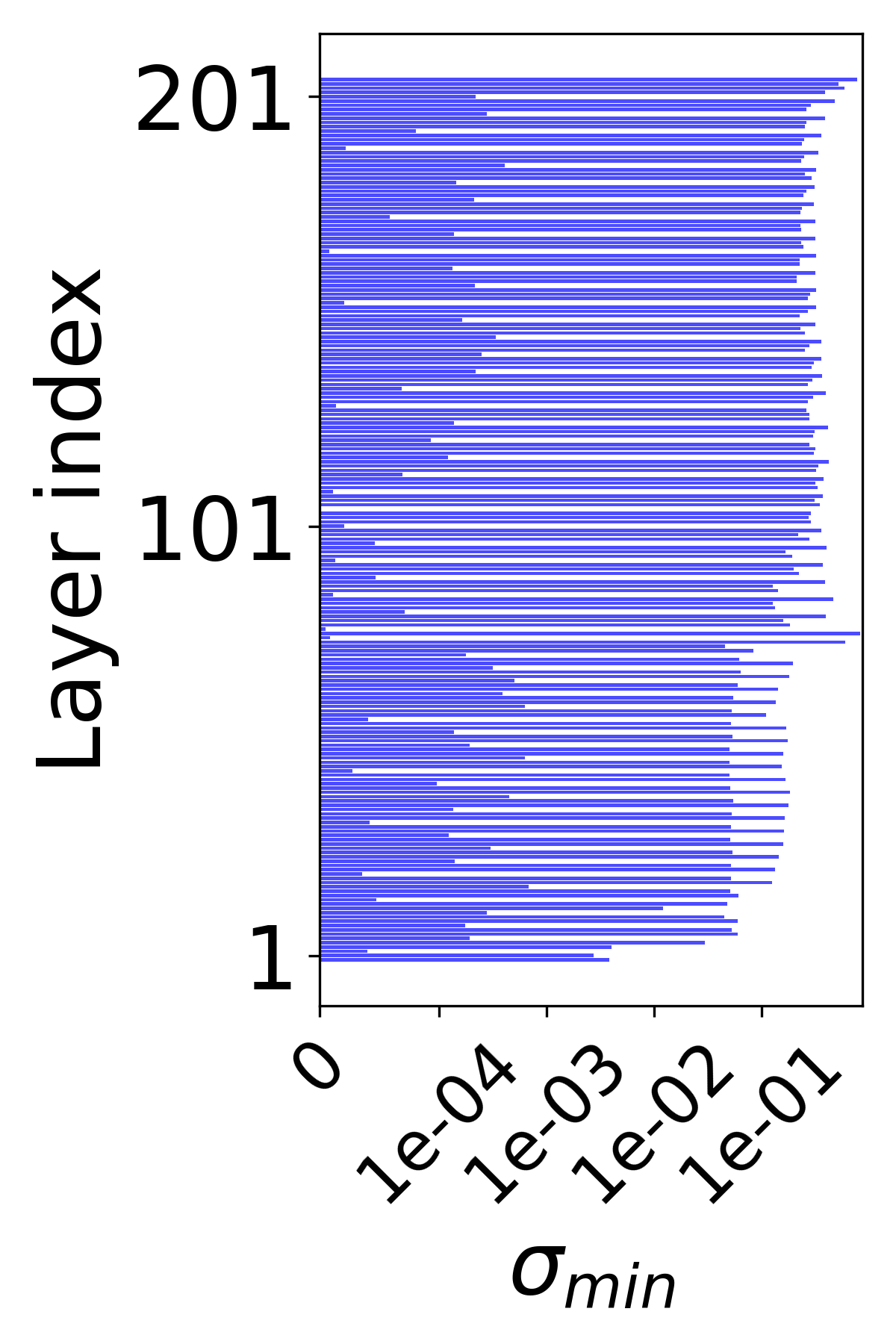}
    \caption{LLaVA-1.5}
\end{subfigure}
\hfill
\begin{subfigure}[t]{0.32\textwidth}
    \centering
    \includegraphics[width=0.48\linewidth]{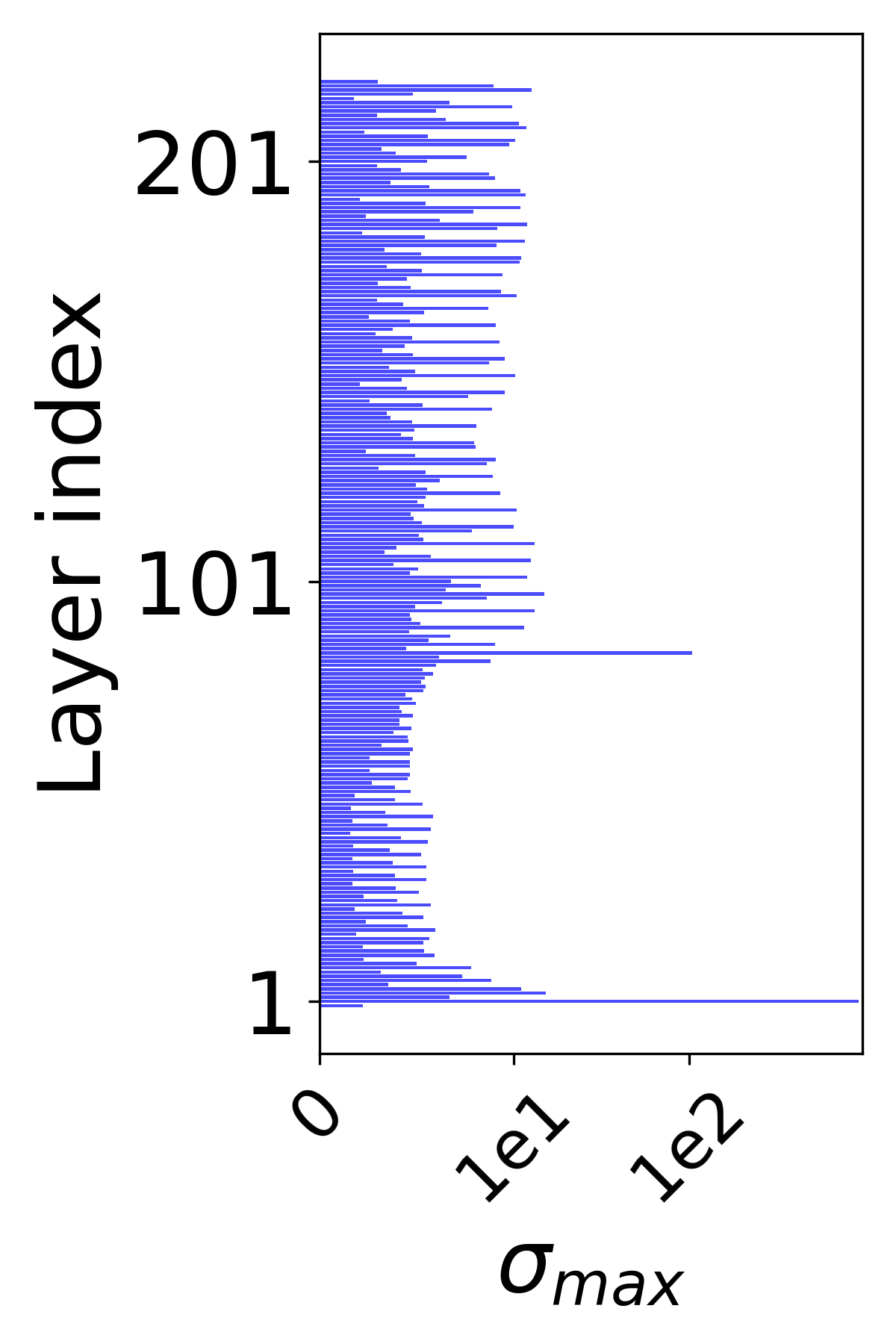}
    \hfill
    \includegraphics[width=0.48\linewidth]{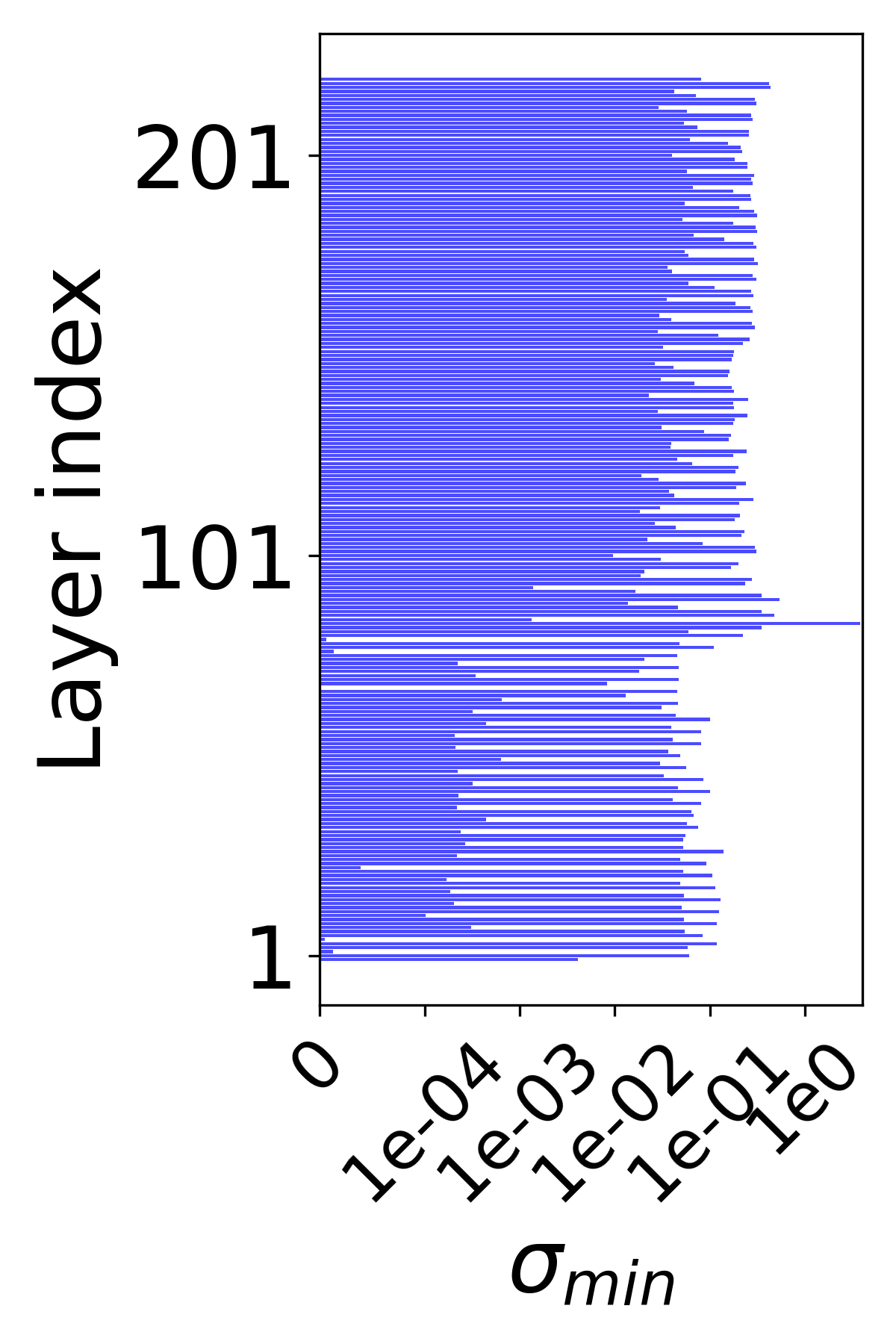}
    \caption{Gemma~3}
\end{subfigure}

\caption{Distribution of the largest (\(\sigma_{\max}\), left) and smallest (\(\sigma_{\min}\), right) singular values across non-attention linear operators in Qwen2.5-VL, LLaVA-1.5, and Gemma~3.}
\label{fig:singular_value_plots}
\end{figure*}

\begin{table*}[t]
\centering
\caption{Distribution of extreme singular values across non-attention linear operators.}
\label{tab:singular_value_statistics_all_models}
\scriptsize
\setlength{\tabcolsep}{4pt}
\begin{tabular}{lcccccc}
\toprule
\multirow{2}{*}{\textbf{Threshold}}
& \multicolumn{2}{c}{\textbf{Qwen2.5-VL}}
& \multicolumn{2}{c}{\textbf{LLaVA-1.5}}
& \multicolumn{2}{c}{\textbf{Gemma~3}} \\
\cmidrule(lr){2-3}
\cmidrule(lr){4-5}
\cmidrule(lr){6-7}
& Count & Percentage
& Count & Percentage
& Count & Percentage \\
\midrule

\multicolumn{7}{c}{\textbf{Largest Singular Value (\(\sigma_{\max}\))}} \\
\midrule
\(\sigma_{\max} > 10^{1}\)
& 46/245 & 18.78\%
& 40/206 & 19.42\%
& 25/221 & 11.31\% \\

\(\sigma_{\max} > 10^{2}\)
& 0/245 & 0.00\%
& 0/206 & 0.00\%
& 2/221 & 0.90\% \\

\midrule
\multicolumn{7}{c}{\textbf{Smallest Singular Value (\(\sigma_{\min}\))}} \\
\midrule
\(\sigma_{\min} < 10^{-2}\)
& 62/245 & 25.31\%
& 60/206 & 29.13\%
& 32/221 & 14.48\% \\

\(\sigma_{\min} < 10^{-3}\)
& 61/245 & 24.90\%
& 57/206 & 27.67\%
& 27/221 & 12.22\% \\

\(\sigma_{\min} < 10^{-4}\)
& 24/245 & 9.80\%
& 26/206 & 12.62\%
& 6/221 & 2.71\% \\

\(\sigma_{\min} < 10^{-5}\)
& 3/245 & 1.22\%
& 4/206 & 1.94\%
& 3/221 & 1.36\% \\

\(\sigma_{\min} < 10^{-6}\)
& 0/245 & 0.00\%
& 1/206 & 0.49\%
& 0/221 & 0.00\% \\

\bottomrule
\end{tabular}
\end{table*}
\noindent\textbf{Singular Value Analysis of intermediate linear operators of VLMs. }We analyze the distributions of the largest and smallest singular values of all non-attention linear operators in the evaluated models. Figure~\ref{fig:singular_value_plots} visualizes these distributions, while Table~\ref{tab:singular_value_statistics_all_models} summarizes the prevalence of extreme singular values. Across all three models, near-null singular directions occur substantially more frequently than strongly amplifying ones. For example, $\sigma_{\min}<10^{-3}$ is observed in $\textbf{24.90\%}$, $\textbf{27.67\%}$, and $\textbf{12.22\%}$ of operators in Qwen2.5-VL, LLaVA-1.5, and Gemma~3, respectively, with similar trends persisting at smaller thresholds. In contrast, singular values satisfying $\sigma_{\max}>10$ occur much less frequently, accounting for only $\textbf{18.78\%}$, $\textbf{19.42\%}$, and $\textbf{11.31\%}$ of operators. Singular values exceeding $10^2$ are nearly absent across all models (Table~\ref{tab:singular_value_statistics_all_models}), consistent with prior work showing that large singular values are commonly constrained or implicitly regularized during training.

These spectral characteristics help explain the quantitative results in Section~\ref{sec:quantitativeComparision}. Qwen2.5-VL and LLaVA-1.5 contain approximately twice as many near-null singular directions as Gemma~3, providing substantially larger bottom singular-vector subspaces that SSGRA can exploit. Consequently, these models exhibit much larger relative degradation over existing attacks than Gemma 3.

\begin{figure}[t]
\centering
\begin{subfigure}[t]{0.45\linewidth}
    \centering
    \includegraphics[width=\linewidth, trim=0cm 1.1cm 0cm 0cm, clip]{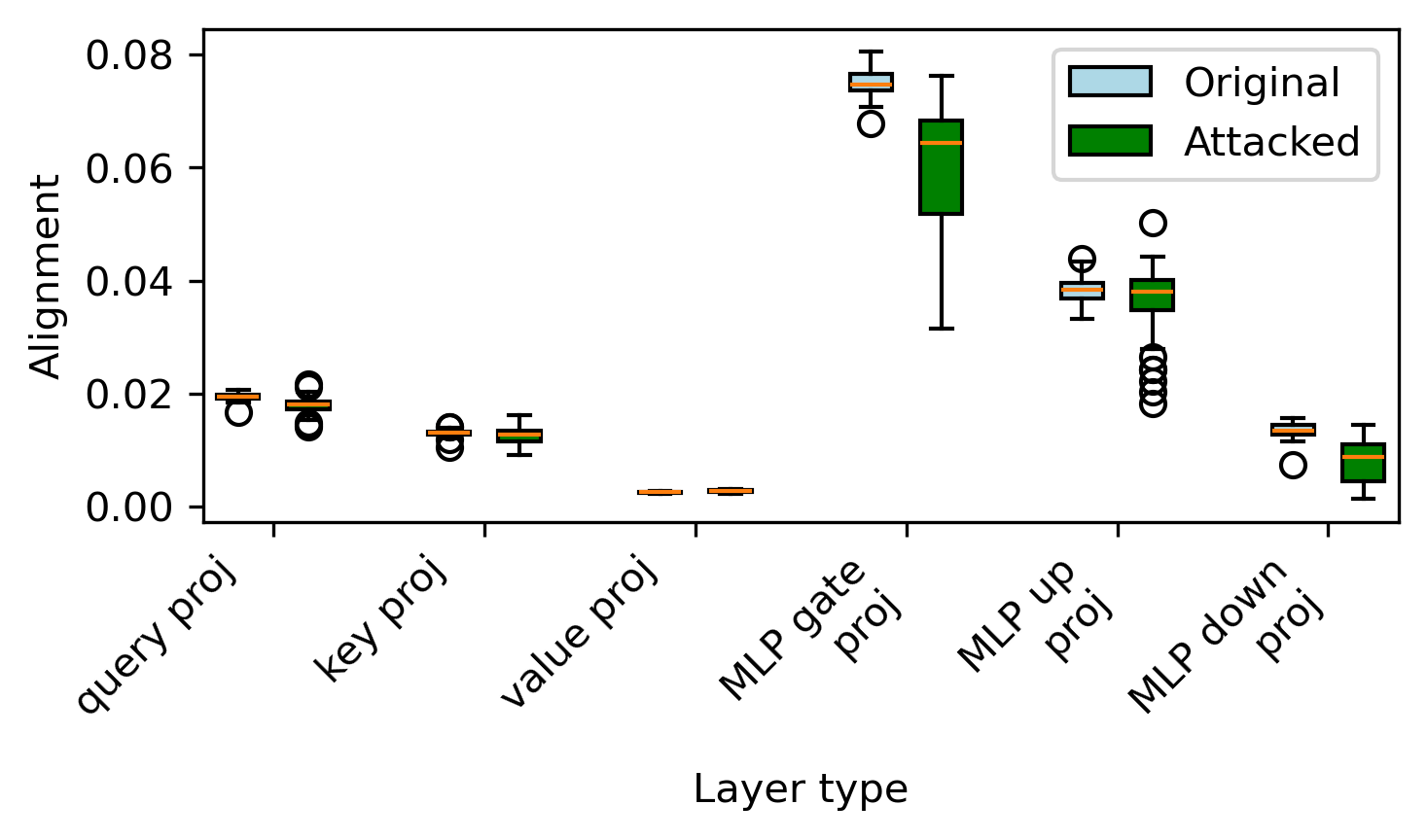}
    \caption{BSA, Top k = 10}
    \label{fig:top_subspace_alignment}
\end{subfigure}
\begin{subfigure}[t]{0.45\linewidth}
    \centering
    \includegraphics[width=\linewidth, trim=0cm 1.1cm 0cm 0cm, clip]{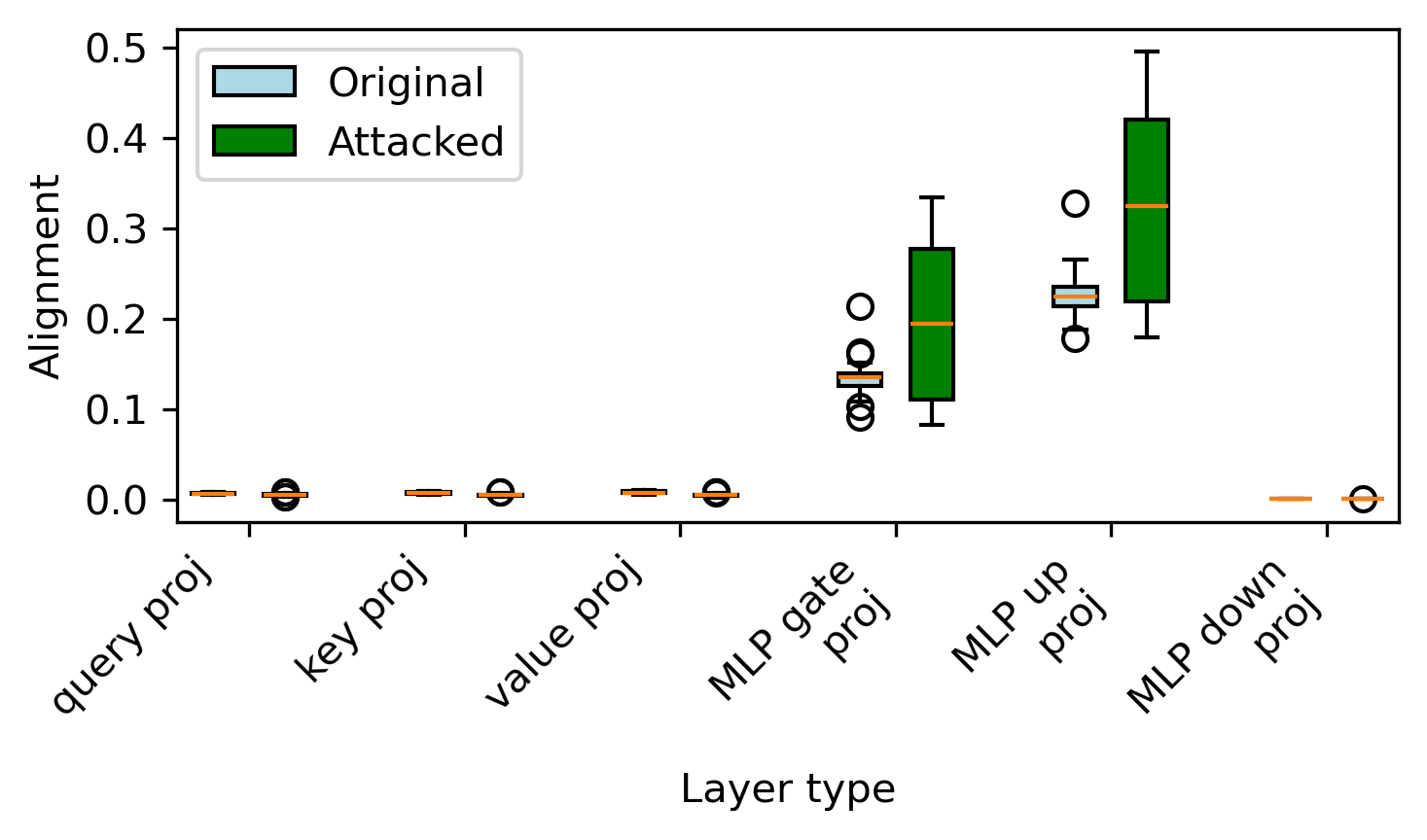}
    \caption{BSA, Bottom k = 10}
    \label{fig:bottom_subspace_alignment}
\end{subfigure}
\caption{Distribution of the spectral alignment measure $\Psi_k$ before and after BSA optimization for the top-10 and bottom-10 right singular-vector subspaces.}
\label{fig:spectral_alignment_language_layer3}
\end{figure}

\noindent\textbf{Post-Attack Spectral Alignment Analysis. }
To examine whether spectral alignment emerges naturally during adversarial optimization, we analyze adversarial representations generated by BSA~\cite{yin2023vlattack}, which does not optimize spectral objectives. We evaluate 100 images ($c=0.005$) and compute $\Psi_k$ using Eq.~\eqref{alignmentMeasure} with respect to the top-10 and bottom-10 right singular vectors of selected intermediate layers. Figure~\ref{fig:spectral_alignment_language_layer3} shows the corresponding alignment distributions before and after optimization.


At the \emph{MLP gate proj} and \emph{MLP up proj} layers, adversarial representations exhibit increased alignment with the bottom-$k$ singular subspace, whereas alignment with the top-$k$ subspace remains unchanged or decreases. This trend is absent in attention layers, likely due to their more complex transformations. Since BSA does not optimize spectral alignment, the results suggest that untargeted adversarial optimization naturally steers representations toward information-attenuating bottom singular subspaces. SSGRA strengthens the attack by explicitly promoting this alignment.

\begin{figure}[t]
\centering
\begin{subfigure}[b]{0.24\linewidth}
    \centering
    \includegraphics[width=\linewidth, trim=7.9cm 1.8cm 5.1cm 3.1cm, clip]{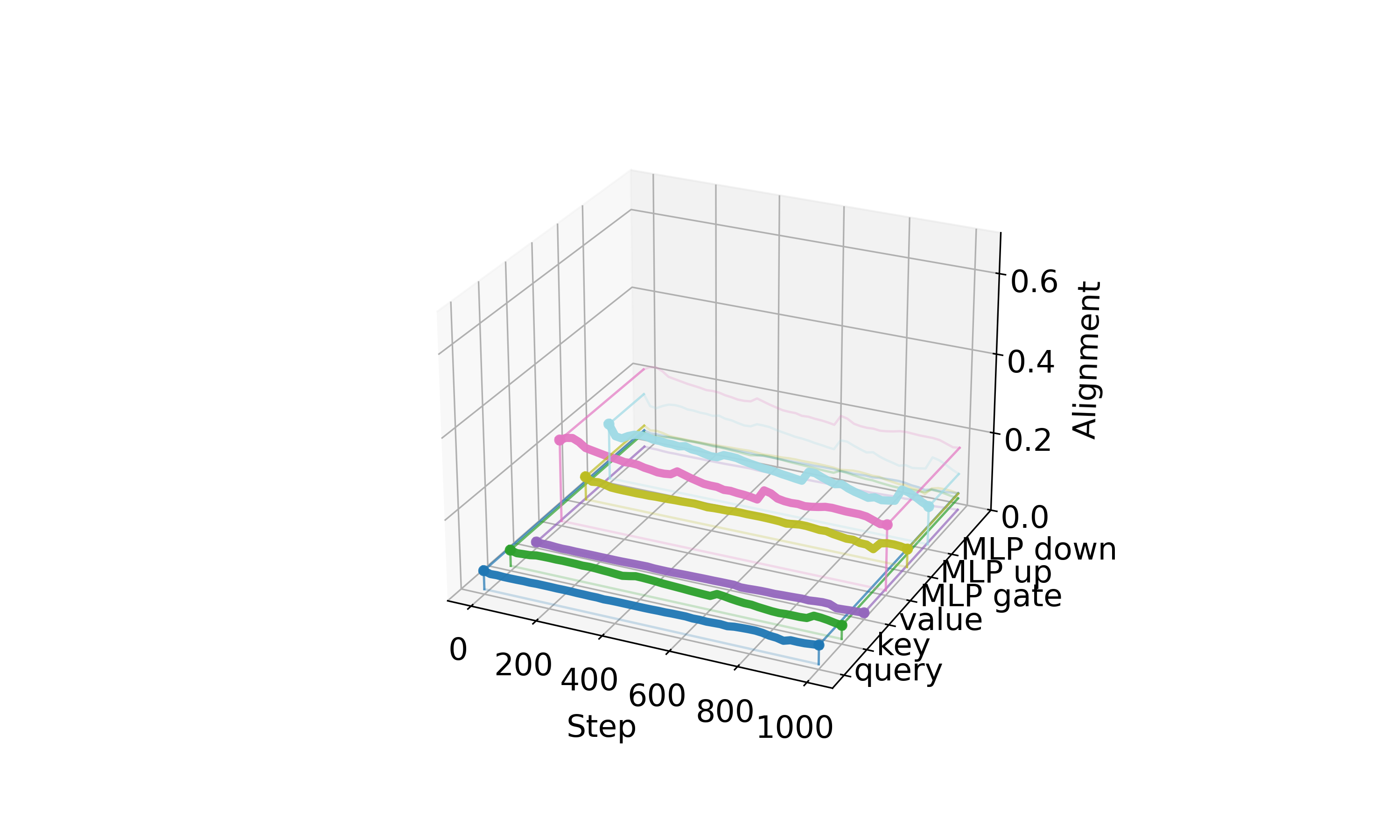}
    \caption{$\Psi_{\mathrm{top}}(v)$\\ for sample 1}
\end{subfigure}
\begin{subfigure}[b]{0.24\linewidth}
    \centering
    \includegraphics[width=\linewidth, trim=7.9cm 1.8cm 5.1cm 3.1cm, clip]{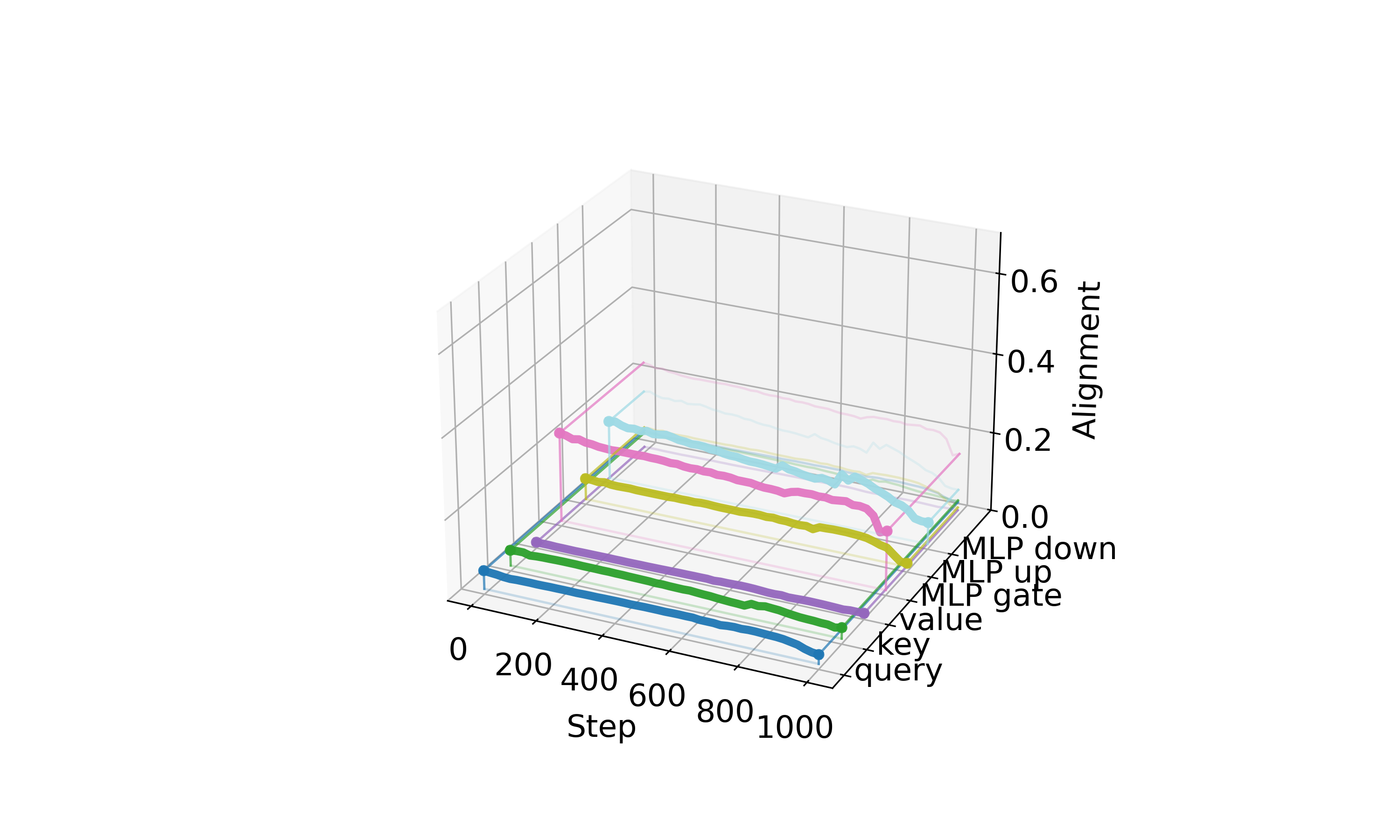}
    \caption{$\Psi_{\mathrm{top}}(v)$\\ for sample 2}
\end{subfigure}
\begin{subfigure}[b]{0.24\linewidth}
    \centering
    \includegraphics[width=\linewidth, trim=7.9cm 1.8cm 5.1cm 3.1cm, clip]  {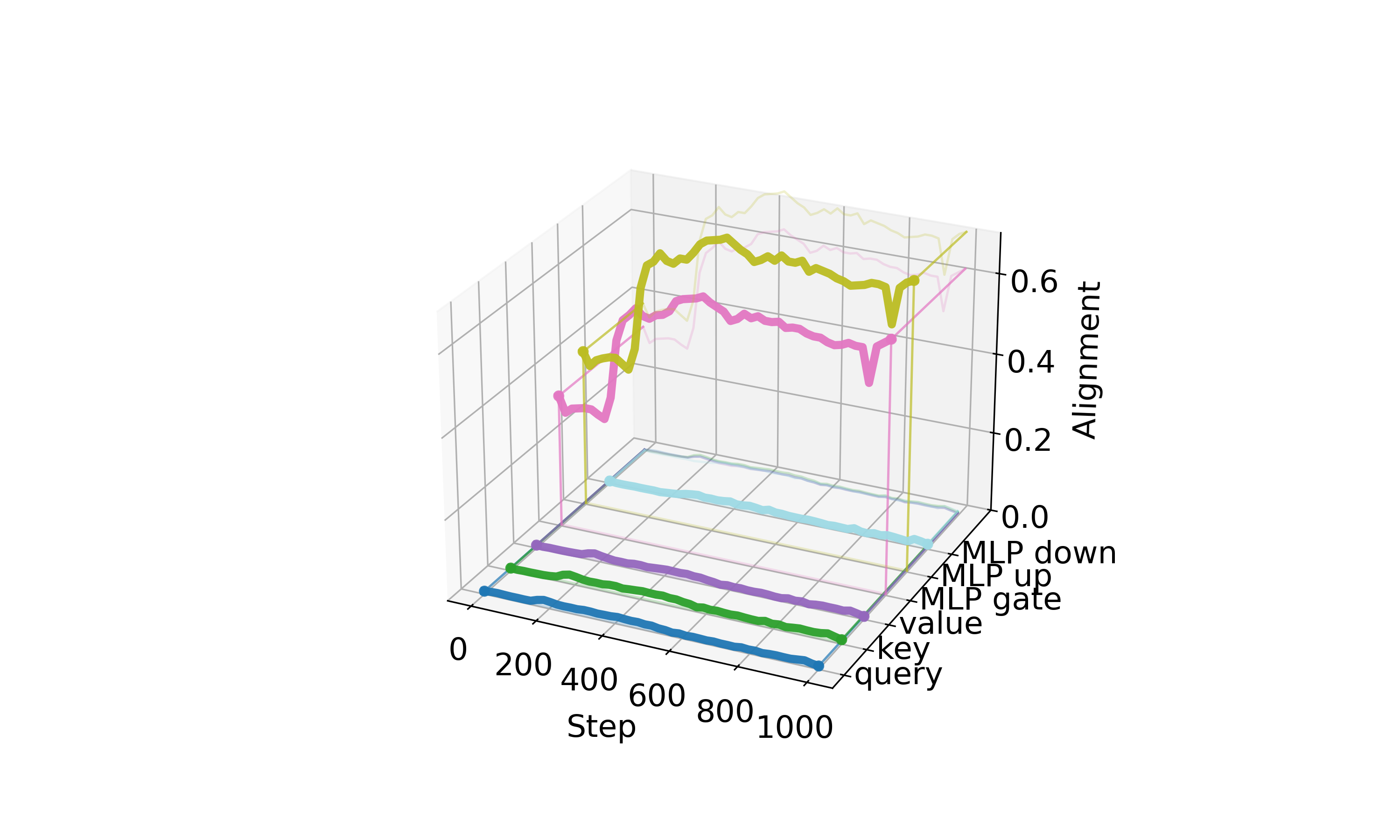}
    \caption{$\Psi_{\mathrm{bottom}}(v)$\\ for sample 1}
\end{subfigure}
\begin{subfigure}[b]{0.24\linewidth}
    \centering
    \includegraphics[width=\linewidth, trim=7.9cm 1.8cm 5.1cm 3.1cm, clip] {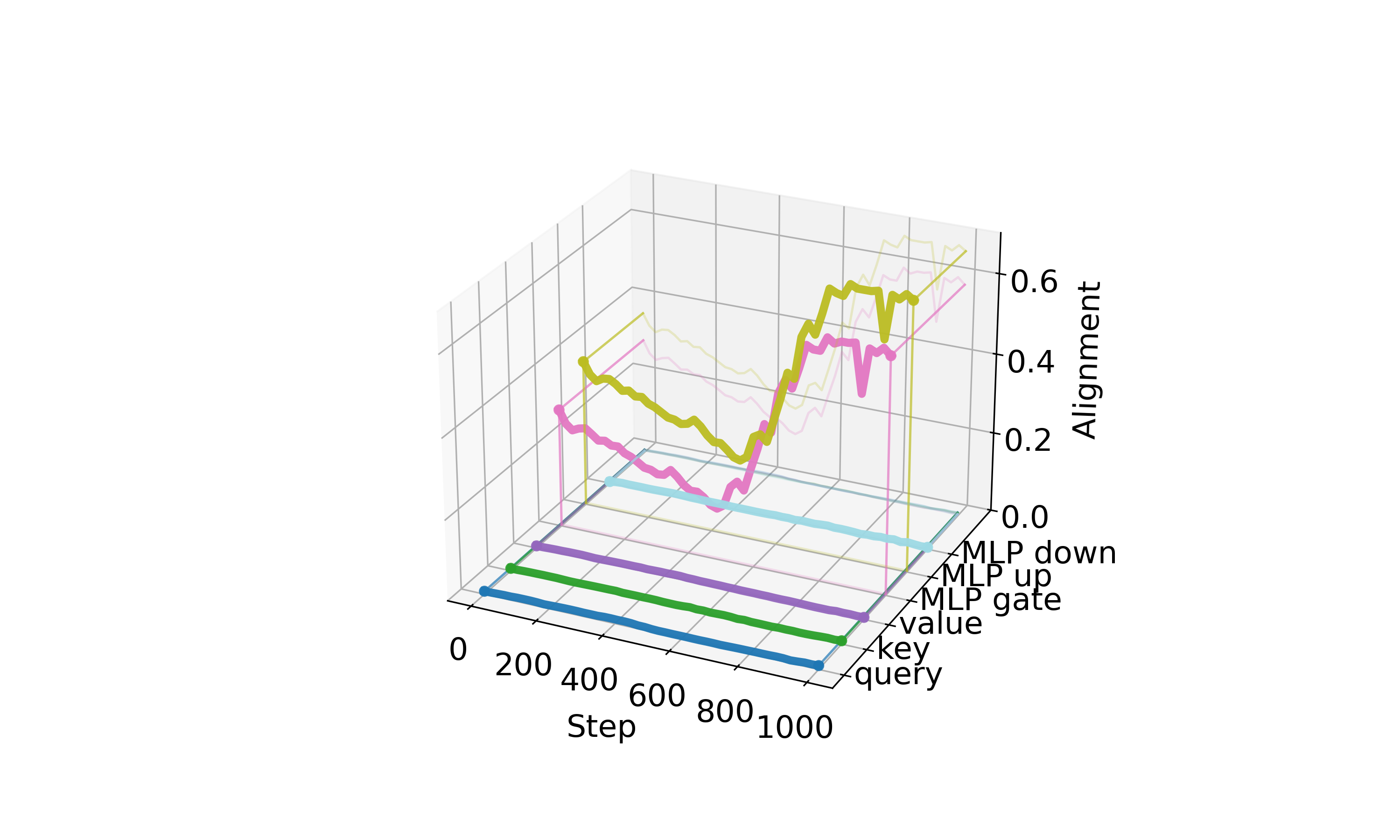}
    \caption{$\Psi_{\mathrm{bottom}}(v)$\\ for sample 2}
\end{subfigure}
\caption{Layer-wise evolution of top and bottom most singular vectors alignment during adversarial optimization on Gemma~3 (two representative samples).}
\label{fig.Eg1topBottomSing}
\end{figure}
\noindent\textbf{Spectral Alignment Dynamics During Attack Optimization. }We analyze spectral alignment during adversarial optimization on Gemma~3. For each target layer, adversarial examples are generated using Eq.~\eqref{eq:layerwise_probe}, and alignment with the largest and smallest singular vectors is computed using Eq.~\eqref{alignmentMeasure}. Figure~\ref{fig.Eg1topBottomSing} shows trajectories from the first multimodal block. Adversarial optimization progressively increases alignment with the bottom singular vectors, while alignment with the top singular vectors remains nearly unchanged or slightly decreases, supporting our hypothesis.

\subsection{Computational Complexity}
Table~\ref{tab:flops} reports the computational complexity of the evaluated attacks in FLOPs. Across all three VLMs, SSGRA has computational cost comparable to representative optimization-based attacks (BSA, EGA, and CE) while achieving improved attack effectiveness. FLOPs should be compared only within the same VLM, as they depend not only on parameter count but also on the vision encoder,  attention mechanism, input image resolution, and the number of visual tokens processed by the multimodal model.

For Qwen2.5-VL, SSGRA requires fewer FLOPs than all other methods while achieving the strongest attack performance (Section~\ref{sec:quantitativeComparision}), since hyperparameter tuning selected $\lambda=0$, reducing Eq.~\eqref{eq:ssgra} to just spectral alignment objective.

\begin{table*}[t]
\centering
\caption{Computational complexity of the attacks measured in floating-point operations (FLOPs). Lower values indicate higher computational efficiency.}
\label{tab:flops}
\scriptsize
\setlength{\tabcolsep}{5pt}
\begin{tabular}{lccc}
\toprule
\textbf{Method} &
\textbf{Qwen2.5-VL} &
\textbf{LLaVA-1.5} &
\textbf{Gemma~3} \\
\midrule

BSA   & $2.94\times10^{13}$ & $1.02\times10^{13}$ & $2.59\times10^{13}$ \\
DRA   & $1.42\times10^{13}$ & $8.68\times10^{12}$ & $1.01\times10^{13}$ \\
FDA   & $6.14\times10^{12}$ & $1.14\times10^{12}$ & $1.01\times10^{13}$ \\
SSPA   & $6.14\times10^{12}$ & $1.14\times10^{12}$ & $1.56\times10^{13}$ \\
EGA   & $2.68\times10^{13}$ & $2.69\times10^{13}$ & $1.95\times10^{13}$ \\
CE   & $2.69\times10^{13}$ & $2.69\times10^{13}$ & $1.96\times10^{13}$ \\
SSGRA & $\mathbf{1.31\times10^{13}}$ & $\mathbf{2.61\times10^{13}}$ & $\mathbf{2.60\times10^{13}}$ \\
\bottomrule
\end{tabular}
\end{table*}

\subsection{Ablation Study}

\begin{figure*}[t]

\includegraphics[width=0.45\textwidth]{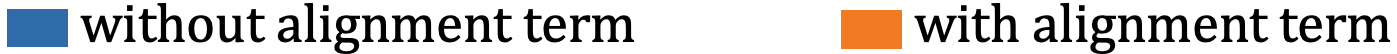}
\vspace{0.5mm}

\begin{subfigure}[t]{0.48\textwidth}
    \centering
    \begin{subfigure}[b]{0.49\linewidth}
        \centering
        \includegraphics[width=\linewidth,
        trim=0cm 0cm 0cm 1.7cm,clip]
        {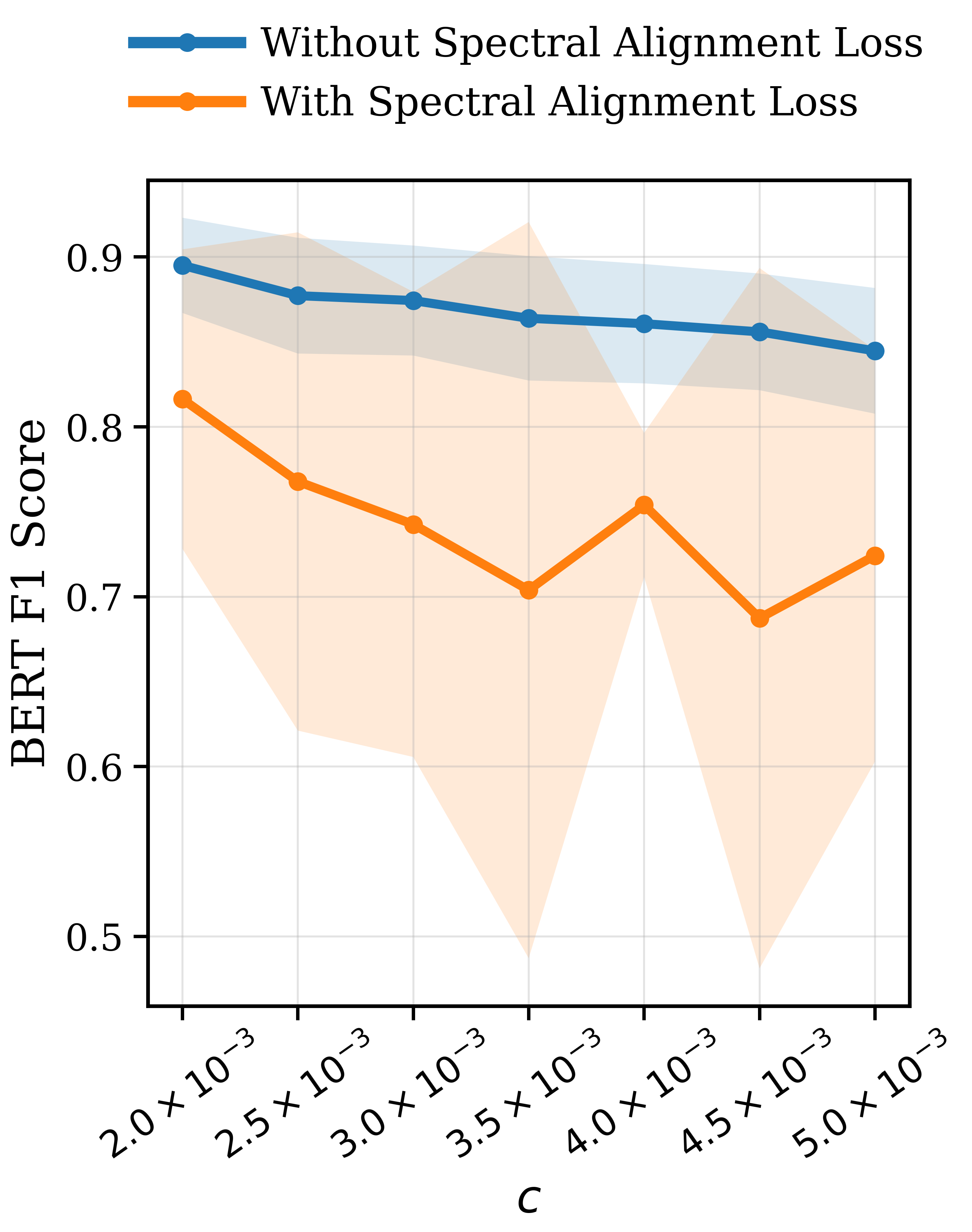}
    \end{subfigure}
    \hfill
    \begin{subfigure}[b]{0.49\linewidth}
        \centering
        \includegraphics[width=\linewidth,
        trim=0cm 0cm 0cm 1.7cm,clip]
        {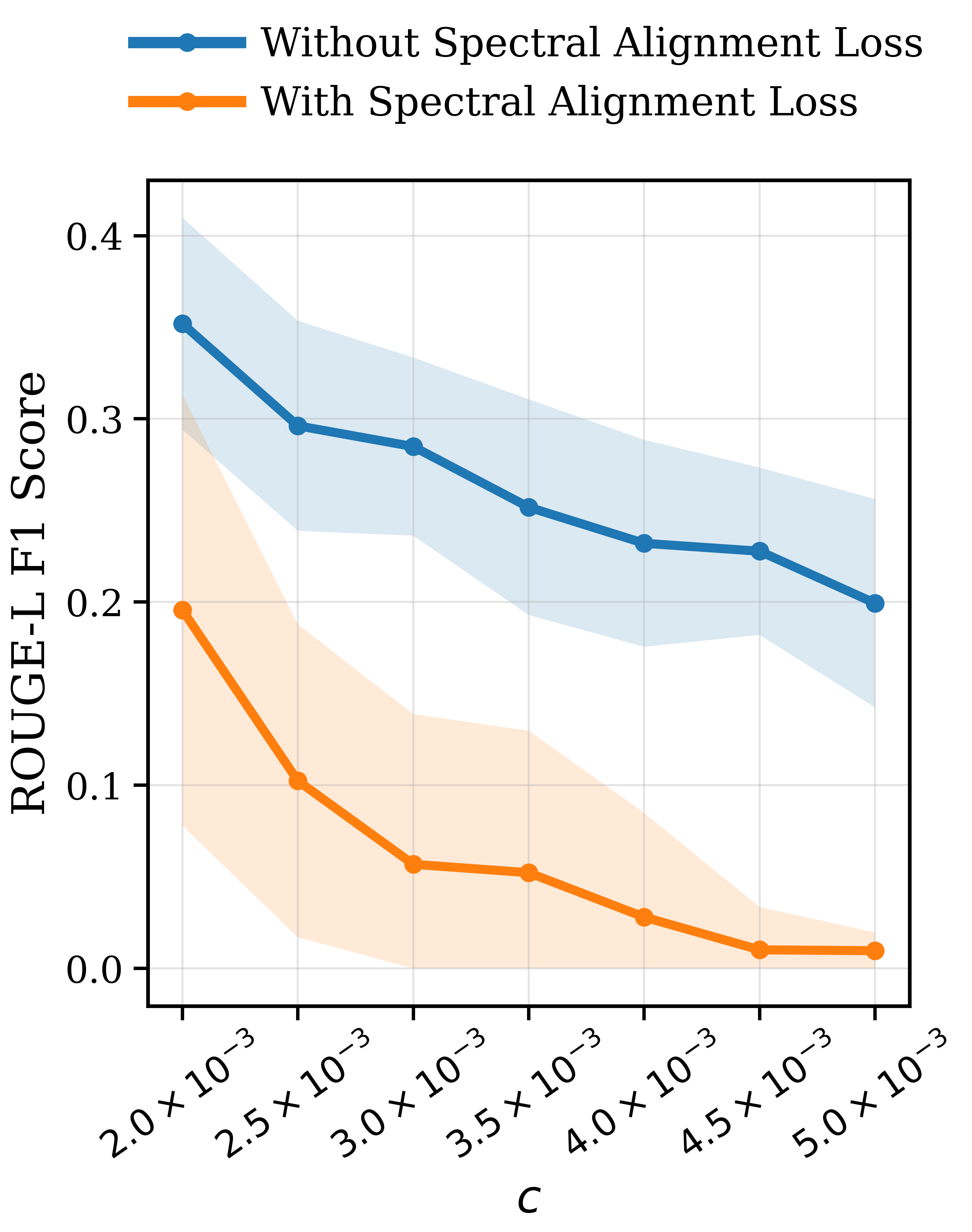}
    \end{subfigure}

    \caption{Effect of the spectral alignment loss.}
    \label{fig:ablation_loss}
\end{subfigure}
\hfill
\begin{subfigure}[t]{0.48\textwidth}
    \centering
    \begin{subfigure}[b]{0.49\linewidth}
        \centering
        \includegraphics[width=\linewidth,
        trim=0cm 0cm 0cm 0.7cm,clip]
        {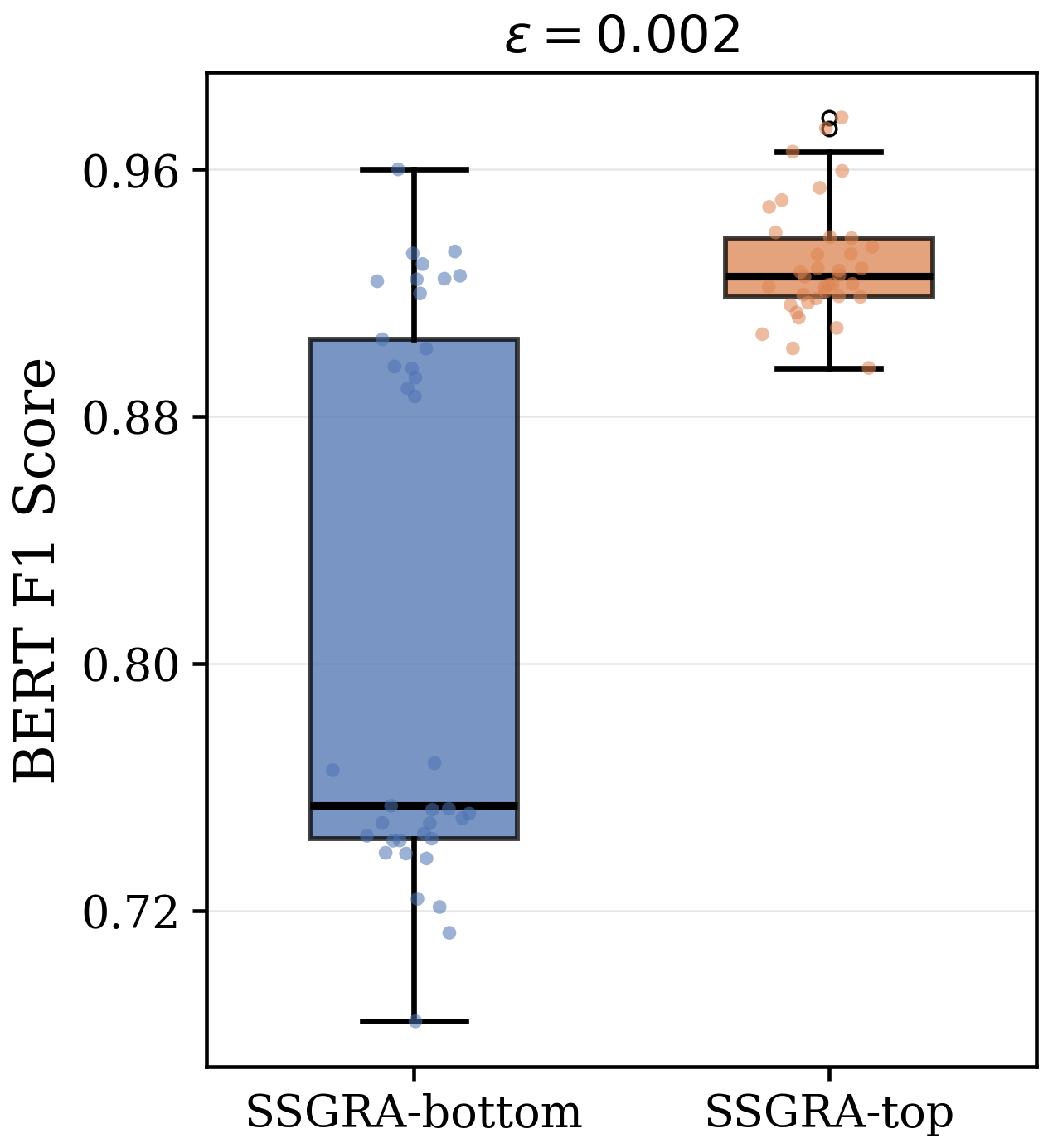}
    \end{subfigure}
    \hfill
    \begin{subfigure}[b]{0.49\linewidth}
        \centering
        \includegraphics[width=\linewidth,
        trim=0cm 0cm 0cm 0.7cm,clip]
        {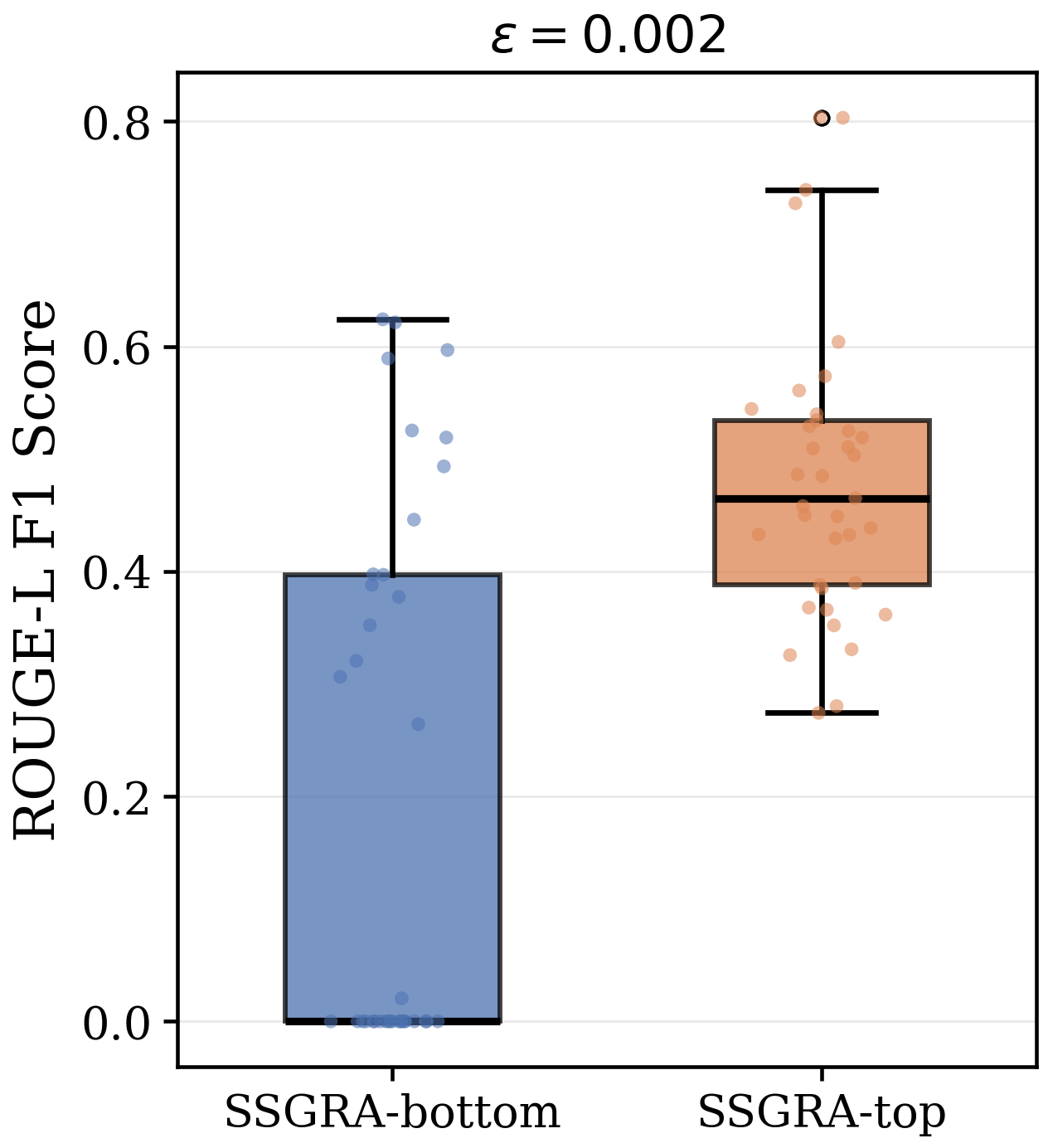}
    \end{subfigure}

    \caption{Bottom- vs. top-singular subspace alignment.}
    \label{fig:ablation_subspace}
\end{subfigure}
\caption{Ablation analysis validating the design choices of SSGRA.}
\label{fig:attack_comparison}
\end{figure*}

We ablate SSGRA on Qwen2.5-VL to evaluate (1) the contribution of the spectral alignment objective and (2) the effect of aligning adversarial representations with bottom versus top singular-vector subspaces.

\noindent\textbf{Effect of the Spectral Alignment Term.} Figure~\ref{fig:ablation_loss} compares SSGRA with and without the spectral alignment term in Eq.~\eqref{eq:ssgra}. Removing this term increases both BERTScore F1 and ROUGE-L F1 across all perturbation budgets, indicating weaker attacks and confirming that spectral alignment is the primary contributor to the improved attack effectiveness. The performance gap widens with increasing $c$, showing that the benefit of spectral guidance increases with the perturbation budget.

\noindent\textbf{Bottom vs.\ Top Singular Subspace.} Figure~\ref{fig:ablation_subspace} compares SSGRA-bottom and SSGRA-top, which align adversarial representations with the bottom-$k$ and top-$k$ singular subspaces, respectively. SSGRA-bottom achieves lower BERTScore and ROUGE-L scores, indicating stronger attacks, whereas SSGRA-top causes only minor degradation. These results support the hypothesis that bottom singular-vector subspaces constitute the primary spectral attack surface in VLMs.

\section{Conclusion}

We presented a spectral perspective on adversarial vulnerability in transformer-based VLMs by analyzing their intermediate linear transformations. We identified bottom singular-vector subspaces as a previously overlooked spectral attack surface and proposed SSGRA, which exploits this insight to improve attack effectiveness on three state-of-the-art VLMs. Our analyses show that near-null singular directions are substantially more prevalent than strongly amplifying ones and that untargeted adversarial optimization naturally tends to increase alignment with these information-attenuating subspaces. These findings suggest that, alongside existing spectral-norm regularization techniques for large singular values, controlling near-null singular directions may provide a complementary approach to improving adversarial robustness.

\noindent\textbf{Limitations}. This work focuses on untargeted white-box attacks to isolate and analyze the spectral mechanisms underlying adversarial vulnerability. The applicability of the proposed framework to transfer-based and black-box settings has not yet been investigated and remains future work.
\clearpage  

%
%
\bibliographystyle{splncs04}
\bibliography{main}

\clearpage

\appendix
\section{Appendix}

\subsection{Spectral Subspace Guided Representation Attack (SSGRA) Algorithm}
\begin{algorithm}[ht]
\caption{Spectral Subspace Guided Representation Attack (SSGRA)}
\label{alg:ssgra}
\textbf{Input}: Image $x$, prompt $p$, VLM $\mathcal{F}$, perturbation budget $c$, step size $\eta$, number of steps $T$, selected linear transformations $\mathcal{S}$, trade-off parameter $\lambda$, bottom-subspace dimension $s$\\
\textbf{Output}: Adversarial image $x_a^*$
\begin{algorithmic}[1]
\STATE Initialize perturbation with small random noise:
\STATE \hspace{1em} $\delta \sim \mathcal{U}(-\xi, \xi)$, where $\xi \ll c$
\STATE Fix prompt $p$ and suppress it in the representation notation.
\STATE Compute clean representations $\{h_k^{v,(j)}(x)\}_{k,j}$ and $\{H_\ell^{(i)}(x)\}_{\ell,i}$
\FOR{each selected transformation $m\in\mathcal{S}$}
    \STATE Compute SVD of its weight matrix:
    $
    W_m = U_m\Sigma_m V_m^\top
    $
    \STATE Extract the bottom-$s$ right singular-vector subspace $V^{\mathrm{bottom}}_{m,s}$
\ENDFOR
\FOR{$\tau=1$ to $T$}
    \STATE Construct the adversarial image:
    \STATE \hspace{1em}
    $x_a \leftarrow \mathrm{clip}(x+\delta,\;x-c,\;x+c)$
    \STATE Compute adversarial representations $\{h_k^{v,(j)}(x_a)\}_{k,j}$ and $\{H_\ell^{(i)}(x_a)\}_{\ell,i}$
    \STATE Compute BSA representation discrepancy term:    $$
    \mathcal{L}_{\mathrm{BSA}}
    =
    \sum_{k=1}^{K}\sum_{j=1}^{N_v}
    \cos\!\left(h_k^{v,(j)}(x),h_k^{v,(j)}(x_a)\right)
    +
    \sum_{\ell=1}^{L}\sum_{i=1}^{N}
    \cos\!\left(H_\ell^{(i)}(x),H_\ell^{(i)}(x_a)\right)
    $$
    \STATE Compute spectral subspace alignment term:
    \STATE \hspace{1em}
    $
    \mathcal{L}_{\mathrm{SS}}
    =
    \sum_{m\in\mathcal{S}}
    \Psi_s\!\left(
    z_m(x_a),
    V^{\mathrm{bottom}}_{m,s}
    \right)
    $
    \STATE Compute SSGRA objective:
    \STATE \hspace{1em}
    $
    \mathcal{L}_{\mathrm{SSGRA}}
    =
    -\lambda \mathcal{L}_{\mathrm{BSA}}
    +
    (1-\lambda)\mathcal{L}_{\mathrm{SS}}
    $
    \STATE Update perturbation by Adam ascent:
    \STATE \hspace{1em}
    $
    \delta \leftarrow \mathrm{AdamStep}\!\left(\delta,\nabla_\delta \mathcal{L}_{\mathrm{SSGRA}}\right)
    $
    \STATE Project perturbation onto the $L_\infty$ ball:
    \STATE \hspace{1em}
    $
    \delta \leftarrow \mathrm{clip}(\delta,-c,c)
    $
\ENDFOR
\STATE Construct the final adversarial image:
\STATE \hspace{1em}
$x_a^* \leftarrow \mathrm{clip}(x+\delta,\;x-c,\;x+c)$
\STATE \textbf{return} $x_a^*$
\end{algorithmic}
\end{algorithm}


Algorithm~\ref{alg:ssgra} summarizes the optimization procedure of SSGRA. The attack first fixes the textual prompt and computes the clean intermediate representations of the input image across the visual encoder and language-model blocks. For each selected intermediate linear transformation, SSGRA performs an SVD of the corresponding weight matrix and extracts the bottom-$s$ right singular-vector subspace. These subspaces define the information-attenuating directions used by the spectral alignment objective.

During optimization, the adversarial image is constructed by adding a learnable perturbation $\delta$ to the clean image and clipping it within the prescribed $L_\infty$ budget. At each iteration, the model is evaluated on the current adversarial image to obtain the corresponding intermediate representations. SSGRA then combines two objectives. The first is the BSA representation-discrepancy term, which reduces the similarity between clean and adversarial feature representations across visual and language layers. The second is the proposed spectral subspace alignment term, which encourages adversarial representations before the selected transformations to align with the bottom singular-vector subspaces. The trade-off parameter $\lambda$ balances these two effects.

The perturbation is updated by Adam ascent on the combined SSGRA objective, since the attack maximizes representation disruption and spectral alignment. After each update, the perturbation is projected back onto the $L_\infty$ ball to ensure that the adversarial image remains visually close to the original input. The final adversarial example is obtained by applying the optimized perturbation to the clean image and clipping it to satisfy the perturbation constraint. The algorithm explicitly guides adversarial representations toward directions that are strongly attenuated by intermediate linear transformations, thereby weakening semantic information propagation through the VLM.

\subsection{Evaluation Metrics Details}
\label{sec:evalDetails}
We evaluate attack effectiveness using two complementary text-based metrics 
that compare the adversarial model output $\hat{y}_a$ against the clean 
output $\hat{y}_c$.

\noindent\textbf{BERTScore.} BERTScore~\cite{zhang2019bertscore} computes token-level 
semantic similarity between $\hat{y}_a$ and $\hat{y}_c$ using contextual 
embeddings from a pretrained language model (RoBERTa-large). For each token 
$a_i \in \hat{y}_a$ and $c_j \in \hat{y}_c$, cosine similarity is computed 
in embedding space. Precision, recall, and F1 are defined as:
\begin{equation}
    P_{\text{BERT}} = \frac{1}{|\hat{y}_a|}\sum_{a_i \in \hat{y}_a} \max_{c_j \in \hat{y}_c} \cos(\mathbf{e}_{a_i}, \mathbf{e}_{c_j}),
\end{equation}
\begin{equation}
    R_{\text{BERT}} = \frac{1}{|\hat{y}_c|}\sum_{c_j \in \hat{y}_c} \max_{a_i \in \hat{y}_a} \cos(\mathbf{e}_{a_i}, \mathbf{e}_{c_j}),
\end{equation}
\begin{equation}
    F1_{\text{BERT}} = \frac{2 \cdot P_{\text{BERT}} \cdot R_{\text{BERT}}}{P_{\text{BERT}} + R_{\text{BERT}}},
\end{equation}

where $\mathbf{e}_{a_i}$ and $\mathbf{e}_{c_j}$ are the contextual embeddings 
of tokens $a_i$ and $c_j$ respectively. Lower scores indicate greater semantic 
degradation of the adversarial output relative to the clean output.

\noindent\textbf{ROUGE-L.} ROUGE-L~\cite{lin2004rouge} measures lexical overlap via 
the Longest Common Subsequence (LCS) between $\hat{y}_a$ and $\hat{y}_c$. 
Let $\text{LCS}(\hat{y}_a, \hat{y}_c)$ denote the length of the longest common 
subsequence. Precision, recall, and F1 are:
\begin{equation}
    P_{\text{R}} = \frac{|\text{LCS}(\hat{y}_a, \hat{y}_c)|}{|\hat{y}_a|}, \quad
    R_{\text{R}} = \frac{|\text{LCS}(\hat{y}_a, \hat{y}_c)|}{|\hat{y}_c|}, \quad
    F1_{\text{R}} = \frac{2 \cdot P_{\text{R}} \cdot R_{\text{R}}}{P_{\text{R}} + R_{\text{R}}}.
\end{equation}
Unlike BERTScore, ROUGE-L is sensitive to structural content loss: a low recall 
indicates that the adversarial output fails to reproduce key content from the 
clean description. The two metrics are complementary — BERTScore captures 
semantic similarity robust to paraphrase, while ROUGE-L captures lexical 
fidelity and structural degradation. We report mean and standard deviation 
over 100 images per experimental configuration.

\begin{table*}[ht]
\centering
\caption{
Performance of different attack methods under varying perturbation budgets.
\textbf{(a)} BERTScore F1.
\textbf{(b)} ROUGE-L F1.
Lower values indicate stronger attack effectiveness.
}
\label{tab:all_results}

\scriptsize
\setlength{\tabcolsep}{2pt}
\renewcommand{\arraystretch}{0.72}

\resizebox{\textwidth}{!}{%
{\scriptsize
\begin{tabular}{lccccccc}
\toprule
\multicolumn{8}{c}{\textbf{(a) BERTScore F1}}\\
\midrule
Method
& $c=0.002$
& $c=0.0025$
& $c=0.003$
& $c=0.0035$
& $c=0.004$
& $c=0.0045$
& $c=0.005$\\
\midrule

\multicolumn{8}{c}{\textbf{Qwen2.5-VL}} \\
\midrule
BSA
& $0.895\pm0.028$
& $\underline{0.877}\pm0.035$
& $0.874\pm0.032$
& $\underline{0.864}\pm0.037$
& $\underline{0.861}\pm0.035$
& $\underline{0.856}\pm0.038$
& $\underline{0.845}\pm0.037$ \\
DRA
& $0.938\pm0.026$
& $0.943\pm0.023$
& $0.936\pm0.023$
& $0.934\pm0.023$
& $0.935\pm0.021$
& $0.936\pm0.023$
& $0.931\pm0.018$ \\
FDA
& $0.933\pm0.021$
& $0.928\pm0.025$
& $0.929\pm0.023$
& $0.925\pm0.022$
& $0.922\pm0.023$
& $0.918\pm0.022$
& $0.918\pm0.024$ \\
SSPA
& $0.926\pm0.018$
& $0.928\pm0.020$
& $0.919\pm0.020$
& $0.922\pm0.022$
& $0.917\pm0.020$
& $0.913\pm0.019$
& $0.916\pm0.019$ \\
EGA
& $0.896\pm0.022$
& $0.883\pm0.022$
& $0.883\pm0.024$
& $0.865\pm0.053$
& $0.872\pm0.031$
& $\underline{0.856}\pm0.061$
& $0.865\pm0.034$ \\
CE
& $\underline{0.886}\pm0.016$
& $0.884\pm0.017$
& $\underline{0.873}\pm0.028$
& $0.875\pm0.034$
& $0.871\pm0.030$
& $0.865\pm0.048$
& $0.856\pm0.048$ \\
SSGRA
& $\mathbf{0.816}\pm0.088$
& $\mathbf{0.768}\pm0.147$
& $\mathbf{0.743}\pm0.137$
& $\mathbf{0.704}\pm0.217$
& $\mathbf{0.754}\pm0.042$
& $\mathbf{0.687}\pm0.206$
& $\mathbf{0.724}\pm0.121$ \\

\textbf{Gain over Best (\%)} 
& \textbf{7.90 \%}
& \textbf{12.40 \%}
& \textbf{14.89 \%}
& \textbf{18.52 \%}
& \textbf{12.43 \%}
& \textbf{19.74 \%}
& \textbf{14.32\%} \\

\midrule

\multicolumn{8}{c}{\textbf{LLaVa-1.5}} \\
\midrule
BSA
& $0.926\pm0.034$
& ${0.911}\pm0.034$
& $\underline{0.902}\pm0.028$
& $\underline{0.895}\pm0.019$
& $\underline{0.892}\pm0.020$
& $0.886\pm0.028$
& $0.884\pm0.019$ \\

DRA
& $0.951\pm0.033$
& $0.940\pm0.022$
& $0.935\pm0.027$
& $0.932\pm0.030$
& $0.926\pm0.027$
& $0.914\pm0.033$
& $0.915\pm0.028$ \\

FDA
& $0.954\pm0.027$
& $0.946\pm0.026$
& $0.945\pm0.027$
& $0.944\pm0.027$
& $0.939\pm0.026$
& $0.942\pm0.026$
& $0.939\pm0.025$ \\

SSPA
& $0.949\pm0.027$
& $0.947\pm0.024$
& $0.936\pm0.021$
& $0.934\pm0.023$
& $0.933\pm0.026$
& $0.928\pm0.022$
& $0.934\pm0.026$ \\

EGA
& $0.933\pm0.031$
& $0.922\pm0.023$
& $0.918\pm0.026$
& $0.913\pm0.027$
& $0.903\pm0.027$
& $\underline{0.885}\pm0.131$
& $\underline{0.875}\pm0.133$ \\

CE
& $\underline{0.921}\pm0.022$
& $0.916\pm0.022$
& $0.911\pm0.020$
& $0.909\pm0.018$
& $0.907\pm0.019$
& $0.888\pm0.129$
& $0.883\pm0.130$ \\


SSGRA
& $\mathbf{0.916}\pm0.029$
& $0.912\pm0.027$
& $\mathbf{0.882}\pm0.133$
& $\mathbf{0.837}\pm0.216$
& $\mathbf{0.853}\pm0.179$
& $\mathbf{0.819}\pm0.213$
& $\mathbf{0.816}\pm0.214$ \\

\textbf{Gain over Best (\%)} 
& \textbf{0.54 \%}
& --
& \textbf{2.22 \%}
& \textbf{6.48 \%}
& \textbf{4.37 \%}
& \textbf{7.46 \%}
& \textbf{6.74\%} \\
\midrule

\multicolumn{8}{c}{\textbf{Gemma 3}} \\
\midrule
BSA
& $0.905\pm0.035$
& $0.890\pm0.034$
& $\underline{0.880}\pm0.035$
& $\underline{0.870}\pm0.043$
& $\underline{0.861}\pm0.031$
& $\underline{0.856}\pm0.031$
& ${0.847}\pm0.037$ \\

DRA
& $0.919\pm0.027$
& $0.918\pm0.025$
& $0.914\pm0.029$
& $0.911\pm0.025$
& $0.908\pm0.029$
& $0.909\pm0.028$
& $0.903\pm0.030$ \\

FDA
& $0.927\pm0.036$
& $0.924\pm0.037$
& $0.923\pm0.032$
& $0.919\pm0.034$
& $0.923\pm0.032$
& $0.921\pm0.030$
& $0.918\pm0.026$ \\

SSPA
& $0.923\pm0.022$
& $0.918\pm0.028$
& $0.916\pm0.032$
& $0.913\pm0.029$
& $0.913\pm0.031$
& $0.905\pm0.027$
& $0.909\pm0.031$ \\

EGA
& $0.911\pm0.029$
& $0.918\pm0.031$
& $0.908\pm0.030$
& $0.907\pm0.031$
& $0.901\pm0.031$
& $0.908\pm0.030$
& $0.902\pm0.030$ \\

CE
& ${0.886}\pm0.026$
& ${0.883}\pm0.025$
& $0.879\pm0.030$
& $0.879\pm0.026$
& $0.877\pm0.033$
& $0.881\pm0.034$
& $0.867\pm0.041$ \\


SSGRA
& $0.903\pm0.038$
& ${0.883}\pm0.042$
& $\mathbf{0.873}\pm0.036$
& $\mathbf{0.868}\pm0.032$
& $\mathbf{0.858}\pm0.034$
& $\mathbf{0.855}\pm0.037$
& ${0.849}\pm0.030$ \\

\textbf{Gain over Best (\%)} 
& --
& --
& \textbf{0.80 \%}
& \textbf{0.23 \%}
& \textbf{0.35 \%}
& \textbf{0.12 \%}
& -- \\

\bottomrule
\end{tabular}
}
}

\vspace{0.5em}

\resizebox{\textwidth}{!}{%
{\scriptsize
\begin{tabular}{lccccccc}
\toprule
\multicolumn{8}{c}{\textbf{(b) ROUGE-L F1}}\\
\midrule
Method
& $c=0.002$
& $c=0.0025$
& $c=0.003$
& $c=0.0035$
& $c=0.004$
& $c=0.0045$
& $c=0.005$\\
\midrule

\multicolumn{8}{c}{\textbf{Qwen2.5-VL}} \\
\midrule
BSA
& $0.352\pm0.116$
& $0.296\pm0.115$
& $0.285\pm0.097$
& $\underline{0.252}\pm0.118$
& $\underline{0.232}\pm0.113$
& $0.228\pm0.091$
& $\underline{0.199}\pm0.114$ \\
DRA
& $0.551\pm0.187$
& $0.577\pm0.171$
& $0.536\pm0.170$
& $0.539\pm0.164$
& $0.528\pm0.149$
& $0.533\pm0.159$
& $0.501\pm0.132$ \\
FDA
& $0.517\pm0.160$
& $0.486\pm0.153$
& $0.494\pm0.153$
& $0.471\pm0.148$
& $0.447\pm0.144$
& $0.440\pm0.131$
& $0.434\pm0.139$ \\
SSPA
& $0.498\pm0.122$
& $0.494\pm0.136$
& $0.437\pm0.099$
& $0.447\pm0.142$
& $0.428\pm0.119$
& $0.406\pm0.107$
& $0.427\pm0.121$ \\
EGA
& $0.337\pm0.113$
& $0.294\pm0.086$
& $0.293\pm0.097$
& $0.277\pm0.100$
& $0.271\pm0.075$
& $0.261\pm0.077$
& $0.274\pm0.065$ \\
CE
& $\underline{0.282}\pm0.066$
& $\underline{0.277}\pm0.058$
& $\underline{0.239}\pm0.071$
& $0.256\pm0.076$
& $0.235\pm0.057$
& $\underline{0.226}\pm0.065$
& $0.217\pm0.081$ \\
SSGRA
& $\mathbf{0.196}\pm0.235$
& $\mathbf{0.102}\pm0.171$
& $\mathbf{0.057}\pm0.164$
& $\mathbf{0.052}\pm0.155$
& $\mathbf{0.028}\pm0.114$
& $\mathbf{0.010}\pm0.047$
& $\mathbf{0.010}\pm0.020$ \\

\textbf{Gain over Best (\%)} 
& \textbf{30.50 \%}
& \textbf{63.18 \%}
& \textbf{76.15 \%}
& \textbf{79.37 \%}
& \textbf{97.93 \%}
& \textbf{95.57 \%}
& \textbf{94.97\%} \\

\midrule

\multicolumn{8}{c}{\textbf{LLaVa-1.5}} \\
\midrule
BSA
& $0.492\pm0.215$
& $0.416\pm0.196$
& $\underline{0.369}\pm0.135$
& $\underline{0.330}\pm0.080$
& $\underline{0.320}\pm0.103$
& $\underline{0.297}\pm0.077$
& $\underline{0.291}\pm0.065$ \\
DRA
& $0.639\pm0.212$
& $0.545\pm0.164$
& $0.525\pm0.183$
& $0.514\pm0.201$
& $0.477\pm0.152$
& $0.426\pm0.164$
& $0.432\pm0.156$ \\
FDA
& $0.644\pm0.195$
& $0.591\pm0.182$
& $0.590\pm0.203$
& $0.585\pm0.192$
& $0.550\pm0.182$
& $0.570\pm0.184$
& $0.558\pm0.182$ \\
SSPA
& $0.620\pm0.202$
& $0.594\pm0.177$
& $0.524\pm0.144$
& $0.520\pm0.154$
& $0.520\pm0.170$
& $0.483\pm0.147$
& $0.518\pm0.173$ \\
EGA
& $0.511\pm0.202$
& $0.456\pm0.138$
& $0.427\pm0.162$
& $0.416\pm0.151$
& $0.371\pm0.138$
& $0.377\pm0.143$
& $0.336\pm0.117$ \\
CE
& $\underline{0.439}\pm0.134$
& $0.400\pm0.137$
& $0.375\pm0.122$
& $0.375\pm0.094$
& $0.357\pm0.096$
& $0.340\pm0.088$
& $0.329\pm0.100$ \\
SSGRA
& $\mathbf{0.422}\pm0.169$
& ${0.406}\pm0.157$
& $\mathbf{0.364}\pm0.184$
& $\mathbf{0.299}\pm0.138$
& $\mathbf{0.297}\pm0.132$
& $\mathbf{0.252}\pm0.121$
& $\mathbf{0.236}\pm0.138$ \\

\textbf{Gain over Best (\%)} 
& \textbf{3.87 \%}
& --
& \textbf{1.35 \%}
& \textbf{9.39 \%}
& \textbf{7.19 \%}
& \textbf{15.15 \%}
& \textbf{18.90\%} \\

\midrule

\multicolumn{8}{c}{\textbf{Gemma 3}} \\
\midrule
BSA
& $0.406\pm0.142$
& $0.331\pm0.116$
& $0.315\pm0.112$
& $0.287\pm0.116$
& ${0.253}\pm0.098$
& ${0.238}\pm0.092$
& $\underline{0.228}\pm0.109$ \\
DRA
& $0.456\pm0.140$
& $0.457\pm0.121$
& $0.433\pm0.131$
& $0.405\pm0.120$
& $0.388\pm0.134$
& $0.402\pm0.126$
& $0.365\pm0.125$ \\
FDA
& $0.516\pm0.186$
& $0.509\pm0.186$
& $0.492\pm0.166$
& $0.471\pm0.157$
& $0.478\pm0.139$
& $0.474\pm0.152$
& $0.450\pm0.121$ \\
SSPA
& $0.464\pm0.118$
& $0.441\pm0.131$
& $0.440\pm0.141$
& $0.418\pm0.134$
& $0.422\pm0.129$
& $0.388\pm0.128$
& $0.404\pm0.126$ \\
EGA
& $0.423\pm0.118$
& $0.463\pm0.144$
& $0.408\pm0.125$
& $0.410\pm0.128$
& $0.382\pm0.127$
& $0.420\pm0.131$
& $0.390\pm0.119$ \\
CE
& ${0.289}\pm0.080$
& ${0.285}\pm0.071$
& ${0.269}\pm0.079$
& $\underline{0.275}\pm0.082$
& $0.273\pm0.088$
& $0.285\pm0.081$
& $0.259\pm0.085$ \\
SSGRA
& $0.407\pm0.136$
& $0.319\pm0.118$
& $0.288\pm0.119$
& $\mathbf{0.272}\pm0.105$
& $0.259\pm0.104$
& $0.241\pm0.088$
& $\mathbf{0.221}\pm0.093$ \\

\textbf{Gain over Best (\%)} 
& --
& --
& --
& \textbf{1.09 \%}
& --
& --
& \textbf{3.07\%} \\

\bottomrule
\end{tabular}
}
}

\end{table*}

\begin{figure*}[t]
\centering
\begin{subfigure}[b]{0.7\textwidth}
    \centering
    \includegraphics[width=\linewidth]{pictures/legend/legend.png}
\end{subfigure}
\begin{subfigure}[t]{0.32\textwidth}
    \centering
    \includegraphics[width=\linewidth, trim=0cm 0cm 0cm 2.4cm, clip]{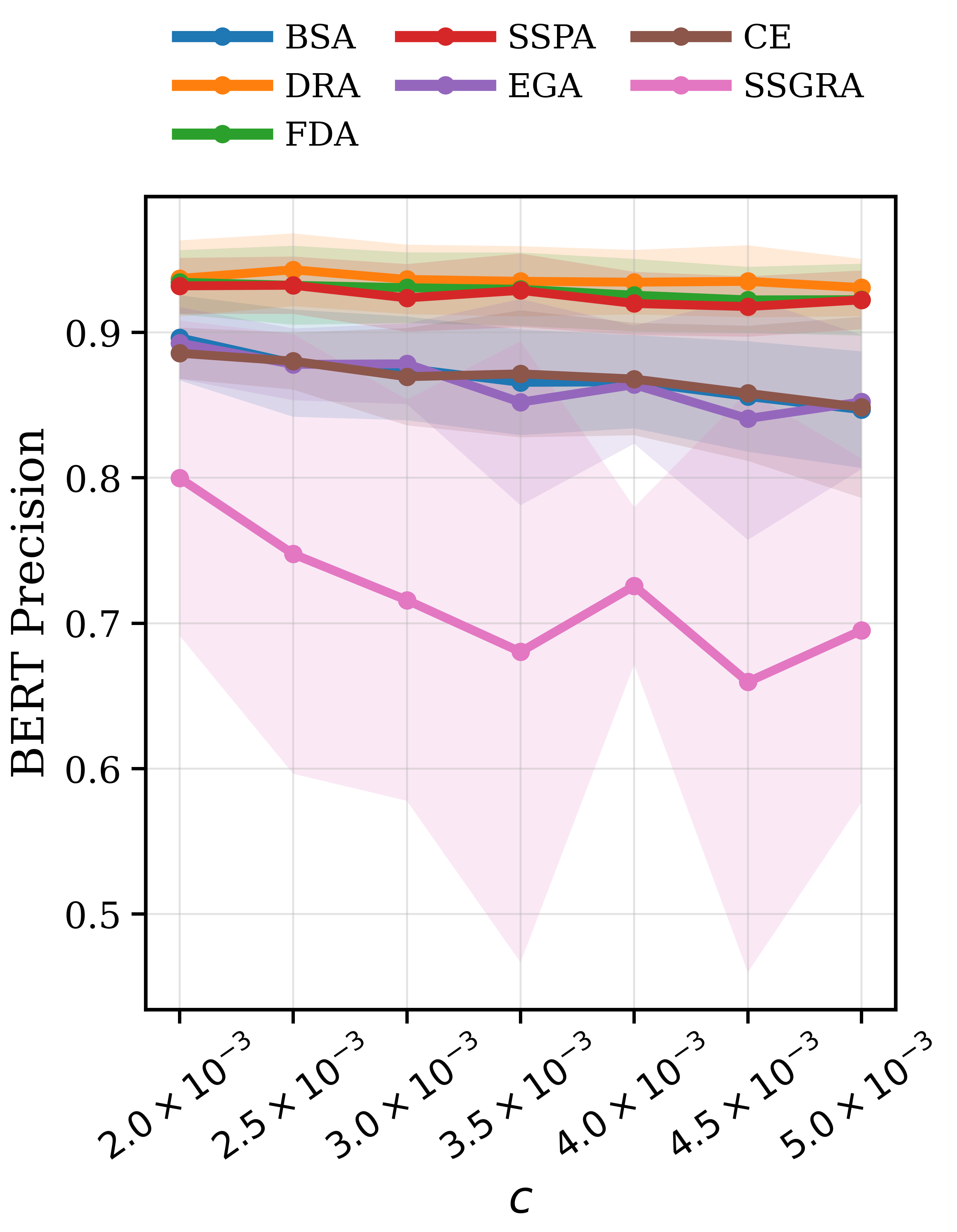}
    \caption{Precision(Qwen2.5-VL)}
\end{subfigure}
\hfill
\begin{subfigure}[t]{0.32\textwidth}
    \centering
    \includegraphics[width=\linewidth, trim=0cm 0cm 0cm 2.4cm, clip]{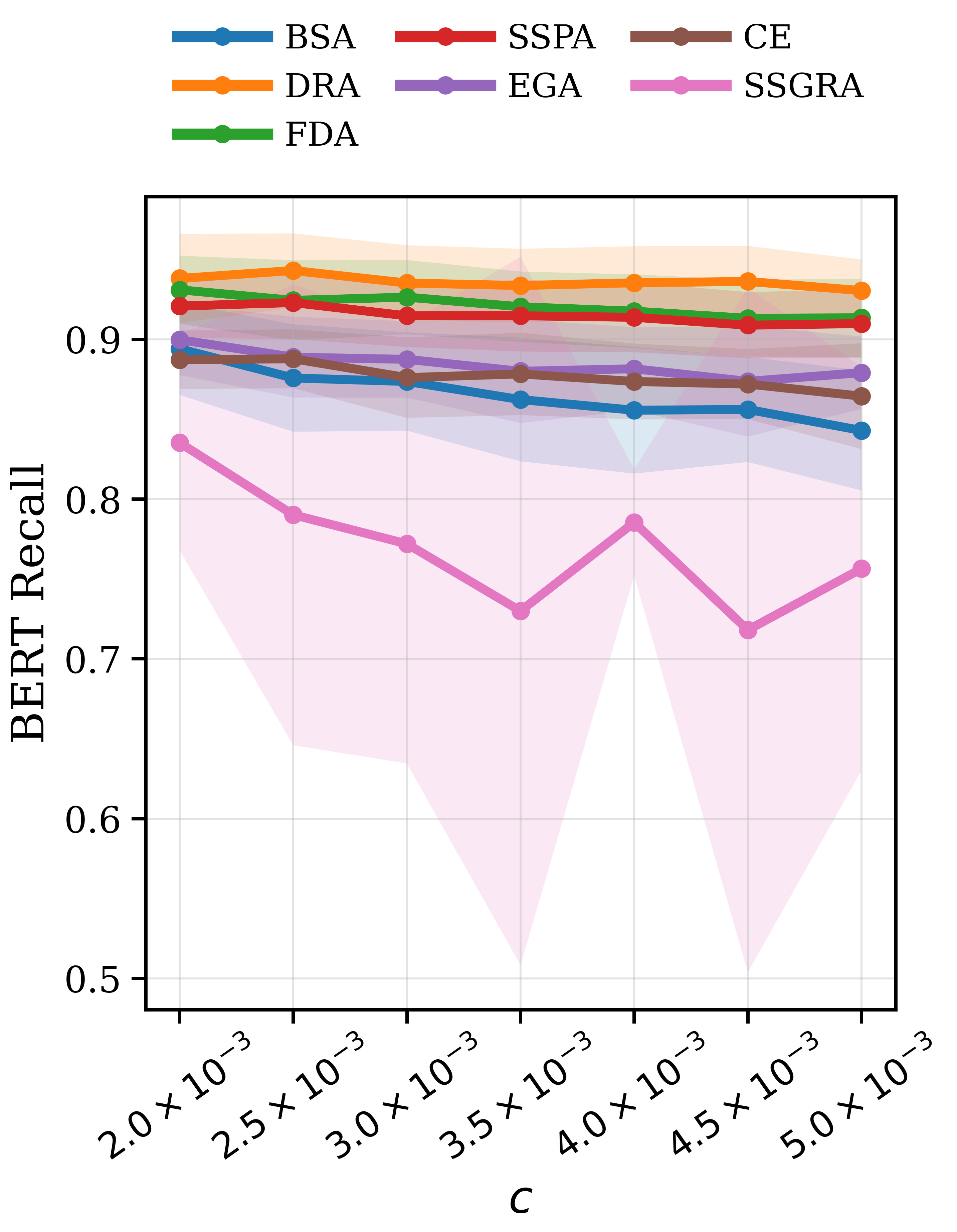}
    \caption{Recall (Qwen2.5-VL)}
\end{subfigure}
\hfill
\begin{subfigure}[t]{0.32\textwidth}
    \centering
    \includegraphics[width=\linewidth, trim=0cm 0cm 0cm 2.4cm, clip]{pictures/qwen2p5/epsilonSeries1/F1ComparisionSeries_num_steps_1000_AttackStartLayer_0_towardsNull_0.5_numSamplesConsidered_38_gate_proj_gate_proj.png}
    \caption{F1 (Qwen2.5-VL)}
\end{subfigure}

\vspace{0.6em}

\begin{subfigure}[t]{0.32\textwidth}
    \centering
    \includegraphics[width=\linewidth, trim=0cm 0cm 0cm 2.4cm, clip]{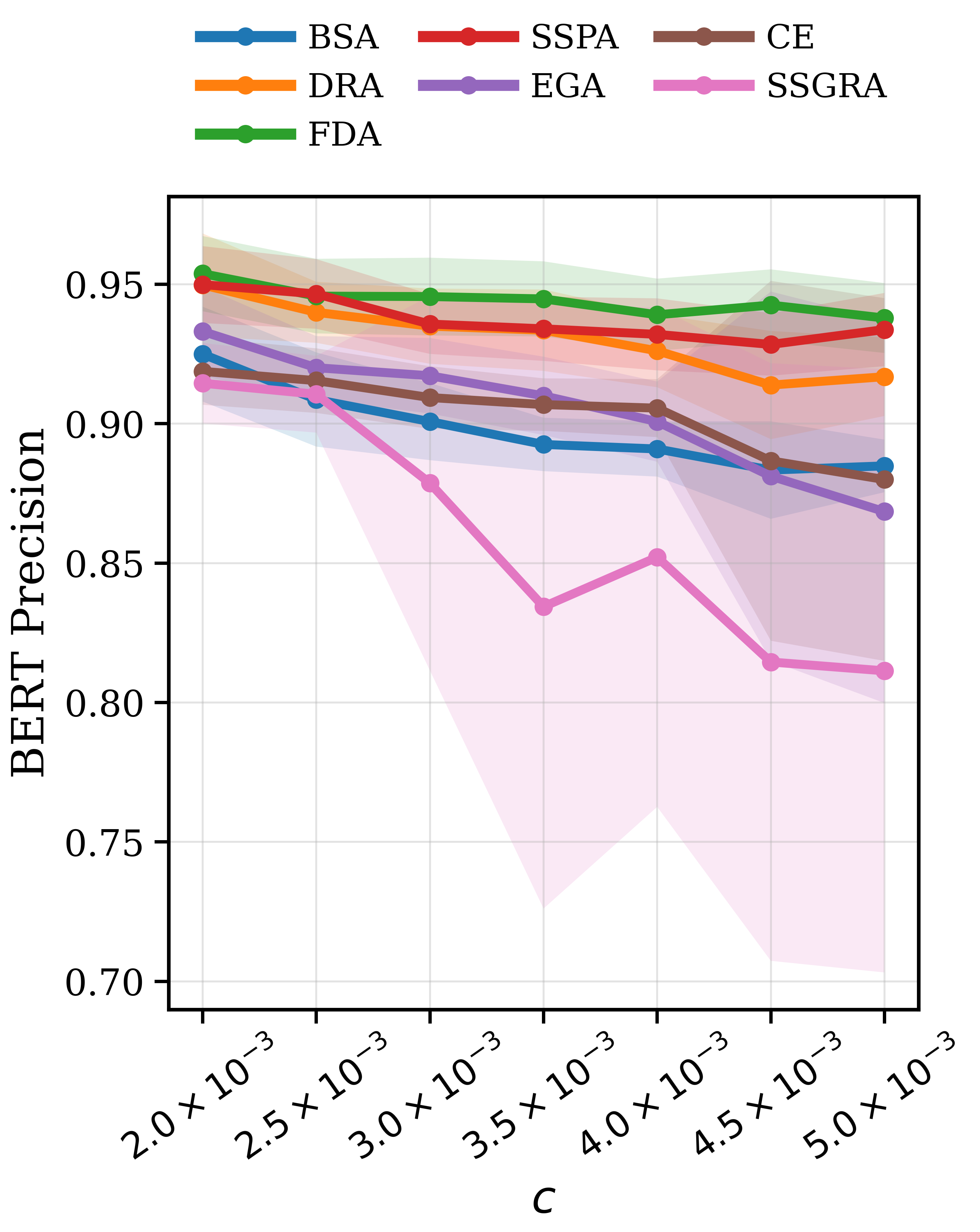}
    \caption{Precision (LLaVa-1.5)}
\end{subfigure}
\hfill
\begin{subfigure}[t]{0.32\textwidth}
    \centering
    \includegraphics[width=\linewidth, trim=0cm 0cm 0cm 2.4cm, clip]{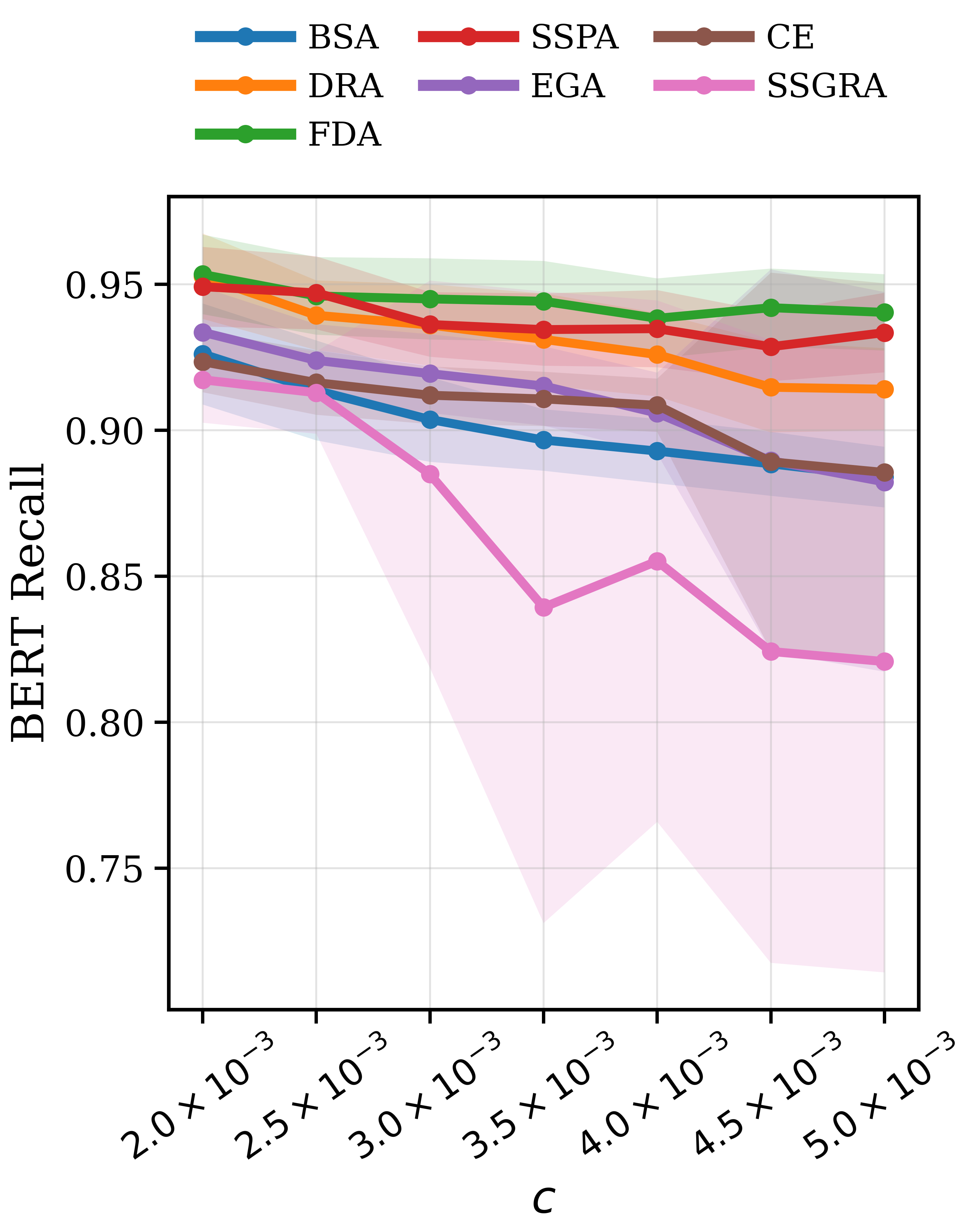}
    \caption{Recall (LLaVa-1.5)}
\end{subfigure}
\hfill
\begin{subfigure}[t]{0.32\textwidth}
    \centering
    \includegraphics[width=\linewidth, trim=0cm 0cm 0cm 2.4cm, clip]{pictures/Llava1p5/epsilonSeries1/F1ComparisonSeries_AllAttacksnum_steps_1000_AttackStartLayer_0_towardsNull_0.1_numSamplesConsidered_50.png}
    \caption{F1 (LLaVa-1.5)}
\end{subfigure}

\vspace{0.6em}

\begin{subfigure}[t]{0.32\textwidth}
    \centering
    \includegraphics[width=\linewidth, trim=0cm 0cm 0cm 2.4cm, clip]{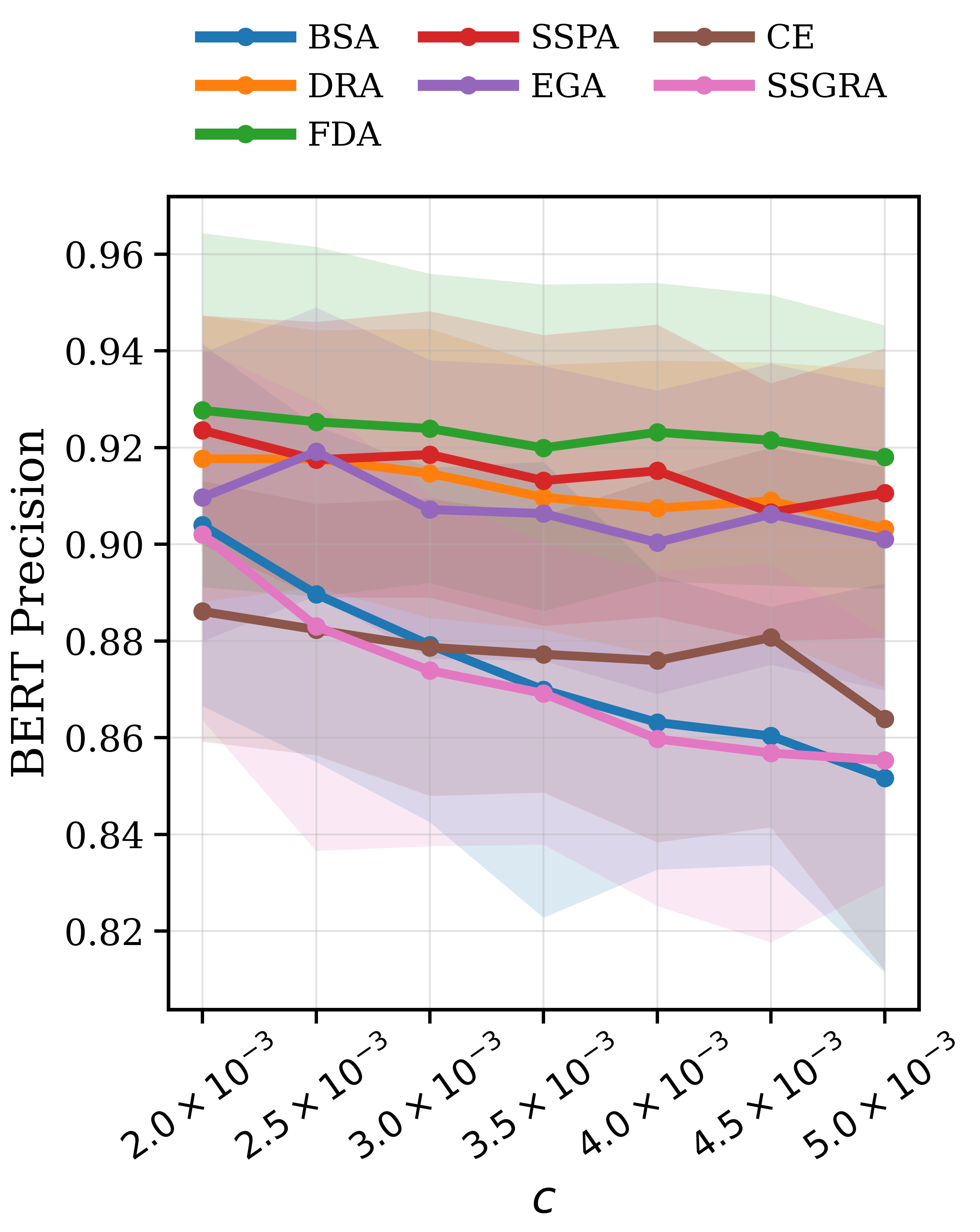}
    \caption{Precision (Gemma 3)} 
\end{subfigure}
\hfill
\begin{subfigure}[t]{0.32\textwidth}
    \centering
    \includegraphics[width=\linewidth, trim=0cm 0cm 0cm 2.4cm, clip]{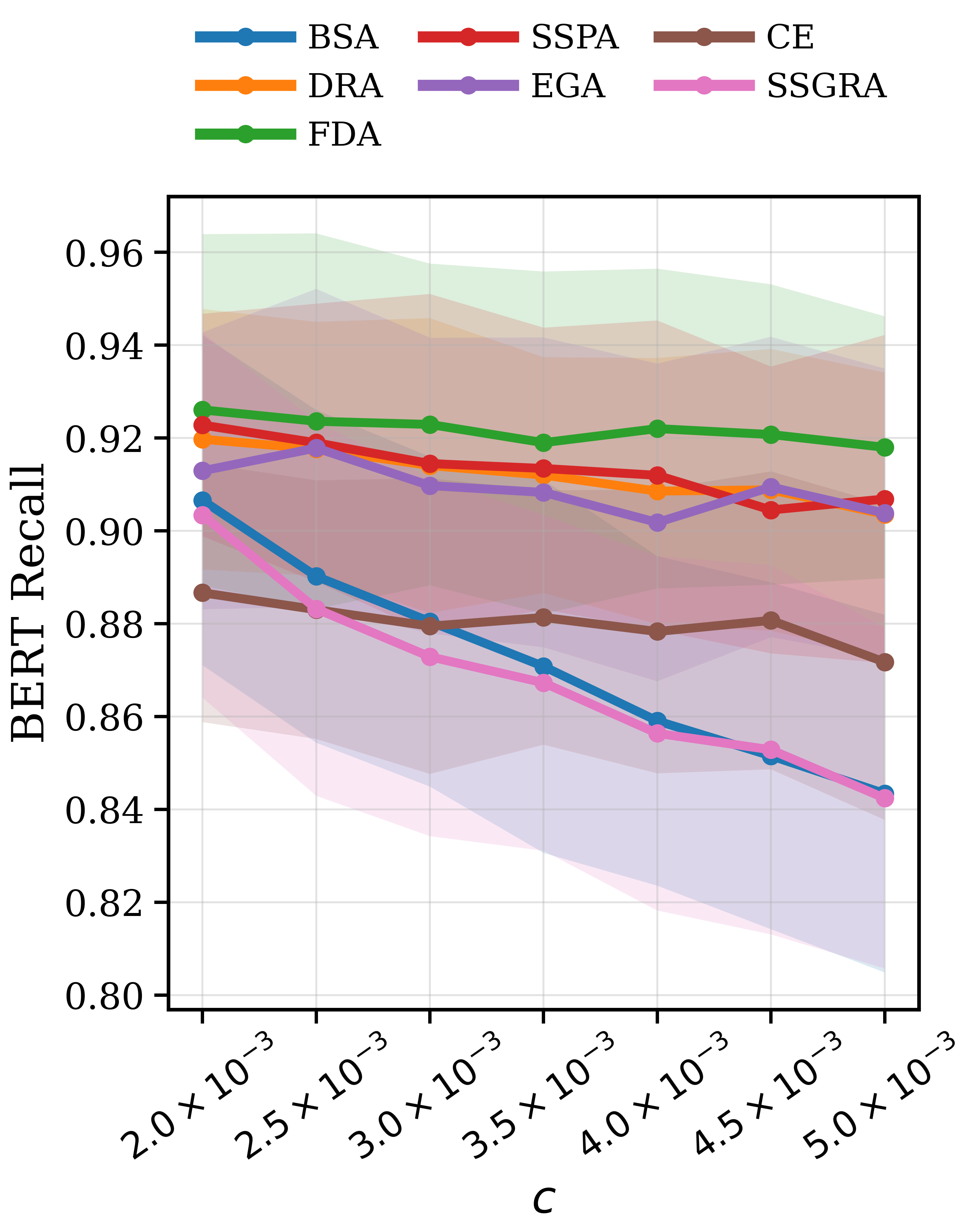}
    \caption{Recall (Gemma 3)}
\end{subfigure}
\hfill
\begin{subfigure}[t]{0.32\textwidth}
    \centering
    \includegraphics[width=\linewidth, trim=0cm 0cm 0cm 2.4cm, clip]{pictures/gemma3/epsilonSeries1/F1ComparisonSeries_num_steps_1000_AttackStartLayer_0_AttackStartLayerVis_11_towardsNull_0.1.png}
    \caption{F1 (Gemma 3)}
\end{subfigure}

\caption{BERT-score comparison of different adversarial attack methods across perturbation budgets for Qwen~2.5-VL, LLaVa~1.5, and Gemma~3. Each row corresponds to a model, while the columns show Precision, Recall, and F1 score, respectively.}
\label{fig:bertscore_comparison_all_models}

\end{figure*}

\begin{figure*}[t]
\centering
\begin{subfigure}[b]{0.7\textwidth}
    \centering
    \includegraphics[width=\linewidth]{pictures/legend/legend.png}
\end{subfigure}

\begin{subfigure}[t]{0.32\linewidth}
    \centering
    \includegraphics[width=\linewidth, trim=0cm 0cm 0cm 2.4cm, clip]{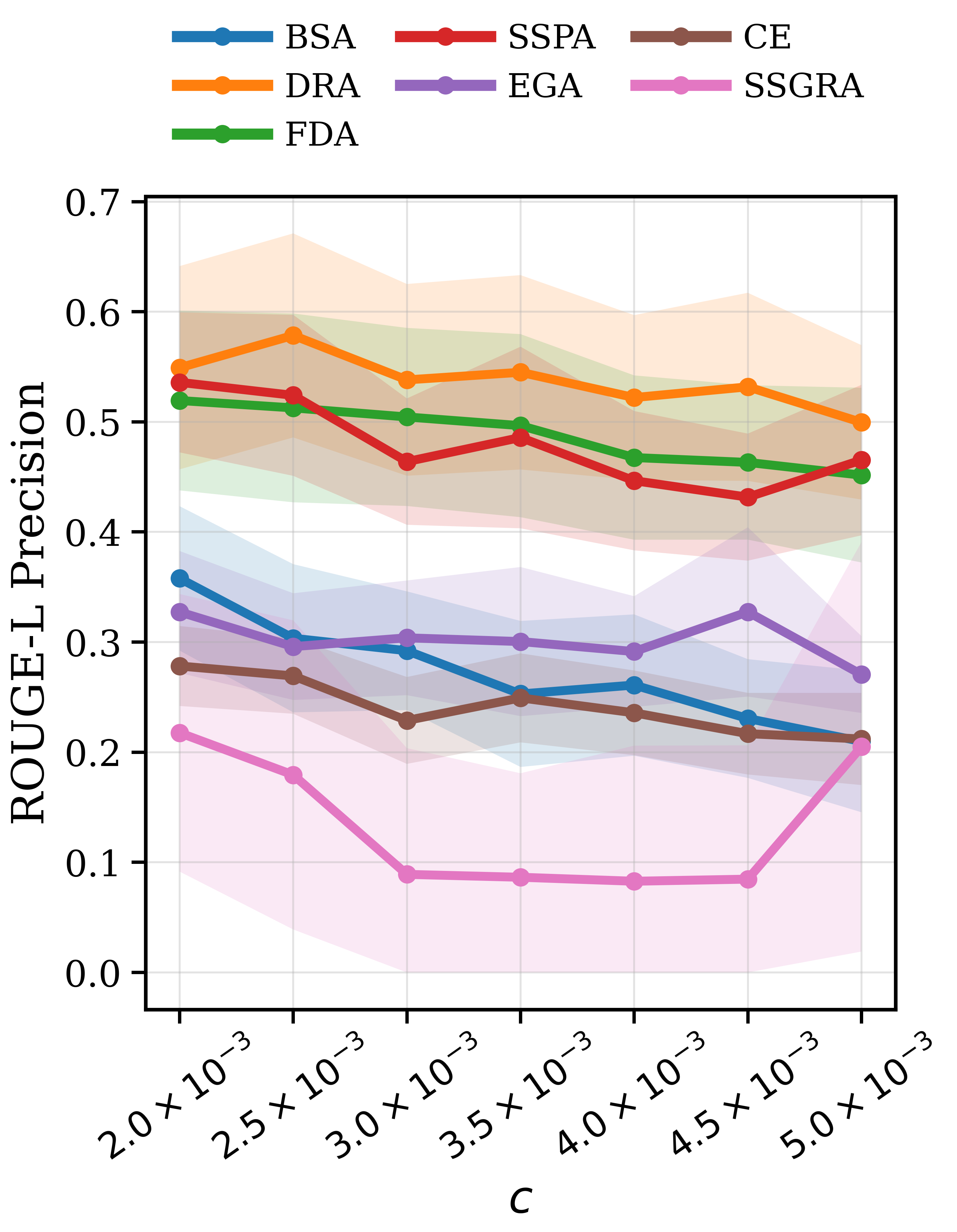}
    \caption{Precision (Qwen2.5-VL)}
\end{subfigure}
\hfill
\begin{subfigure}[t]{0.32\linewidth}
    \centering
    \includegraphics[width=\linewidth, trim=0cm 0cm 0cm 2.4cm, clip]{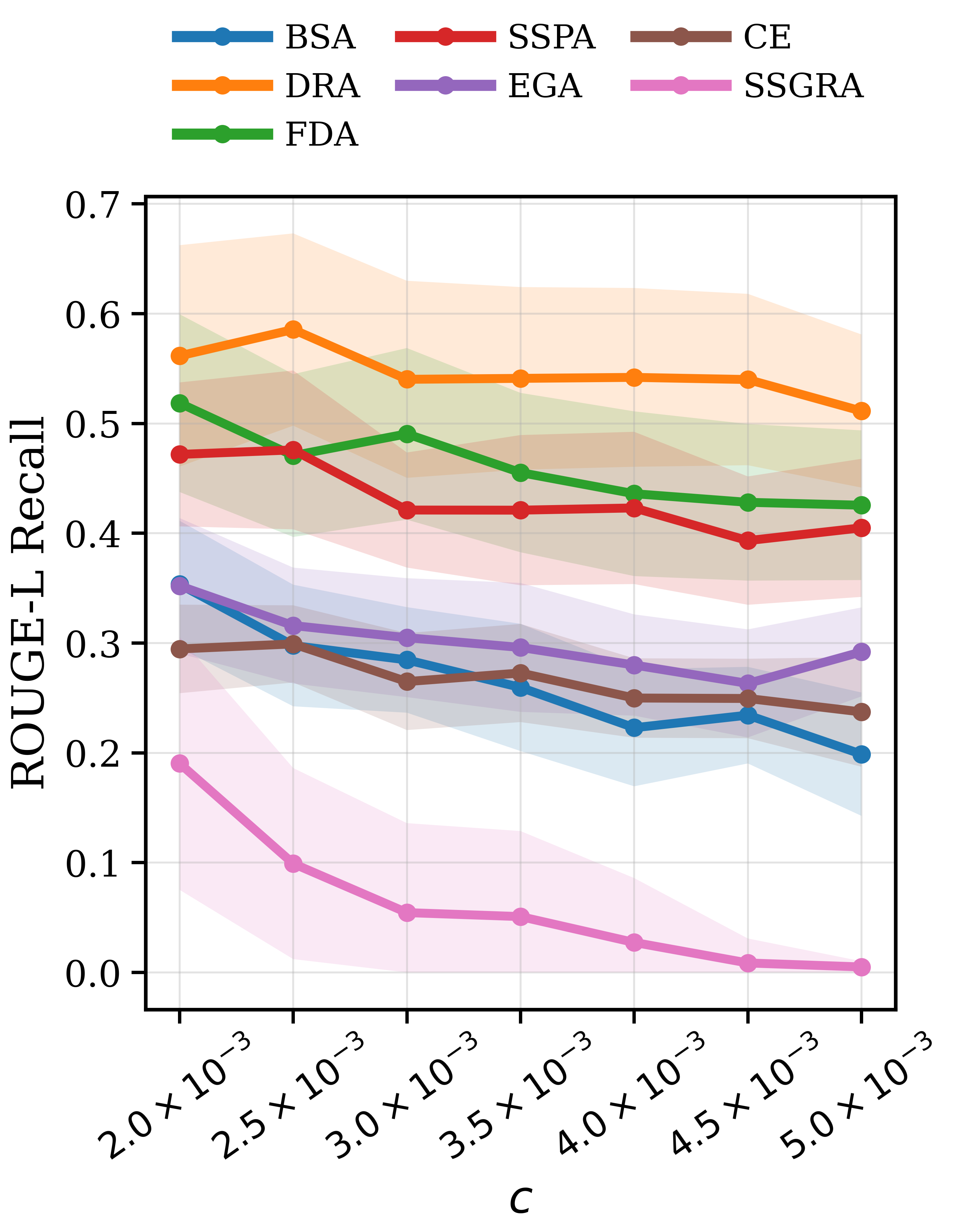}
    \caption{Recall (Qwen2.5-VL)}
\end{subfigure}
\hfill
\begin{subfigure}[t]{0.32\linewidth}
    \centering
    \includegraphics[width=\linewidth, trim=0cm 0cm 0cm 2.4cm, clip]{pictures/qwen2p5/epsilonSeriesRougeL/F1ComparisionSeries_num_steps_1000_.png}
    \caption{F1 (Qwen2.5-VL)}
\end{subfigure}

\vspace{0.6em}


\begin{subfigure}[t]{0.32\linewidth}
    \centering
    \includegraphics[width=\linewidth, trim=0cm 0cm 0cm 2.4cm, clip]{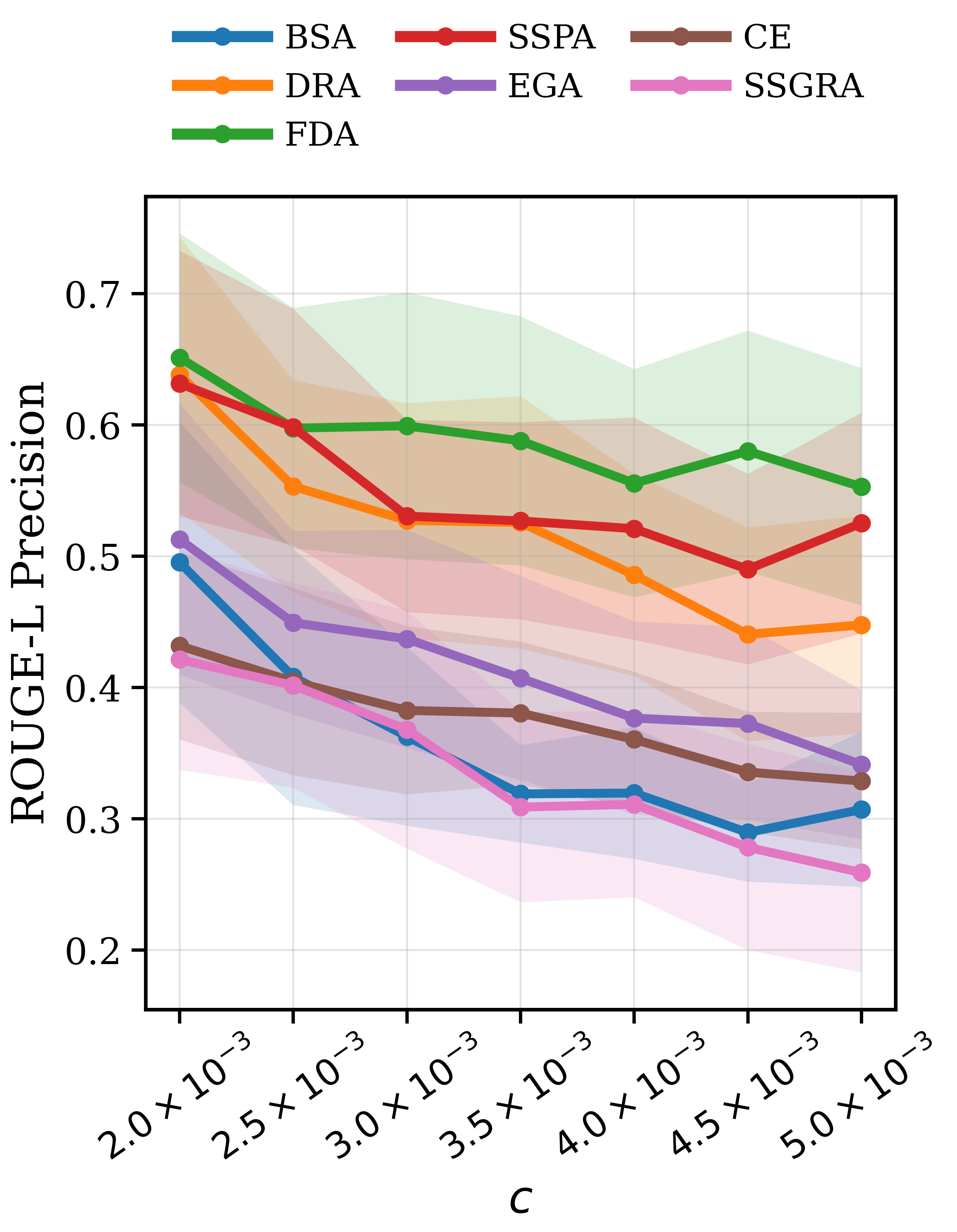}
    \caption{Precision (LLaVa-1.5)}
\end{subfigure}
\hfill
\begin{subfigure}[t]{0.32\linewidth}
    \centering
    \includegraphics[width=\linewidth, trim=0cm 0cm 0cm 2.4cm, clip]{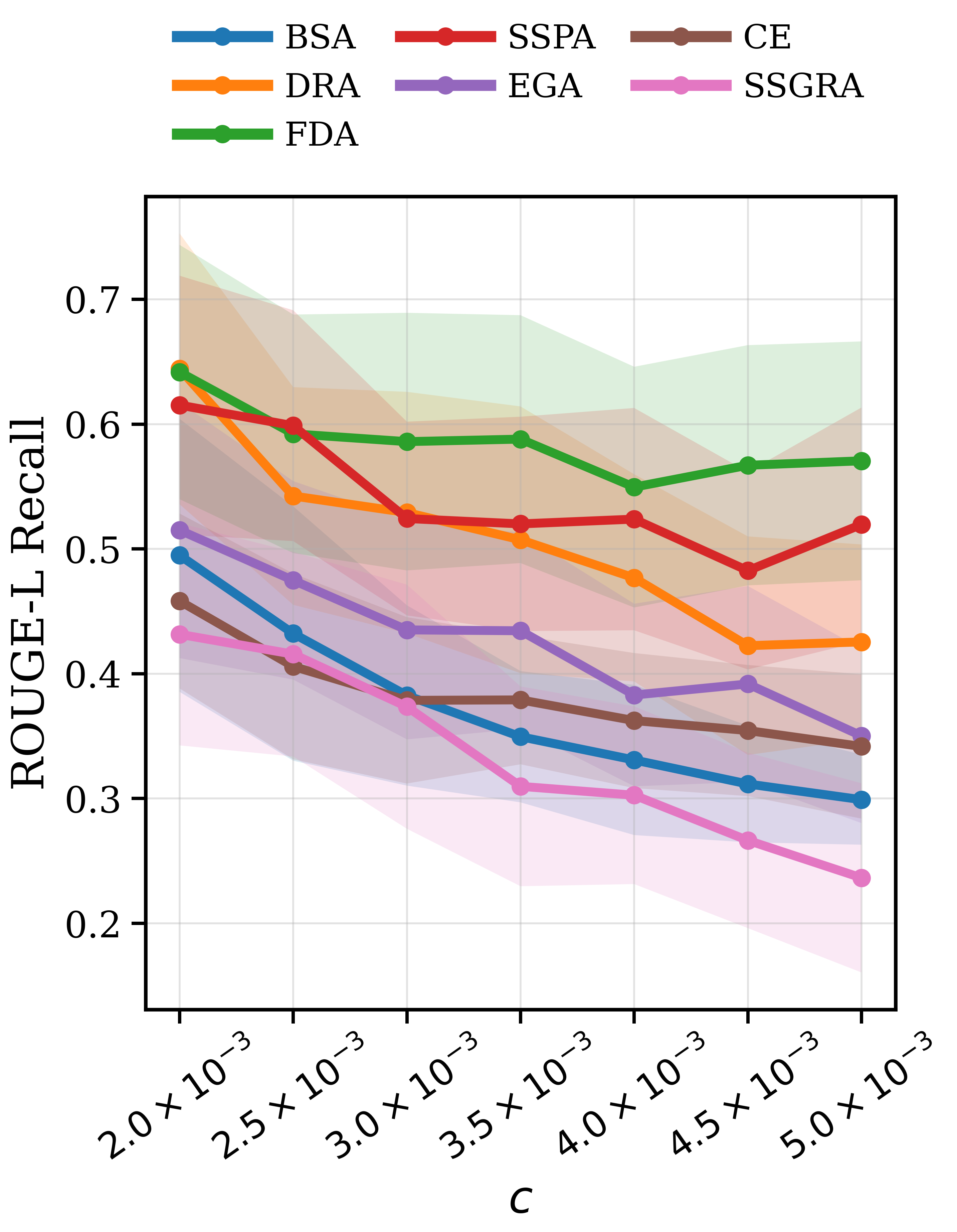}
    \caption{Recall (LLaVa-1.5)}
\end{subfigure}
\hfill
\begin{subfigure}[t]{0.32\linewidth}
    \centering
    \includegraphics[width=\linewidth, trim=0cm 0cm 0cm 2.4cm, clip]{pictures/Llava1p5/epsilonSeriesRougeL1/F1ComparisionSeries_num_steps_1000_.png}
    \caption{F1 (LLaVa-1.5)}
\end{subfigure}

\vspace{0.6em}


\begin{subfigure}[t]{0.32\linewidth}
    \centering
    \includegraphics[width=\linewidth, trim=0cm 0cm 0cm 2.4cm, clip]{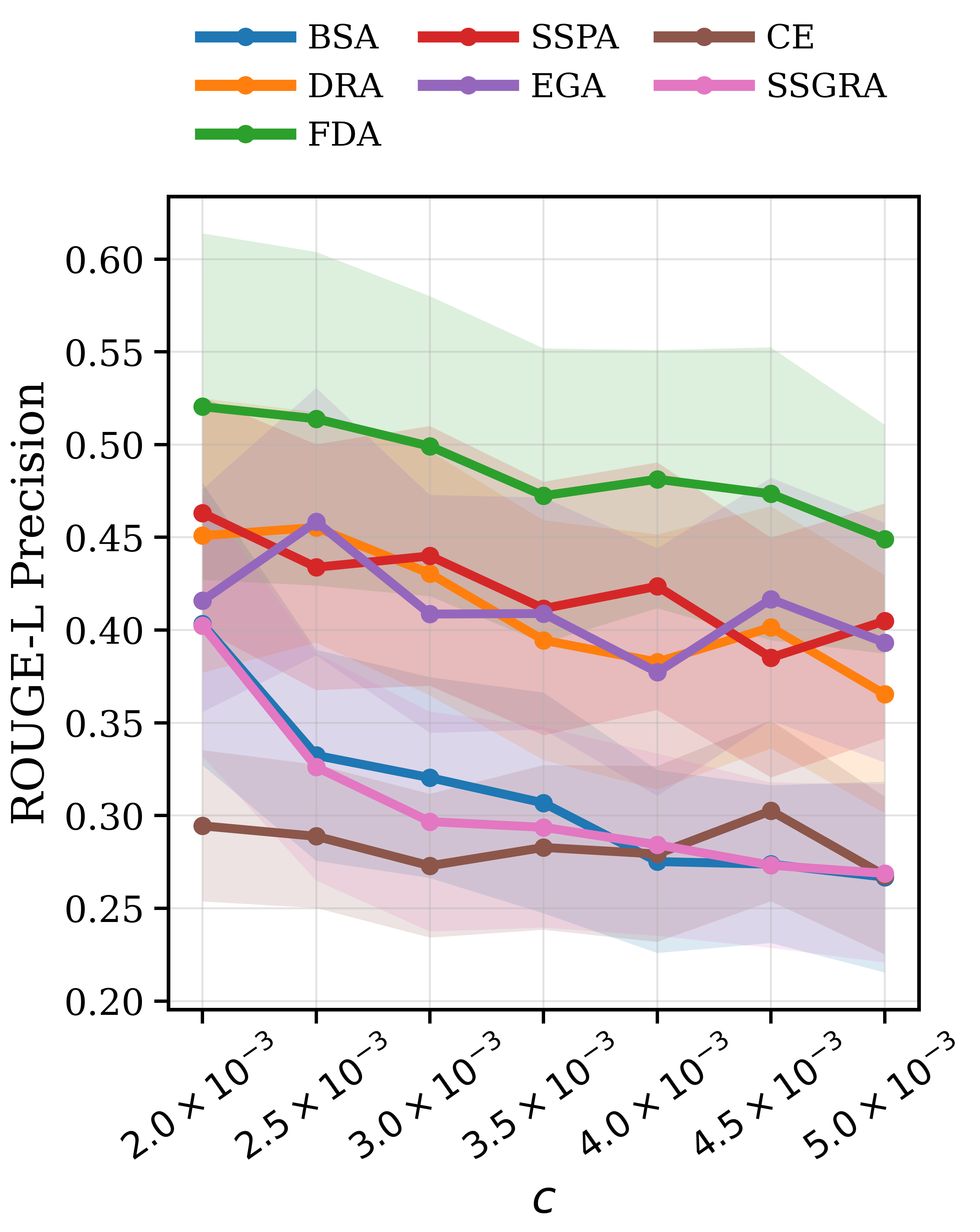}
    \caption{Precision (Gemma 3)}
\end{subfigure}
\hfill
\begin{subfigure}[t]{0.32\linewidth}
    \centering
    \includegraphics[width=\linewidth, trim=0cm 0cm 0cm 2.4cm, clip]{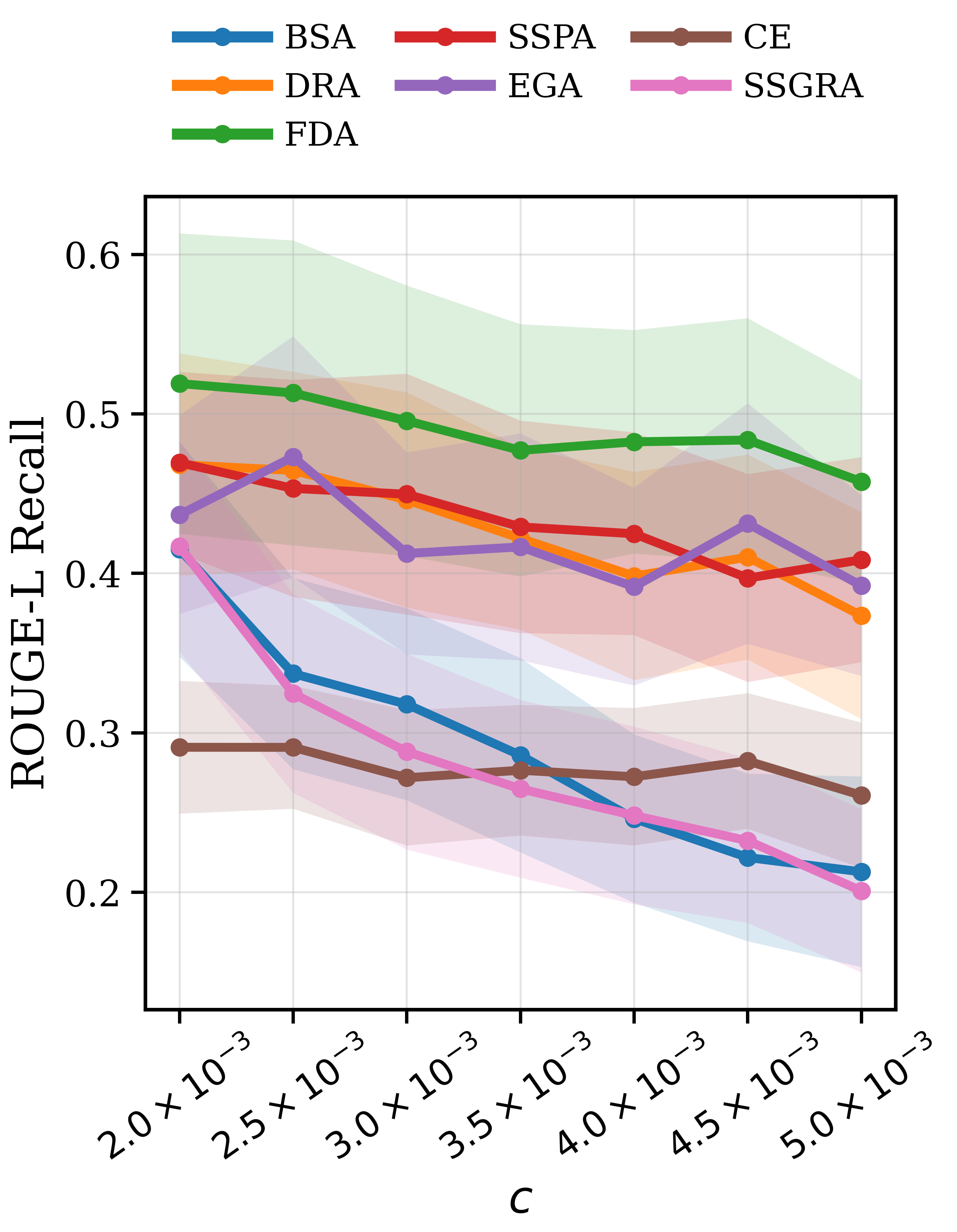}
    \caption{Recall (Gemma 3)}
\end{subfigure}
\hfill
\begin{subfigure}[t]{0.32\linewidth}
    \centering
    \includegraphics[width=\linewidth, trim=0cm 0cm 0cm 2.4cm, clip]{pictures/gemma3/epsilonSeriesRougeL1/F1ComparisonSeries_AllAttacksnum_steps_1000_AttackStartLayer_0_towardsNull_0.1_numSamplesConsidered_50.png}
    \caption{F1 (Gemma 3)}
\end{subfigure}

\caption{ROUGE-L score comparison of sample-specific attacks across Qwen~2.5-VL, LLaVa~1.5, and Gemma~3. The three columns report Precision, Recall, and F1 score, respectively, while each row corresponds to a different vision-language model.}
\label{fig:rougel_comparison_all_models}
\end{figure*}



\subsection{Comprehensive Quantitative Results}

Figures~\ref{fig:bertscore_comparison_all_models} and~\ref{fig:rougel_comparison_all_models} visualize the Precision, Recall, and F1 trends under BERTScore and ROUGE-L across perturbation budgets, while Tables~\ref{tab:Qwen_all_results}, \ref{tab:llava_all_results}, and~\ref{tab:gemma_all_results} report the corresponding numerical results. Across all three VLMs, SSGRA consistently achieves the lowest BERTScore and ROUGE-L scores in most settings, with the largest improvements on Qwen2.5-VL and the smallest on Gemma~3, consistent with the spectral characterization presented in Section \ref{sec:spectralCharacterization}. In addition to the F1 scores reported in Section \ref{sec:quantitativeComparision}, here we provide the corresponding Precision and Recall values, enabling a more detailed analysis of attack behavior. The results show that the improvements achieved by SSGRA are not driven by a single evaluation component but are consistently reflected across all three metrics. Furthermore, the complete numerical results complement the plots by reporting the mean and standard deviation for every perturbation budget, providing a comprehensive view of both attack effectiveness and its variability across the evaluated samples.

\begin{table*}[t]
\centering
\caption{Performance of different attack methods on Qwen2.5-VL under varying perturbation budgets $c$.
Top: BERTScore (Precision, Recall and F1, Mean$\pm$Std).
Bottom: ROUGE-L (Precision, Recall and F1, Mean$\pm$Std).}
\label{tab:Qwen_all_results}

\textbf{(a) BERTScore}

\vspace{2mm}

\resizebox{\textwidth}{!}{
\begin{tabular}{llccccccc}

\toprule
Method & Metric & $c=0.002$ & $c=0.0025$ & $c=0.003$ & $c=0.0035$ & $c=0.004$ & $c=0.0045$ & $c=0.005$ \\
\midrule
\multirow{3}{*}{BSA}
& P  & $0.896\pm0.029$ & $0.879\pm0.037$ & $0.875\pm0.036$ & $0.866\pm0.036$ & $0.866\pm0.032$ & $0.856\pm0.038$ & $0.847\pm0.040$ \\
& R  & $0.894\pm0.029$ & $0.876\pm0.034$ & $0.874\pm0.031$ & $0.862\pm0.039$ & $0.856\pm0.040$ & $0.856\pm0.033$ & $0.843\pm0.038$ \\
& F1 & $0.895\pm0.028$ & $0.877\pm0.034$ & $0.874\pm0.032$ & $0.864\pm0.037$ & $0.861\pm0.035$ & $0.856\pm0.034$ & $0.845\pm0.037$ \\
\midrule
\multirow{3}{*}{DRA}
& P  & $0.937\pm0.026$ & $0.943\pm0.025$ & $0.936\pm0.024$ & $0.935\pm0.024$ & $0.935\pm0.022$ & $0.935\pm0.025$ & $0.931\pm0.020$ \\
& R  & $0.938\pm0.028$ & $0.943\pm0.023$ & $0.935\pm0.024$ & $0.934\pm0.023$ & $0.935\pm0.023$ & $0.936\pm0.022$ & $0.931\pm0.019$ \\
& F1 & $0.938\pm0.026$ & $0.943\pm0.023$ & $0.936\pm0.023$ & $0.934\pm0.023$ & $0.935\pm0.021$ & $0.936\pm0.023$ & $0.931\pm0.018$ \\
\midrule
\multirow{3}{*}{FDA}
& P  & $0.935\pm0.022$ & $0.932\pm0.027$ & $0.931\pm0.024$ & $0.930\pm0.025$ & $0.926\pm0.025$ & $0.922\pm0.023$ & $0.923\pm0.025$ \\
& R  & $0.931\pm0.021$ & $0.924\pm0.025$ & $0.926\pm0.023$ & $0.920\pm0.022$ & $0.918\pm0.023$ & $0.913\pm0.024$ & $0.914\pm0.025$ \\
& F1 & $0.933\pm0.021$ & $0.928\pm0.025$ & $0.929\pm0.023$ & $0.925\pm0.022$ & $0.922\pm0.023$ & $0.918\pm0.022$ & $0.918\pm0.024$ \\
\midrule
\multirow{3}{*}{SSPA}
& P  & $0.932\pm0.019$ & $0.932\pm0.020$ & $0.924\pm0.023$ & $0.929\pm0.025$ & $0.920\pm0.022$ & $0.918\pm0.021$ & $0.922\pm0.020$ \\
& R  & $0.921\pm0.020$ & $0.923\pm0.023$ & $0.915\pm0.020$ & $0.915\pm0.022$ & $0.914\pm0.022$ & $0.909\pm0.021$ & $0.910\pm0.021$ \\
& F1 & $0.926\pm0.018$ & $0.928\pm0.020$ & $0.919\pm0.020$ & $0.922\pm0.022$ & $0.917\pm0.020$ & $0.913\pm0.019$ & $0.916\pm0.019$ \\
\midrule
\multirow{3}{*}{EGA}
& P  & $0.893\pm0.025$ & $0.878\pm0.025$ & $0.878\pm0.028$ & $0.852\pm0.071$ & $0.864\pm0.041$ & $0.841\pm0.084$ & $0.852\pm0.046$ \\
& R  & $0.900\pm0.022$ & $0.889\pm0.025$ & $0.887\pm0.024$ & $0.880\pm0.032$ & $0.882\pm0.026$ & $0.874\pm0.035$ & $0.879\pm0.023$ \\
& F1 & $0.896\pm0.022$ & $0.883\pm0.022$ & $0.883\pm0.024$ & $0.865\pm0.052$ & $0.872\pm0.031$ & $0.856\pm0.061$ & $0.865\pm0.034$ \\
\midrule
\multirow{3}{*}{CE}
& P  & $0.886\pm0.018$ & $0.880\pm0.020$ & $0.870\pm0.033$ & $0.872\pm0.044$ & $0.868\pm0.039$ & $0.858\pm0.046$ & $0.848\pm0.062$ \\
& R  & $0.887\pm0.018$ & $0.888\pm0.018$ & $0.876\pm0.025$ & $0.878\pm0.026$ & $0.874\pm0.024$ & $0.872\pm0.022$ & $0.864\pm0.033$ \\
& F1 & $0.886\pm0.016$ & $0.884\pm0.017$ & $0.873\pm0.028$ & $0.875\pm0.034$ & $0.871\pm0.030$ & $0.865\pm0.034$ & $0.856\pm0.048$ \\
\midrule
\multirow{3}{*}{SSGRA}
& P  & $\mathbf{0.800}\pm0.108$ & $\mathbf{0.748}\pm0.151$ & $\mathbf{0.716}\pm0.138$ & $\mathbf{0.680}\pm0.214$ & $\mathbf{0.726}\pm0.054$ & $\mathbf{0.660}\pm0.200$ & $\mathbf{0.695}\pm0.118$ \\
& R  & $\mathbf{0.835}\pm0.067$ & $\mathbf{0.790}\pm0.144$ & $\mathbf{0.772}\pm0.137$ & $\mathbf{0.730}\pm0.222$ & $\mathbf{0.785}\pm0.033$ & $\mathbf{0.718}\pm0.214$ & $\mathbf{0.756}\pm0.126$ \\
& F1 & $\mathbf{0.816}\pm0.088$ & $\mathbf{0.768}\pm0.147$ & $\mathbf{0.743}\pm0.137$ & $\mathbf{0.704}\pm0.217$ & $\mathbf{0.754}\pm0.042$ & $\mathbf{0.687}\pm0.206$ & $\mathbf{0.724}\pm0.121$ \\
\bottomrule

\end{tabular}
}

\vspace{3mm}

\textbf{(b) ROUGE-L}

\vspace{2mm}

\resizebox{\textwidth}{!}{
\begin{tabular}{llccccccc}

\toprule
Method & Metric & $c=0.002$ & $c=0.0025$ & $c=0.003$ & $c=0.0035$ & $c=0.004$ & $c=0.0045$ & $c=0.005$ \\
\midrule
\multirow{3}{*}{BSA}
& P  & $0.358\pm0.131$ & $0.303\pm0.135$ & $0.292\pm0.108$ & $0.253\pm0.133$ & $0.261\pm0.128$ & $0.231\pm0.108$ & $0.210\pm0.129$ \\
& R  & $0.353\pm0.116$ & $0.298\pm0.111$ & $0.285\pm0.096$ & $0.260\pm0.116$ & $0.223\pm0.107$ & $0.234\pm0.088$ & $0.199\pm0.112$ \\
& F1 & $0.352\pm0.116$ & $0.296\pm0.115$ & $0.285\pm0.097$ & $0.252\pm0.118$ & $0.232\pm0.113$ & $0.228\pm0.091$ & $0.199\pm0.114$ \\
\midrule
\multirow{3}{*}{DRA}
& P  & $0.549\pm0.185$ & $0.579\pm0.185$ & $0.538\pm0.174$ & $0.545\pm0.177$ & $0.522\pm0.150$ & $0.532\pm0.171$ & $0.500\pm0.140$ \\
& R  & $0.562\pm0.202$ & $0.586\pm0.175$ & $0.540\pm0.179$ & $0.541\pm0.166$ & $0.542\pm0.163$ & $0.540\pm0.156$ & $0.511\pm0.139$ \\
& F1 & $0.551\pm0.187$ & $0.577\pm0.171$ & $0.536\pm0.170$ & $0.539\pm0.164$ & $0.528\pm0.149$ & $0.533\pm0.159$ & $0.501\pm0.132$ \\
\midrule
\multirow{3}{*}{FDA}
& P  & $0.519\pm0.163$ & $0.513\pm0.171$ & $0.504\pm0.162$ & $0.497\pm0.166$ & $0.468\pm0.149$ & $0.463\pm0.140$ & $0.452\pm0.159$ \\
& R  & $0.519\pm0.162$ & $0.471\pm0.148$ & $0.491\pm0.156$ & $0.455\pm0.145$ & $0.436\pm0.150$ & $0.428\pm0.143$ & $0.426\pm0.136$ \\
& F1 & $0.517\pm0.160$ & $0.486\pm0.153$ & $0.494\pm0.153$ & $0.471\pm0.148$ & $0.447\pm0.144$ & $0.440\pm0.131$ & $0.434\pm0.139$ \\
\midrule
\multirow{3}{*}{SSPA}
& P  & $0.536\pm0.127$ & $0.524\pm0.146$ & $0.464\pm0.115$ & $0.486\pm0.165$ & $0.447\pm0.127$ & $0.432\pm0.115$ & $0.465\pm0.137$ \\
& R  & $0.472\pm0.131$ & $0.476\pm0.145$ & $0.421\pm0.105$ & $0.421\pm0.137$ & $0.423\pm0.139$ & $0.393\pm0.117$ & $0.405\pm0.126$ \\
& F1 & $0.498\pm0.122$ & $0.494\pm0.136$ & $0.437\pm0.099$ & $0.447\pm0.142$ & $0.428\pm0.119$ & $0.406\pm0.107$ & $0.427\pm0.121$ \\
\midrule
\multirow{3}{*}{EGA}
& P  & $0.327\pm0.111$ & $0.296\pm0.097$ & $0.304\pm0.104$ & $0.300\pm0.135$ & $0.291\pm0.101$ & $0.327\pm0.154$ & $0.271\pm0.070$ \\
& R  & $0.352\pm0.123$ & $0.316\pm0.106$ & $0.305\pm0.108$ & $0.296\pm0.117$ & $0.280\pm0.092$ & $0.263\pm0.098$ & $0.292\pm0.081$ \\
& F1 & $0.337\pm0.113$ & $0.294\pm0.086$ & $0.293\pm0.097$ & $0.277\pm0.100$ & $0.271\pm0.075$ & $0.261\pm0.077$ & $0.274\pm0.065$ \\
\midrule
\multirow{3}{*}{CE}
& P  & $0.278\pm0.073$ & $0.269\pm0.069$ & $0.229\pm0.079$ & $0.249\pm0.081$ & $0.236\pm0.077$ & $0.217\pm0.074$ & $0.212\pm0.084$ \\
& R  & $0.295\pm0.081$ & $0.299\pm0.070$ & $0.265\pm0.088$ & $0.273\pm0.089$ & $0.250\pm0.072$ & $0.250\pm0.072$ & $0.237\pm0.100$ \\
& F1 & $0.282\pm0.066$ & $0.277\pm0.058$ & $0.239\pm0.071$ & $0.256\pm0.076$ & $0.235\pm0.057$ & $0.226\pm0.065$ & $0.217\pm0.081$ \\
\midrule
\multirow{3}{*}{SSGRA}
& P  & $\mathbf{0.218}\pm0.252$ & $\mathbf{0.179}\pm0.281$ & $\mathbf{0.089}\pm0.229$ & $\mathbf{0.086}\pm0.189$ & $\mathbf{0.083}\pm0.246$ & $\mathbf{0.085}\pm0.243$ & $0.205\pm0.372$ \\
& R  & $\mathbf{0.190}\pm0.230$ & $\mathbf{0.099}\pm0.174$ & $\mathbf{0.054}\pm0.163$ & $\mathbf{0.051}\pm0.156$ & $\mathbf{0.027}\pm0.117$ & $\mathbf{0.009}\pm0.045$ & $\mathbf{0.005}\pm0.010$ \\
& F1 & $\mathbf{0.196}\pm0.235$ & $\mathbf{0.102}\pm0.171$ & $\mathbf{0.057}\pm0.164$ & $\mathbf{0.052}\pm0.155$ & $\mathbf{0.028}\pm0.114$ & $\mathbf{0.010}\pm0.047$ & $\mathbf{0.010}\pm0.020$ \\
\bottomrule

\end{tabular}
}

\end{table*}


\begin{table*}[t]
\centering
\caption{Performance of different attack methods on LLaVa-1.5 under varying perturbation budgets $c$.
Top: BERTScore (Precision, Recall and F1, Mean$\pm$Std).
Bottom: ROUGE-L (Precision, Recall and F1, Mean$\pm$Std).}
\label{tab:llava_all_results}

\textbf{(a) BERTScore}

\vspace{2mm}

\resizebox{\textwidth}{!}{
\begin{tabular}{llccccccc}

\toprule
Method & Metric & $c=0.002$ & $c=0.0025$ & $c=0.003$ & $c=0.0035$ & $c=0.004$ & $c=0.0045$ & $c=0.005$ \\
\midrule

\multirow{3}{*}{BSA}
& P  & $0.925\pm0.034$ & $\mathbf{0.909}\pm0.034$ & $0.901\pm0.028$ & $0.893\pm0.019$ & $0.891\pm0.020$ & $0.883\pm0.035$ & $0.885\pm0.019$ \\
& R  & $0.926\pm0.035$ & $0.914\pm0.034$ & $0.904\pm0.029$ & $0.897\pm0.021$ & $0.893\pm0.022$ & $0.888\pm0.022$ & $0.884\pm0.021$ \\
& F1 & $0.926\pm0.034$ & $\mathbf{0.911}\pm0.034$ & $0.902\pm0.028$ & $0.895\pm0.019$ & $0.892\pm0.020$ & $0.886\pm0.028$ & $0.884\pm0.019$ \\
\midrule
\multirow{3}{*}{DRA}
& P  & $0.950\pm0.037$ & $0.940\pm0.022$ & $0.935\pm0.027$ & $0.934\pm0.029$ & $0.926\pm0.026$ & $0.914\pm0.039$ & $0.917\pm0.028$ \\
& R  & $0.953\pm0.029$ & $0.939\pm0.024$ & $0.936\pm0.028$ & $0.931\pm0.031$ & $0.926\pm0.029$ & $0.915\pm0.031$ & $0.914\pm0.028$ \\
& F1 & $0.951\pm0.033$ & $0.940\pm0.022$ & $0.935\pm0.027$ & $0.932\pm0.030$ & $0.926\pm0.027$ & $0.914\pm0.033$ & $0.915\pm0.028$ \\
\midrule
\multirow{3}{*}{FDA}
& P  & $0.954\pm0.027$ & $0.946\pm0.027$ & $0.946\pm0.028$ & $0.945\pm0.027$ & $0.939\pm0.026$ & $0.943\pm0.026$ & $0.938\pm0.025$ \\
& R  & $0.953\pm0.027$ & $0.946\pm0.027$ & $0.945\pm0.028$ & $0.944\pm0.028$ & $0.938\pm0.027$ & $0.942\pm0.027$ & $0.940\pm0.026$ \\
& F1 & $0.954\pm0.027$ & $0.946\pm0.026$ & $0.945\pm0.027$ & $0.944\pm0.027$ & $0.939\pm0.026$ & $0.942\pm0.026$ & $0.939\pm0.025$ \\
\midrule
\multirow{3}{*}{SSPA}
& P  & $0.950\pm0.028$ & $0.947\pm0.025$ & $0.936\pm0.021$ & $0.934\pm0.023$ & $0.932\pm0.026$ & $0.928\pm0.022$ & $0.934\pm0.026$ \\
& R  & $0.949\pm0.027$ & $0.947\pm0.025$ & $0.936\pm0.022$ & $0.934\pm0.025$ & $0.935\pm0.026$ & $0.929\pm0.024$ & $0.933\pm0.027$ \\
& F1 & $0.949\pm0.027$ & $0.947\pm0.024$ & $0.936\pm0.021$ & $0.934\pm0.023$ & $0.933\pm0.026$ & $0.928\pm0.022$ & $0.934\pm0.026$ \\
\midrule
\multirow{3}{*}{EGA}
& P  & $0.933\pm0.032$ & $0.920\pm0.023$ & $0.917\pm0.027$ & $0.910\pm0.028$ & $0.901\pm0.029$ & $0.881\pm0.132$ & $0.868\pm0.137$ \\
& R  & $0.934\pm0.031$ & $0.924\pm0.025$ & $0.919\pm0.027$ & $0.915\pm0.027$ & $0.906\pm0.028$ & $0.890\pm0.131$ & $0.882\pm0.130$ \\
& F1 & $0.933\pm0.031$ & $0.922\pm0.023$ & $0.918\pm0.026$ & $0.913\pm0.027$ & $0.903\pm0.027$ & $0.885\pm0.131$ & $0.875\pm0.133$ \\
\midrule
\multirow{3}{*}{CE}
& P  & $0.919\pm0.024$ & $0.915\pm0.023$ & $0.909\pm0.022$ & $0.907\pm0.019$ & $0.906\pm0.021$ & $0.887\pm0.129$ & $0.880\pm0.130$ \\
& R  & $0.923\pm0.021$ & $0.916\pm0.022$ & $0.912\pm0.020$ & $0.911\pm0.019$ & $0.909\pm0.018$ & $0.889\pm0.130$ & $0.886\pm0.130$ \\
& F1 & $0.921\pm0.022$ & $0.916\pm0.022$ & $0.911\pm0.020$ & $0.909\pm0.018$ & $0.907\pm0.019$ & $0.888\pm0.129$ & $0.883\pm0.130$ \\
\midrule

\multirow{3}{*}{SSGRA}
& P  & $\mathbf{0.915}\pm0.029$ & $0.911\pm0.028$ & $\mathbf{0.879}\pm0.134$ & $\mathbf{0.834}\pm0.217$ & $\mathbf{0.852}\pm0.179$ & $\mathbf{0.815}\pm0.214$ & $\mathbf{0.811}\pm0.216$ \\
& R  & $\mathbf{0.917}\pm0.029$ & $\mathbf{0.913}\pm0.028$ & $\mathbf{0.885}\pm0.133$ & $\mathbf{0.839}\pm0.216$ & $\mathbf{0.855}\pm0.179$ & $\mathbf{0.824}\pm0.213$ & $\mathbf{0.821}\pm0.213$ \\
& F1 & $\mathbf{0.916}\pm0.029$ & $0.912\pm0.027$ & $\mathbf{0.882}\pm0.133$ & $\mathbf{0.837}\pm0.216$ & $\mathbf{0.853}\pm0.179$ & $\mathbf{0.819}\pm0.213$ & $\mathbf{0.816}\pm0.214$ \\

\bottomrule

\end{tabular}
}

\vspace{3mm}

\textbf{(b) ROUGE-L}

\vspace{2mm}

\resizebox{\textwidth}{!}{
\begin{tabular}{llccccccc}

\toprule
Method & Metric & $c=0.002$ & $c=0.0025$ & $c=0.003$ & $c=0.0035$ & $c=0.004$ & $c=0.0045$ & $c=0.005$ \\
\midrule
\multirow{3}{*}{BSA}
& P  & $0.495\pm0.214$ & $0.408\pm0.195$ & $\mathbf{0.363}\pm0.136$ & $0.319\pm0.074$ & $0.320\pm0.100$ & $\mathbf{0.290}\pm0.075$ & $0.307\pm0.118$ \\
& R  & $0.495\pm0.219$ & $0.432\pm0.204$ & $\mathbf{0.383}\pm0.145$ & $0.350\pm0.105$ & $0.331\pm0.120$ & $0.312\pm0.093$ & $0.299\pm0.072$ \\
& F1 & $0.492\pm0.215$ & $0.416\pm0.196$ & $\mathbf{0.369}\pm0.135$ & $0.330\pm0.080$ & $0.320\pm0.103$ & $\mathbf{0.297}\pm0.077$ & $0.291\pm0.065$ \\
\midrule
\multirow{3}{*}{DRA}
& P  & $0.638\pm0.210$ & $0.553\pm0.162$ & $0.527\pm0.179$ & $0.526\pm0.192$ & $0.486\pm0.154$ & $0.441\pm0.163$ & $0.448\pm0.166$ \\
& R  & $0.644\pm0.217$ & $0.542\pm0.174$ & $0.529\pm0.193$ & $0.507\pm0.214$ & $0.477\pm0.166$ & $0.422\pm0.175$ & $0.425\pm0.156$ \\
& F1 & $0.639\pm0.212$ & $0.545\pm0.164$ & $0.525\pm0.183$ & $0.514\pm0.201$ & $0.477\pm0.152$ & $0.426\pm0.164$ & $0.432\pm0.156$ \\
\midrule
\multirow{3}{*}{FDA}
& P  & $0.651\pm0.189$ & $0.598\pm0.183$ & $0.599\pm0.203$ & $0.588\pm0.190$ & $0.556\pm0.174$ & $0.580\pm0.184$ & $0.553\pm0.181$ \\
& R  & $0.642\pm0.204$ & $0.592\pm0.191$ & $0.586\pm0.206$ & $0.588\pm0.199$ & $0.550\pm0.193$ & $0.567\pm0.192$ & $0.571\pm0.191$ \\
& F1 & $0.644\pm0.195$ & $0.591\pm0.182$ & $0.590\pm0.203$ & $0.585\pm0.192$ & $0.550\pm0.182$ & $0.570\pm0.184$ & $0.558\pm0.182$ \\
\midrule
\multirow{3}{*}{SSPA}
& P  & $0.632\pm0.202$ & $0.598\pm0.180$ & $0.531\pm0.146$ & $0.527\pm0.150$ & $0.521\pm0.169$ & $0.490\pm0.145$ & $0.525\pm0.168$ \\
& R  & $0.615\pm0.207$ & $0.599\pm0.185$ & $0.524\pm0.155$ & $0.520\pm0.172$ & $0.524\pm0.178$ & $0.483\pm0.158$ & $0.520\pm0.187$ \\
& F1 & $0.620\pm0.202$ & $0.594\pm0.177$ & $0.524\pm0.144$ & $0.520\pm0.154$ & $0.520\pm0.170$ & $0.483\pm0.147$ & $0.518\pm0.173$ \\
\midrule
\multirow{3}{*}{EGA}
& P  & $0.513\pm0.206$ & $0.449\pm0.140$ & $0.437\pm0.166$ & $0.407\pm0.156$ & $0.377\pm0.147$ & $0.373\pm0.147$ & $0.341\pm0.114$ \\
& R  & $0.515\pm0.205$ & $0.475\pm0.159$ & $0.435\pm0.175$ & $0.434\pm0.157$ & $0.383\pm0.146$ & $0.392\pm0.157$ & $0.350\pm0.140$ \\
& F1 & $0.511\pm0.202$ & $0.456\pm0.138$ & $0.427\pm0.162$ & $0.416\pm0.151$ & $0.371\pm0.138$ & $0.377\pm0.143$ & $0.336\pm0.117$ \\
\midrule
\multirow{3}{*}{CE}
& P  & $0.432\pm0.143$ & $0.405\pm0.143$ & $0.383\pm0.128$ & $0.381\pm0.109$ & $0.360\pm0.103$ & $0.336\pm0.092$ & $0.329\pm0.104$ \\
& R  & $0.458\pm0.140$ & $0.406\pm0.149$ & $0.379\pm0.133$ & $0.379\pm0.103$ & $0.362\pm0.108$ & $0.354\pm0.105$ & $0.342\pm0.116$ \\
& F1 & $0.439\pm0.134$ & $0.400\pm0.137$ & $0.375\pm0.122$ & $0.375\pm0.094$ & $0.357\pm0.096$ & $0.340\pm0.088$ & $0.329\pm0.100$ \\
\midrule
\multirow{3}{*}{SSGRA}
& P  & $\mathbf{0.421}\pm0.168$ & $\mathbf{0.402}\pm0.156$ & $0.368\pm0.182$ & $\mathbf{0.309}\pm0.145$ & $\mathbf{0.311}\pm0.142$ & $0.278\pm0.157$ & $\mathbf{0.259}\pm0.152$ \\
& R  & $\mathbf{0.432}\pm0.178$ & $\mathbf{0.416}\pm0.164$ & $0.374\pm0.196$ & $\mathbf{0.310}\pm0.160$ & $\mathbf{0.303}\pm0.142$ & $\mathbf{0.266}\pm0.141$ & $\mathbf{0.236}\pm0.152$ \\
& F1 & $\mathbf{0.422}\pm0.169$ & $\mathbf{0.406}\pm0.157$ & $0.364\pm0.184$ & $\mathbf{0.299}\pm0.138$ & $\mathbf{0.297}\pm0.132$ & $0.252\pm0.121$ & $\mathbf{0.236}\pm0.138$ \\
\bottomrule

\end{tabular}
}

\end{table*}

\begin{table*}[t]
\centering
\caption{Performance of different attack methods on Gemma~3 under varying perturbation budgets $c$.
Top: BERTScore (Precision, Recall and F1, Mean$\pm$Std).
Bottom: ROUGE-L (Precision, Recall and F1, Mean$\pm$Std).}
\label{tab:gemma_all_results}

\textbf{(a) BERTScore}

\vspace{2mm}

\resizebox{\textwidth}{!}{
\begin{tabular}{llccccccc}
\toprule
Method & Metric & $c=0.002$ & $c=0.0025$ & $c=0.003$ & $c=0.0035$ & $c=0.004$ & $c=0.0045$ & $c=0.005$ \\
\midrule
\multirow{3}{*}{BSA}
& P  & $0.904\pm0.037$ & $0.890\pm0.035$ & $0.879\pm0.037$ & $0.870\pm0.047$ & $0.863\pm0.030$ & $0.860\pm0.027$ & $\mathbf{0.852}\pm0.040$ \\
& R  & $0.907\pm0.036$ & $0.890\pm0.036$ & $0.880\pm0.036$ & $0.871\pm0.040$ & $0.859\pm0.035$ & $0.852\pm0.037$ & $0.843\pm0.039$ \\
& F1 & $0.905\pm0.035$ & $0.890\pm0.034$ & $0.880\pm0.035$ & $0.870\pm0.043$ & $0.861\pm0.031$ & $0.856\pm0.031$ & $\mathbf{0.847}\pm0.037$ \\
\midrule
\multirow{3}{*}{DRA}
& P  & $0.918\pm0.030$ & $0.918\pm0.027$ & $0.915\pm0.030$ & $0.910\pm0.027$ & $0.907\pm0.031$ & $0.909\pm0.029$ & $0.903\pm0.033$ \\
& R  & $0.920\pm0.028$ & $0.918\pm0.027$ & $0.914\pm0.032$ & $0.912\pm0.025$ & $0.909\pm0.029$ & $0.909\pm0.030$ & $0.904\pm0.031$ \\
& F1 & $0.919\pm0.027$ & $0.918\pm0.025$ & $0.914\pm0.029$ & $0.911\pm0.025$ & $0.908\pm0.029$ & $0.909\pm0.028$ & $0.903\pm0.030$ \\
\midrule
\multirow{3}{*}{FDA}
& P  & $0.928\pm0.037$ & $0.925\pm0.036$ & $0.924\pm0.032$ & $0.920\pm0.034$ & $0.923\pm0.031$ & $0.921\pm0.030$ & $0.918\pm0.027$ \\
& R  & $0.926\pm0.038$ & $0.924\pm0.040$ & $0.923\pm0.035$ & $0.919\pm0.037$ & $0.922\pm0.034$ & $0.921\pm0.032$ & $0.918\pm0.028$ \\
& F1 & $0.927\pm0.036$ & $0.924\pm0.037$ & $0.923\pm0.032$ & $0.919\pm0.034$ & $0.923\pm0.032$ & $0.921\pm0.030$ & $0.918\pm0.026$ \\
\midrule
\multirow{3}{*}{SSPA}
& P  & $0.924\pm0.024$ & $0.917\pm0.028$ & $0.919\pm0.030$ & $0.913\pm0.030$ & $0.915\pm0.030$ & $0.907\pm0.027$ & $0.911\pm0.030$ \\
& R  & $0.923\pm0.024$ & $0.919\pm0.030$ & $0.915\pm0.037$ & $0.913\pm0.030$ & $0.912\pm0.033$ & $0.905\pm0.031$ & $0.907\pm0.035$ \\
& F1 & $0.923\pm0.022$ & $0.918\pm0.028$ & $0.916\pm0.032$ & $0.913\pm0.029$ & $0.913\pm0.031$ & $0.905\pm0.027$ & $0.909\pm0.031$ \\
\midrule
\multirow{3}{*}{EGA}
& P  & $0.910\pm0.030$ & $0.919\pm0.030$ & $0.907\pm0.031$ & $0.906\pm0.030$ & $0.900\pm0.031$ & $0.906\pm0.031$ & $0.901\pm0.031$ \\
& R  & $0.913\pm0.030$ & $0.918\pm0.034$ & $0.910\pm0.032$ & $0.908\pm0.033$ & $0.902\pm0.034$ & $0.909\pm0.032$ & $0.904\pm0.031$ \\
& F1 & $0.911\pm0.029$ & $0.918\pm0.031$ & $0.908\pm0.030$ & $0.907\pm0.031$ & $0.901\pm0.031$ & $0.908\pm0.030$ & $0.902\pm0.030$ \\
\midrule
\multirow{3}{*}{CE}
& P  & $\mathbf{0.886}\pm0.027$ & $\mathbf{0.882}\pm0.026$ & $0.879\pm0.031$ & $0.877\pm0.029$ & $0.876\pm0.038$ & $0.881\pm0.039$ & $0.864\pm0.052$ \\
& R  & $\mathbf{0.887}\pm0.028$ & $\mathbf{0.883}\pm0.028$ & $0.879\pm0.032$ & $0.881\pm0.027$ & $0.878\pm0.031$ & $0.881\pm0.032$ & $0.872\pm0.034$ \\
& F1 & $\mathbf{0.886}\pm0.026$ & $\mathbf{0.883}\pm0.025$ & $0.879\pm0.030$ & $0.879\pm0.026$ & $0.877\pm0.033$ & $0.881\pm0.034$ & $0.867\pm0.041$ \\
\midrule
\multirow{3}{*}{SSGRA}
& P  & $0.902\pm0.038$ & $0.883\pm0.046$ & $\mathbf{0.874}\pm0.036$ & $\mathbf{0.869}\pm0.031$ & $\mathbf{0.860}\pm0.035$ & $\mathbf{0.857}\pm0.039$ & $0.855\pm0.026$ \\
& R  & $0.903\pm0.039$ & $\mathbf{0.883}\pm0.040$ & $\mathbf{0.873}\pm0.039$ & $\mathbf{0.867}\pm0.036$ & $\mathbf{0.856}\pm0.038$ & $\mathbf{0.853}\pm0.040$ & $\mathbf{0.842}\pm0.037$ \\
& F1 & $0.903\pm0.038$ & $\mathbf{0.883}\pm0.042$ & $\mathbf{0.873}\pm0.036$ & $\mathbf{0.868}\pm0.032$ & $\mathbf{0.858}\pm0.034$ & $\mathbf{0.855}\pm0.037$ & $0.849\pm0.030$ \\
\bottomrule
\end{tabular}
}

\vspace{3mm}

\textbf{(b) ROUGE-L}

\vspace{2mm}

\resizebox{\textwidth}{!}{
\begin{tabular}{llccccccc}
\toprule
Method & Metric & $c=0.002$ & $c=0.0025$ & $c=0.003$ & $c=0.0035$ & $c=0.004$ & $c=0.0045$ & $c=0.005$ \\
\midrule
\multirow{3}{*}{BSA}
& P  & $0.403\pm0.152$ & $0.332\pm0.114$ & $0.320\pm0.108$ & $0.307\pm0.119$ & $\mathbf{0.275}\pm0.099$ & $0.274\pm0.085$ & $\mathbf{0.267}\pm0.103$ \\
& R  & $0.415\pm0.135$ & $0.337\pm0.120$ & $0.318\pm0.120$ & $0.286\pm0.122$ & $\mathbf{0.246}\pm0.106$ & $\mathbf{0.222}\pm0.105$ & $0.213\pm0.119$ \\
& F1 & $0.406\pm0.142$ & $0.331\pm0.116$ & $0.315\pm0.112$ & $0.287\pm0.116$ & $\mathbf{0.253}\pm0.098$ & $\mathbf{0.238}\pm0.092$ & $0.228\pm0.109$ \\
\midrule
\multirow{3}{*}{DRA}
& P  & $0.451\pm0.148$ & $0.455\pm0.124$ & $0.430\pm0.132$ & $0.395\pm0.129$ & $0.383\pm0.138$ & $0.401\pm0.131$ & $0.365\pm0.128$ \\
& R  & $0.468\pm0.139$ & $0.465\pm0.124$ & $0.446\pm0.135$ & $0.422\pm0.115$ & $0.398\pm0.131$ & $0.410\pm0.129$ & $0.373\pm0.130$ \\
& F1 & $0.456\pm0.140$ & $0.457\pm0.121$ & $0.433\pm0.131$ & $0.405\pm0.120$ & $0.388\pm0.134$ & $0.402\pm0.126$ & $0.365\pm0.125$ \\
\midrule
\multirow{3}{*}{FDA}
& P  & $0.520\pm0.187$ & $0.514\pm0.180$ & $0.499\pm0.162$ & $0.472\pm0.159$ & $0.481\pm0.139$ & $0.473\pm0.158$ & $0.449\pm0.123$ \\
& R  & $0.519\pm0.188$ & $0.513\pm0.191$ & $0.496\pm0.170$ & $0.477\pm0.158$ & $0.482\pm0.140$ & $0.484\pm0.153$ & $0.457\pm0.127$ \\
& F1 & $0.516\pm0.186$ & $0.509\pm0.186$ & $0.492\pm0.166$ & $0.471\pm0.157$ & $0.478\pm0.139$ & $0.474\pm0.152$ & $0.450\pm0.121$ \\
\midrule
\multirow{3}{*}{SSPA}
& P  & $0.463\pm0.125$ & $0.434\pm0.133$ & $0.440\pm0.140$ & $0.412\pm0.137$ & $0.424\pm0.133$ & $0.385\pm0.129$ & $0.405\pm0.127$ \\
& R  & $0.469\pm0.114$ & $0.453\pm0.136$ & $0.450\pm0.151$ & $0.429\pm0.133$ & $0.425\pm0.127$ & $0.397\pm0.130$ & $0.409\pm0.129$ \\
& F1 & $0.464\pm0.118$ & $0.441\pm0.131$ & $0.440\pm0.141$ & $0.418\pm0.134$ & $0.422\pm0.129$ & $0.388\pm0.128$ & $0.404\pm0.126$ \\
\midrule
\multirow{3}{*}{EGA}
& P  & $0.416\pm0.120$ & $0.458\pm0.144$ & $0.409\pm0.128$ & $0.409\pm0.125$ & $0.377\pm0.133$ & $0.417\pm0.131$ & $0.393\pm0.129$ \\
& R  & $0.437\pm0.125$ & $0.473\pm0.151$ & $0.412\pm0.127$ & $0.417\pm0.142$ & $0.392\pm0.124$ & $0.431\pm0.151$ & $0.392\pm0.113$ \\
& F1 & $0.423\pm0.118$ & $0.463\pm0.144$ & $0.408\pm0.125$ & $0.410\pm0.128$ & $0.382\pm0.127$ & $0.420\pm0.131$ & $0.390\pm0.119$ \\
\midrule
\multirow{3}{*}{CE}
& P  & $\mathbf{0.294}\pm0.082$ & $\mathbf{0.289}\pm0.077$ & $\mathbf{0.273}\pm0.077$ & $\mathbf{0.283}\pm0.089$ & $0.279\pm0.095$ & $0.303\pm0.098$ & $0.268\pm0.085$ \\
& R  & $\mathbf{0.291}\pm0.083$ & $\mathbf{0.291}\pm0.077$ & $\mathbf{0.272}\pm0.085$ & $0.277\pm0.082$ & $0.272\pm0.086$ & $0.282\pm0.085$ & $0.261\pm0.091$ \\
& F1 & $\mathbf{0.289}\pm0.080$ & $\mathbf{0.285}\pm0.071$ & $\mathbf{0.269}\pm0.079$ & $0.275\pm0.082$ & $0.273\pm0.088$ & $0.285\pm0.081$ & $0.259\pm0.085$ \\
\midrule
\multirow{3}{*}{SSGRA}
& P  & $0.402\pm0.141$ & $0.326\pm0.122$ & $0.297\pm0.119$ & $0.294\pm0.108$ & $0.284\pm0.098$ & $\mathbf{0.273}\pm0.089$ & $0.269\pm0.096$ \\
& R  & $0.417\pm0.132$ & $0.325\pm0.125$ & $0.288\pm0.123$ & $\mathbf{0.265}\pm0.112$ & $0.248\pm0.111$ & $0.232\pm0.103$ & $\mathbf{0.201}\pm0.102$ \\
& F1 & $0.407\pm0.136$ & $0.319\pm0.118$ & $0.288\pm0.119$ & $\mathbf{0.272}\pm0.105$ & $0.259\pm0.104$ & $0.241\pm0.088$ & $\mathbf{0.221}\pm0.093$ \\
\bottomrule
\end{tabular}
}

\end{table*}

\subsection{Additional Qualitative Results}
\label{sec:additionalQualitative}

\begin{figure}[t]
    \centering
    \begin{subfigure}{\linewidth}
        \centering
        \includegraphics[width=\linewidth]{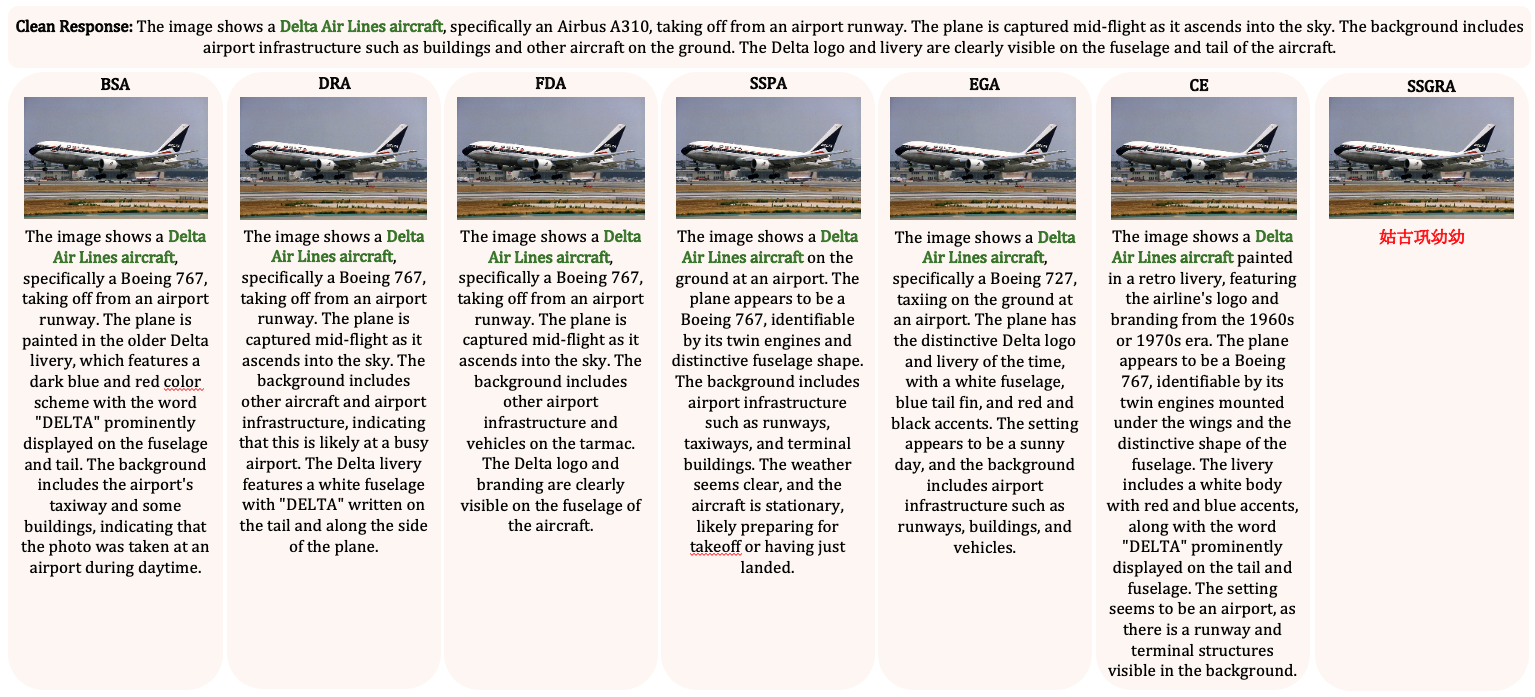}
        \caption{Qwen2.5-VL}
        \label{fig:qual_qwenA1}
    \end{subfigure}
    \begin{subfigure}{\linewidth}
        \centering
        \includegraphics[width=\linewidth]{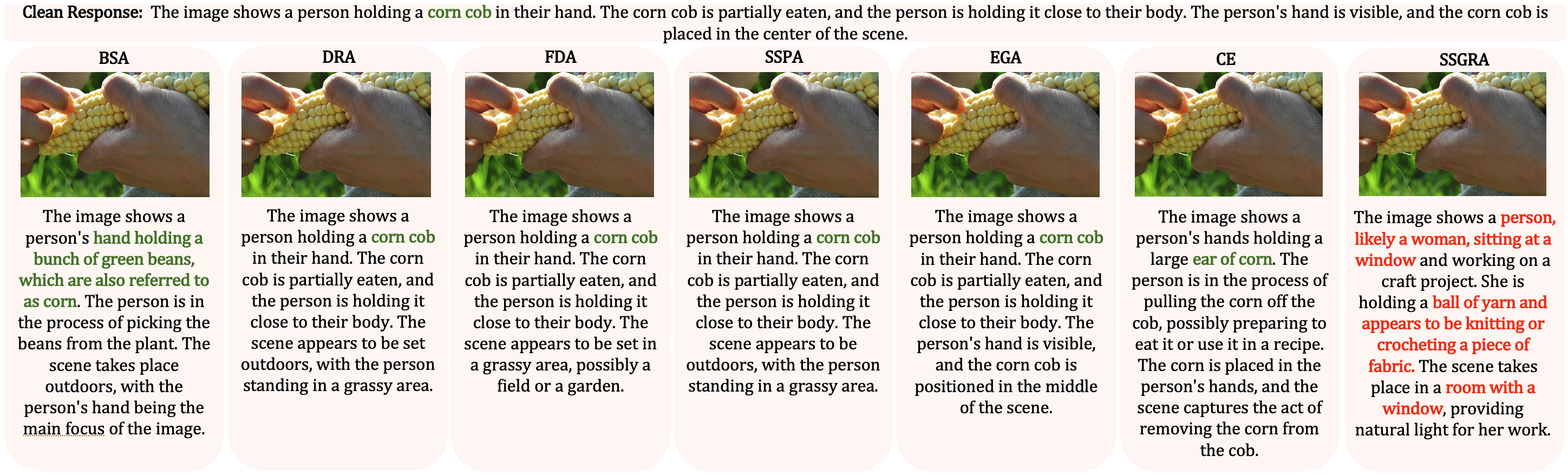}
        \caption{LLaVa-1.5}
        \label{fig:qual_llavaA1}
    \end{subfigure}
    \begin{subfigure}{\linewidth}
        \centering
        \includegraphics[width=\linewidth]{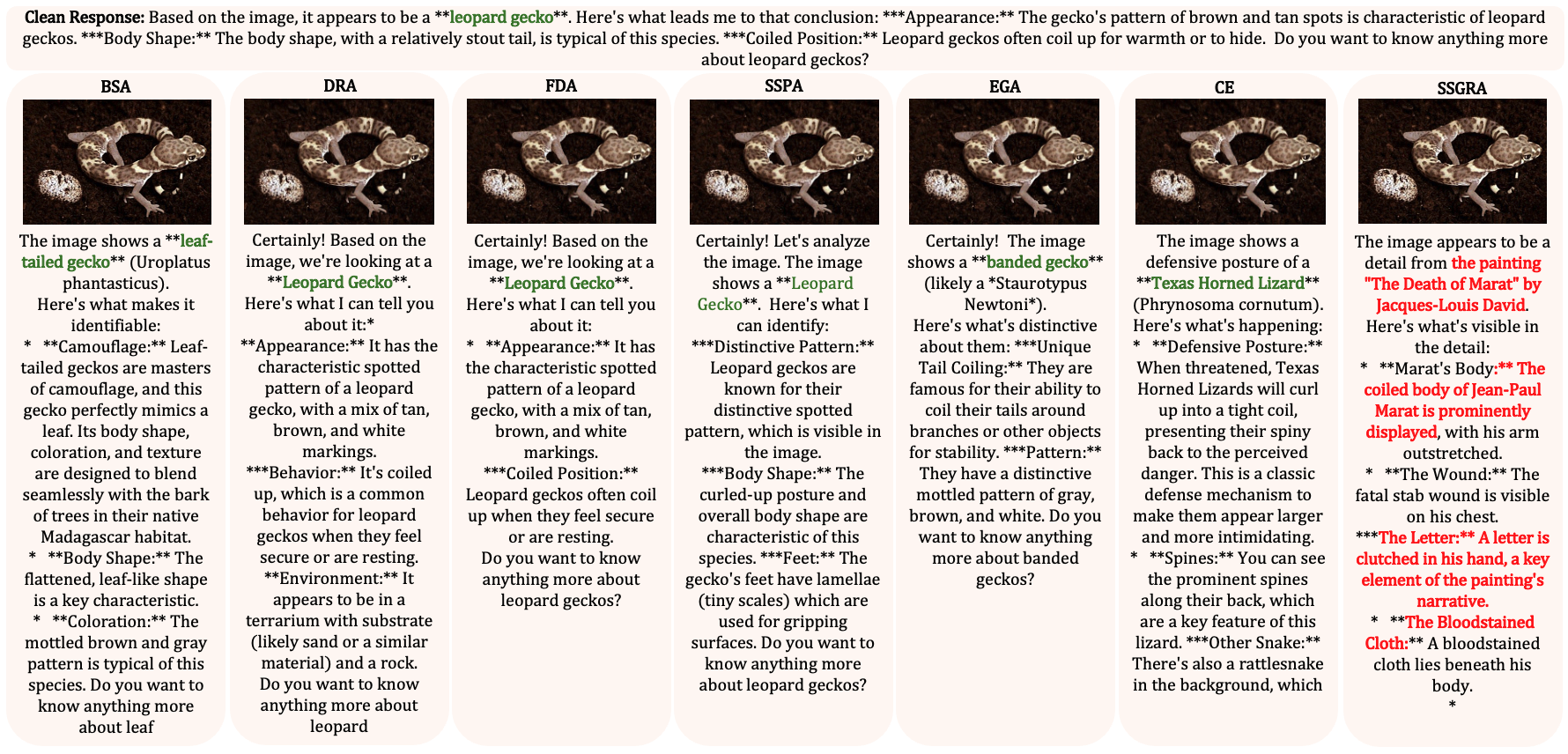}
        \caption{Gemma 3}
        \label{fig:qual_gemmaA1}
    \end{subfigure}
    \caption{Additional qualitative adversarial examples (set 1) generated with a perturbation budget of $c=0.002$ across models.}
    \label{fig:AdditionalQual1}
\end{figure}

\begin{figure}[t]
    \centering
    \begin{subfigure}{\linewidth}
        \centering
        \includegraphics[width=\linewidth]{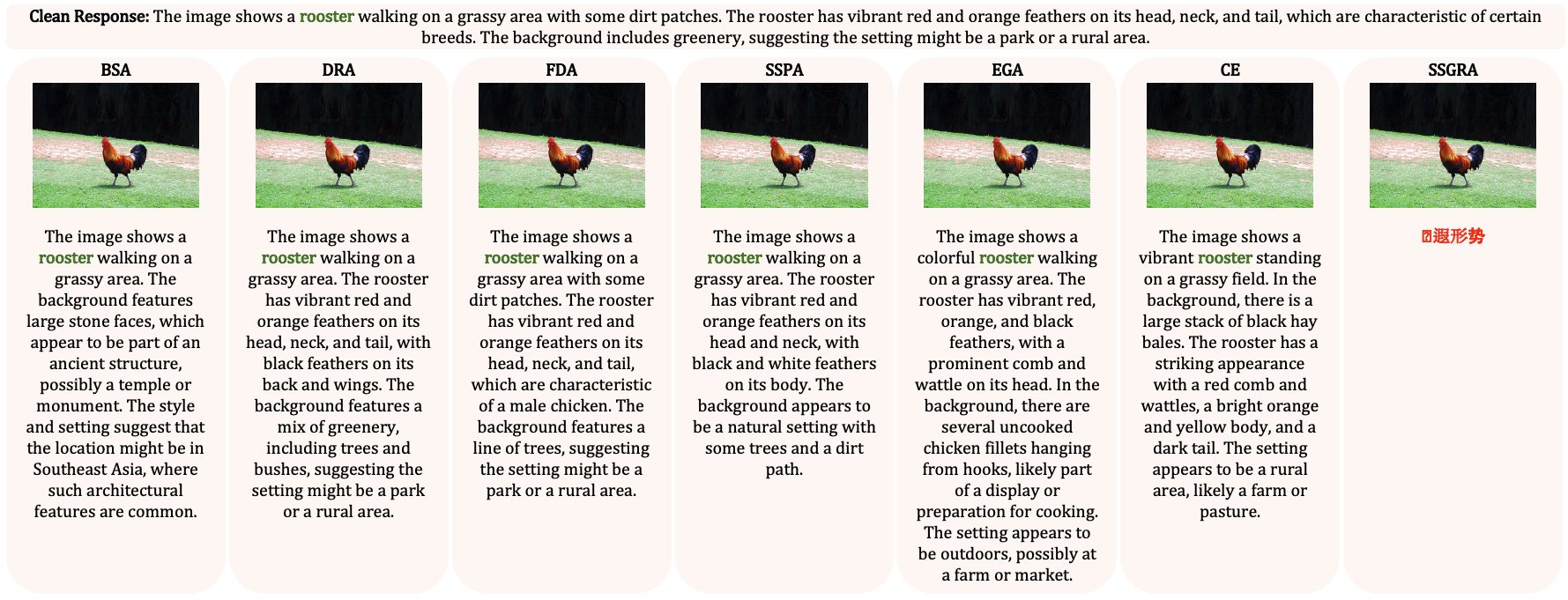}
        \caption{Qwen2.5-VL}
        \label{fig:qual_qwenA2}
    \end{subfigure}
    \begin{subfigure}{\linewidth}
        \centering
        \includegraphics[width=\linewidth]{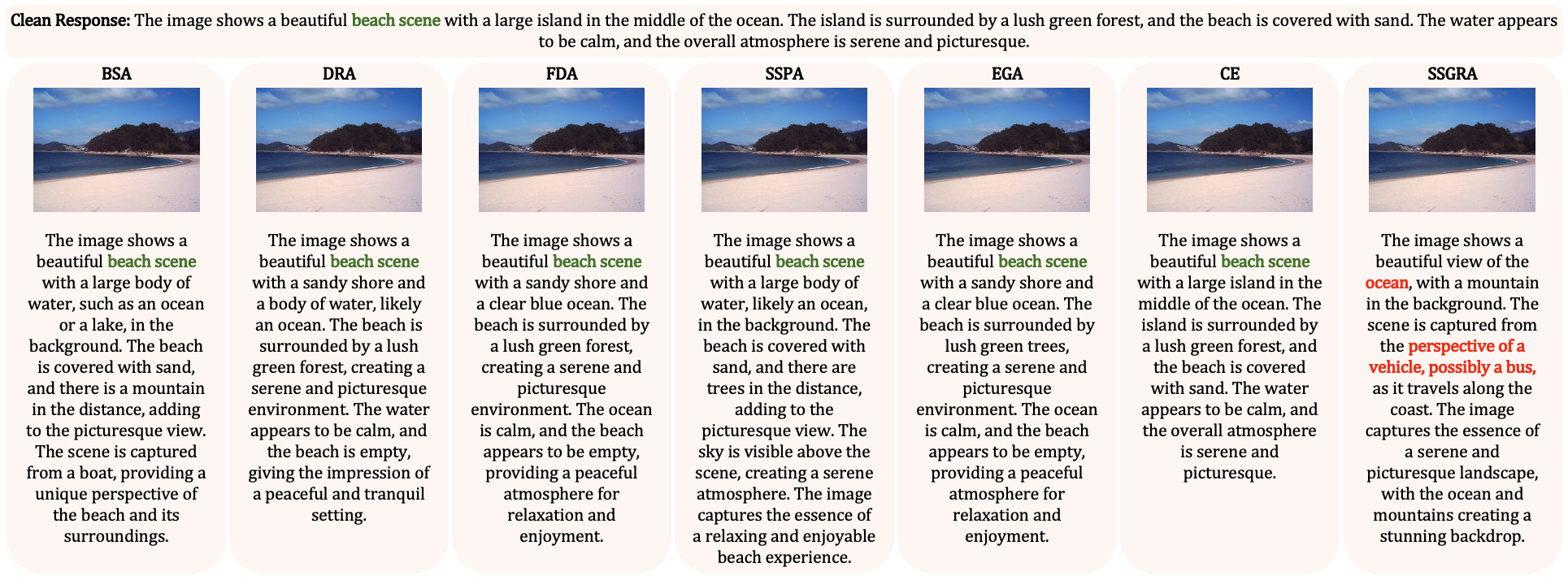}
        \caption{LLaVA-1.5}
        \label{fig:qual_llavaA2}
    \end{subfigure}
    \begin{subfigure}{\linewidth}
        \centering
        \includegraphics[width=\linewidth]{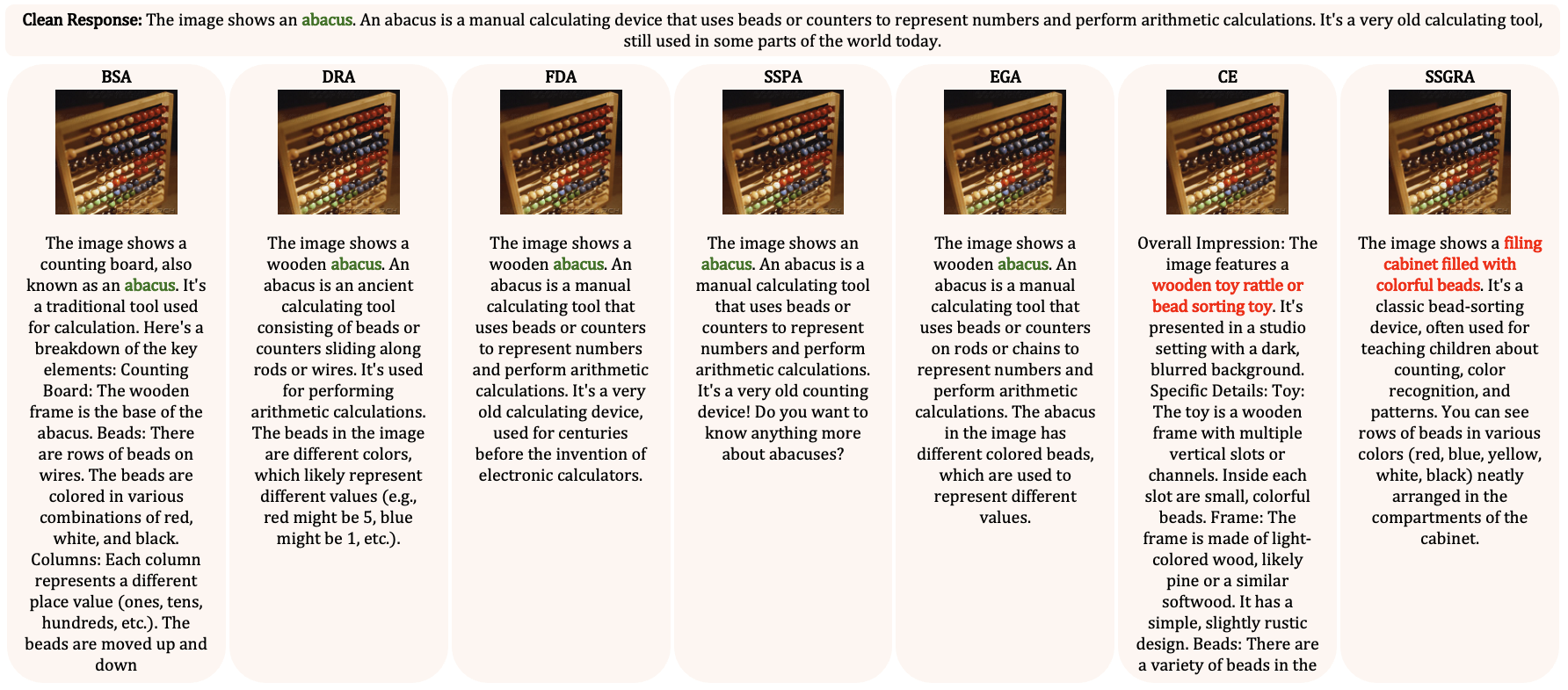}
        \caption{Gemma 3}
        \label{fig:qual_gemmaA2}
    \end{subfigure}
    \caption{Additional qualitative adversarial examples (set 2) generated with a perturbation budget of $c=0.002$ across models.}
    \label{fig:AdditionalQual2}
\end{figure}

\begin{figure}[t]
    \centering
    \begin{subfigure}{\linewidth}
        \centering
        \includegraphics[width=\linewidth]{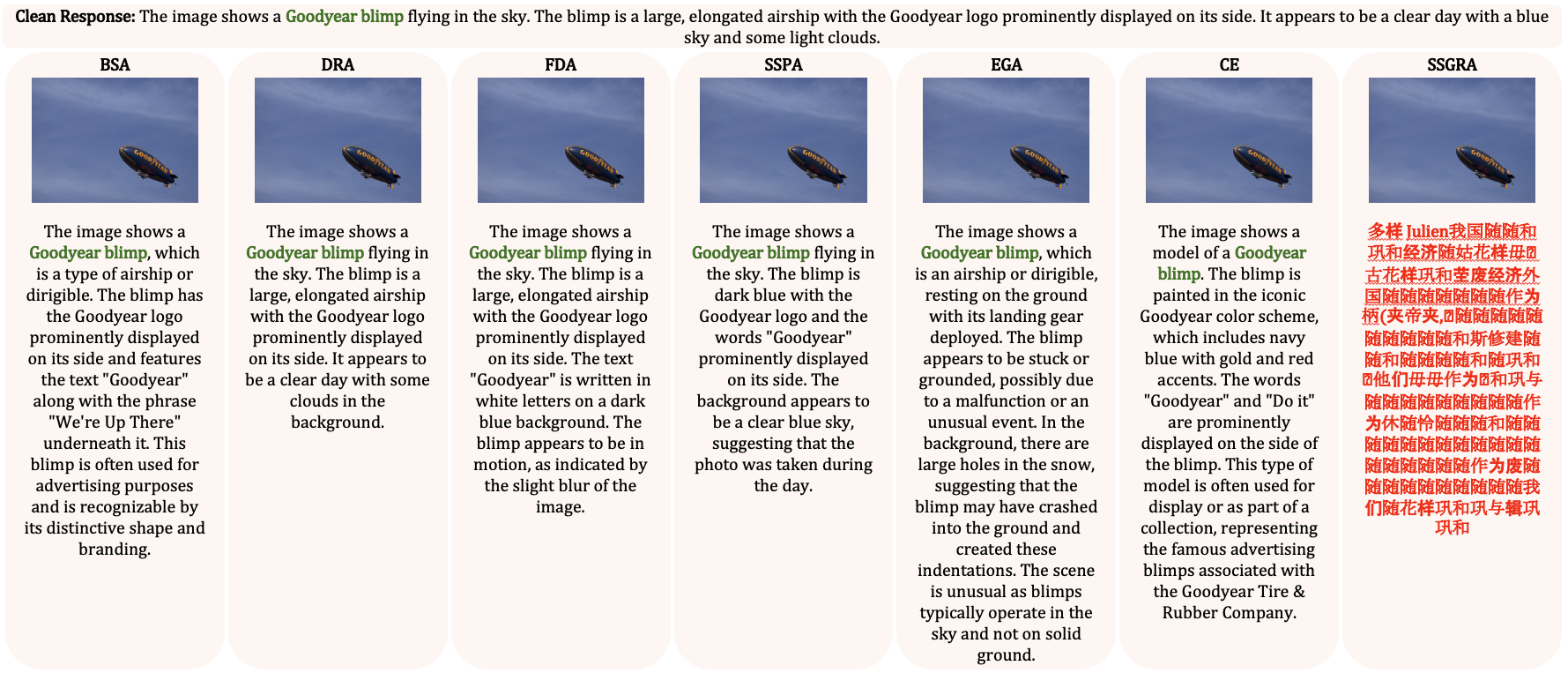}
        \caption{Qwen2.5-VL}
        \label{fig:qual_qwenA3}
    \end{subfigure}
    \begin{subfigure}{\linewidth}
        \centering
        \includegraphics[width=\linewidth]{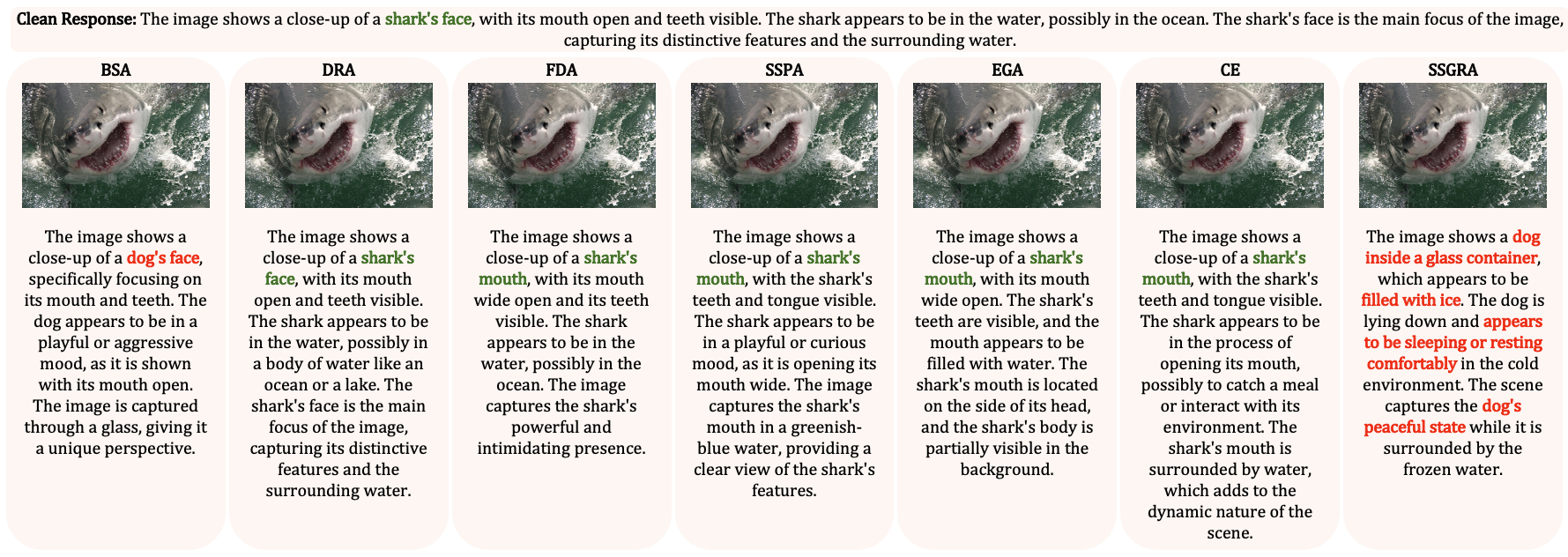}
        \caption{LLaVA-1.5}
        \label{fig:qual_llavaA3}
    \end{subfigure}
    \begin{subfigure}{\linewidth}
        \centering
        \includegraphics[width=\linewidth]{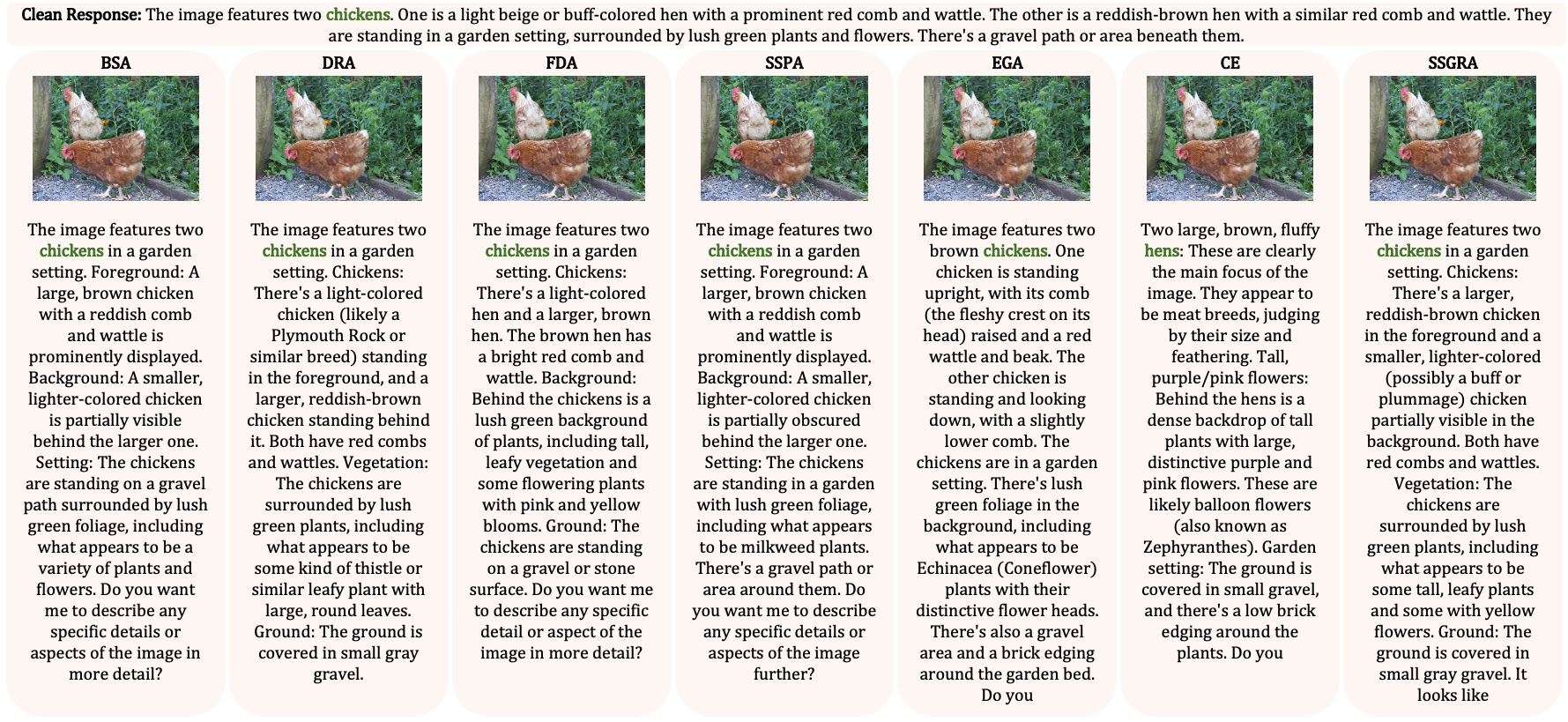}
        \caption{Gemma 3}
        \label{fig:qual_gemmaA3}
    \end{subfigure}
    \caption{Additional qualitative adversarial examples (set 3) generated with a perturbation budget of $c=0.002$ across models.}
    \label{fig:AdditionalQual3}
\end{figure}

\begin{figure}[t]
    \centering
    \begin{subfigure}{\linewidth}
        \centering
        \includegraphics[width=0.94\linewidth]{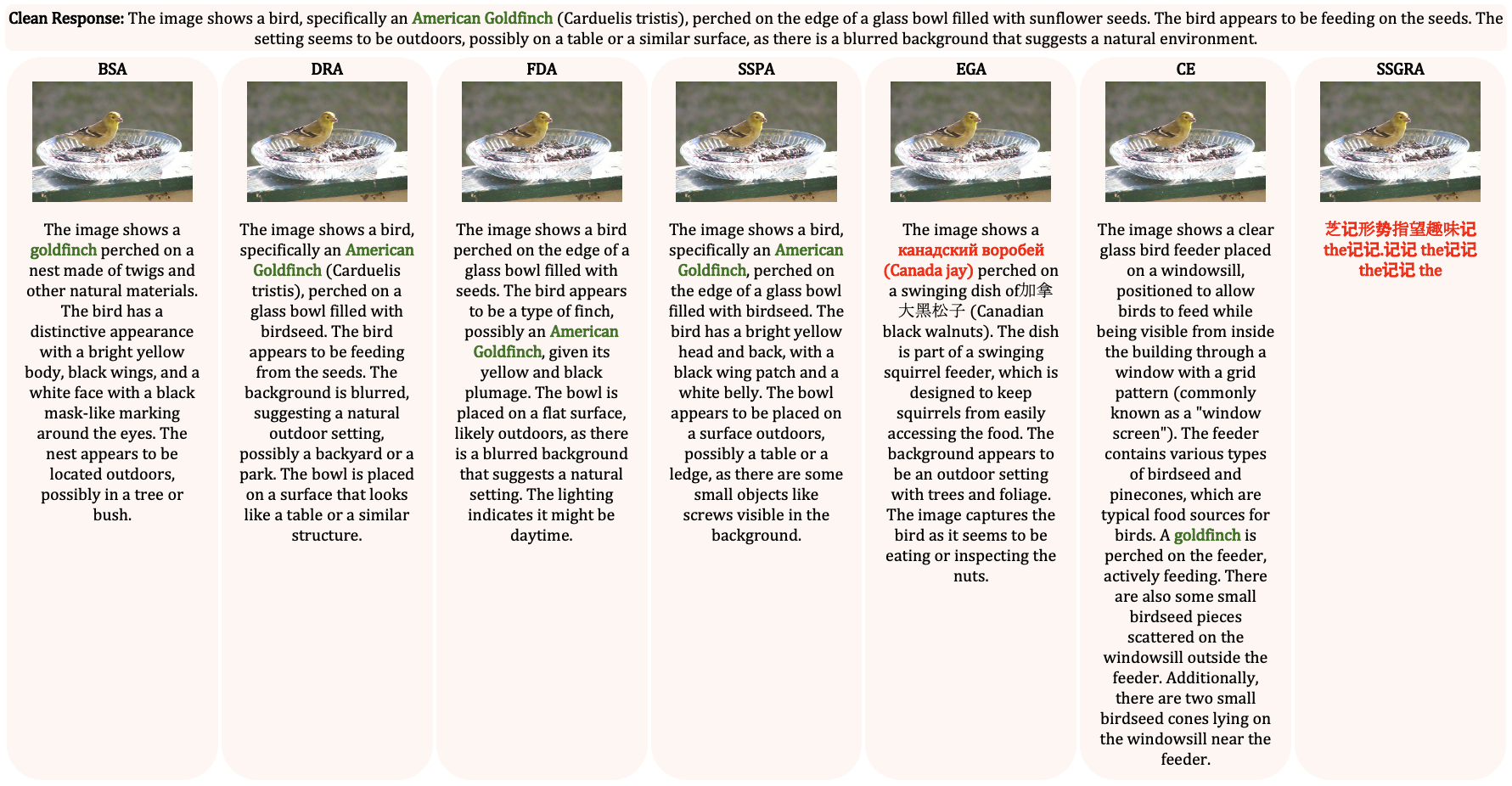}
        \caption{Qwen2.5-VL}
        \label{fig:qual_qwenA4}
    \end{subfigure}
    \begin{subfigure}{\linewidth}
        \centering
        \includegraphics[width=0.94\linewidth]{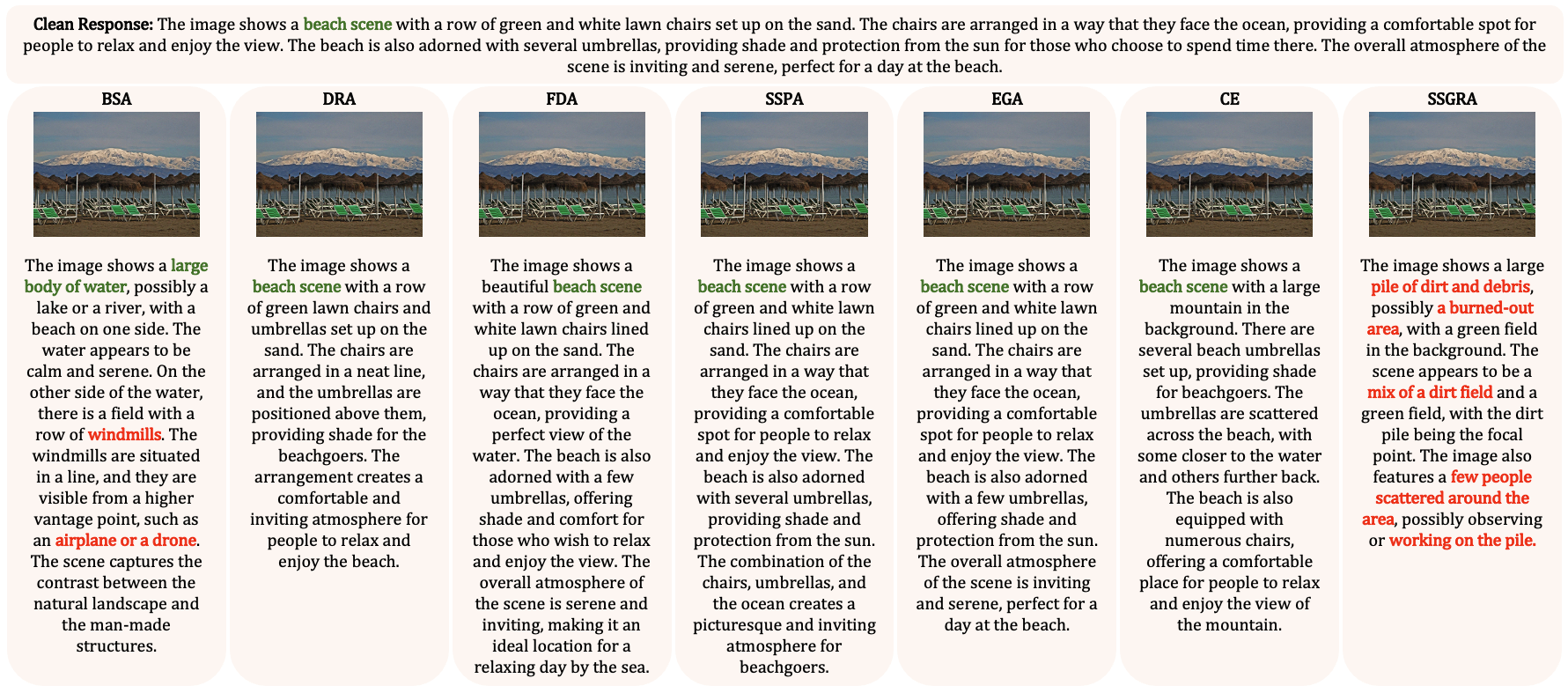}
        \caption{LLaVA-1.5}
        \label{fig:qual_llavaA4}
    \end{subfigure}
    \begin{subfigure}{\linewidth}
        \centering
        \includegraphics[width=0.94\linewidth]{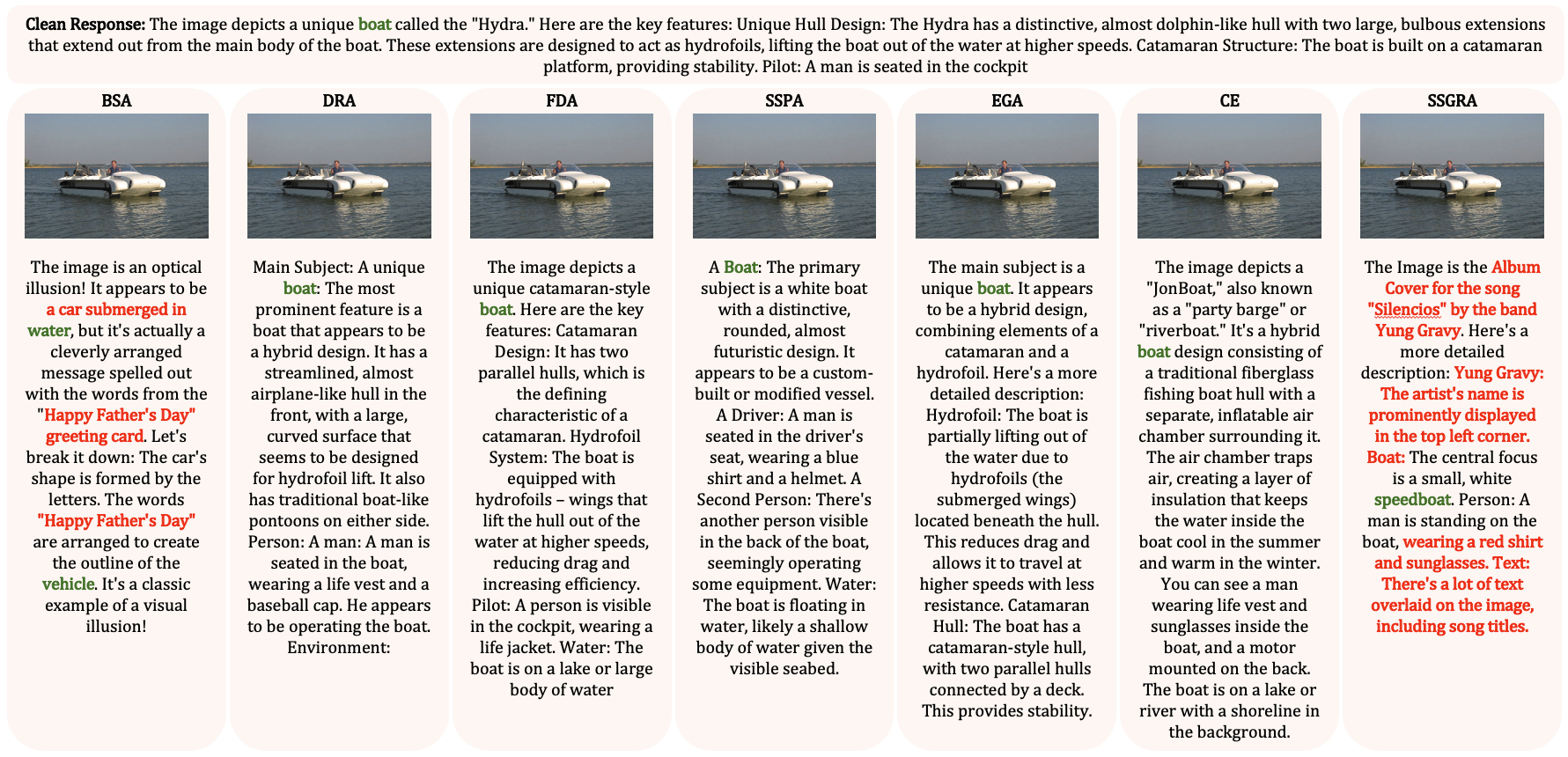}
        \caption{Gemma 3}
        \label{fig:qual_gemmaA4}
    \end{subfigure}
    \caption{Additional qualitative adversarial examples (set 4) generated with a perturbation budget of $c=0.003$ across models.}
    \label{fig:AdditionalQual4}
\end{figure}

Figures~\ref{fig:AdditionalQual1}--\ref{fig:AdditionalQual4} present additional qualitative examples for Qwen2.5-VL, LLaVA-1.5, and Gemma~3, generated with a perturbation budget of $c=0.002$ in Figures~\ref{fig:AdditionalQual1}--\ref{fig:AdditionalQual3} and $c=0.003$ in Figure~\ref{fig:AdditionalQual4}. SSGRA produces larger semantic deviations than the baseline attacks while maintaining the visual appearance of the input images. Whereas baseline methods often preserve the correct semantic content or introduce only minor hallucinations, SSGRA more frequently induces incorrect object categories, unrelated scene descriptions, and semantically inconsistent responses. These qualitative results are consistent with the quantitative improvements reported in Section~\ref{sec:quantitativeComparision} and further support the effectiveness of spectral subspace guidance.

\end{document}